\pdfoutput=1
\documentclass[11pt]{article}
\usepackage{enumerate}
\usepackage[OT1]{fontenc}
\usepackage{amsmath,amssymb}
\usepackage[usenames]{color}
\usepackage{dsfont}
\usepackage{pgfplots}
\usepackage{smile}
\usepackage{multirow}
\usepackage{rotating}
\usepackage{enumerate}
\usepackage{esvect}
\usepackage{tikz}
\usetikzlibrary{patterns}
\usetikzlibrary{arrows}
\usetikzlibrary{bayesnet}
\usepackage{authblk}
\usepackage{caption}

\usepackage[colorlinks,
            linkcolor=red,
            anchorcolor=blue,
            citecolor=blue
            ]{hyperref}
\usepackage{algorithm}
\usepackage{algorithmic}

\def\supp{\mathop{\text{supp}}}

\long\def\comment#1{}

\def\cS{{\mathcal{S}}}

\providecommand{\norm}[1]{\vvvert#1\vvvert}

\newcommand{\bel}{\begin{eqnarray}\label}
\newcommand{\eel}{\end{eqnarray}}
\newcommand{\bes}{\begin{eqnarray*}}
\newcommand{\ees}{\end{eqnarray*}}

\def\rd{\mathrm{d}}

\let\hat\widehat
\let\tilde\widetilde

\def\HH{{\mathbb H}}

\def\supp{\mathop{\text{supp}\kern.2ex}}

\def\given{{\,|\,}}

\def\supp{\mathop{\text{supp}}}

\def\Pie{{\Pi}}
\def\piestar{{\pi^*}}

\def\pie{{\pi}}
\def\pib{{\pi_\text{ob}}}
\def\piepessi{{\hat\pie}}

\theoremstyle{plain}

\usepackage{mathrsfs}
\usepackage{fullpage}

\usepackage{hyperref}
\usepackage[protrusion=false,expansion=true]{microtype}

\def\##1\#{\begin{align}#1\end{align}}
\def\$#1\${\begin{align*}#1\end{align*}}

\usepackage{undertilde}

\theoremstyle{mytheoremstyle}

\def\nend{\nonumber\\}
\def\CI{\text{CI}}

\def\tpr{\tilde{p}}

\def\pr{p}

\def\vT{{\mathcal{T}}}
\def\vhHstar{{\vh_{\vH}^*}}
\def\vhstar{{\vh^*}}
\newcommand\hHstark[1]{{h^*_{\vH, #1}}}
\def\vareH{{\varepsilon_{\vH}}}
\def\vareTheta{{\varepsilon_{\vTheta}}}
\def\vX{X}
\def\vY{Y}
\def\vZ{Z}
\def\vh{h}
\def\vH{\cH}

\def\vTheta{{\Theta}}

\def\vcL{\cL}
\newcommand\thetaTstark[1]{{\theta_{\Theta, #1}^*}}

\def\pob{{\pr_{\text{ob}}}}
\def\EEob{{\EE_{\text{ob}}}}
\def\EEin{{\EE_{\text{in}}}}
\def\CATE{{\text{CATE}}}
\def\aseq{\overset{a.s.}{=}}
\def\CIH{\CI_{\vH, \cD}}
\def\CICATE{\CI_{\cD}}
\newcommand\gpessi[1]{{\hat g^{#1}}}
\def\gHstar{{g_\vH^*}}
\def\gstar{{g^*}}
\newcommand\gH[1]{{g_\cH^#1}}
\newcommand\hstark[1]{h^*_{#1}}
\newcommand\hpessik[2]{\hat h_{#1}^{#2}}
\newcommand\vhpessi[1]{{\hat h^{#1}}}
\def\SubOpt{\text{SubOpt}}
\def\doopt{\text{do}}
\def\vhHpie{{\vh_\vH^\pie}}
\newcommand\hHpiek[1]{{h_{\vH, #1}^\pie}}
\def\gHpie{g_\vH^\pie}
\newcommand\h[2]{h_{#1}^{#2}}
\newcommand{\RNum}[1]{\uppercase\expandafter{\romannumeral #1\relax}}
\def\sF{\mathscr{F}}
\def\sY{\mathscr{Y}}
\newcommand\pin[1]{{p_{\text{in}}^#1}}
\def\tcO{{\tilde\cO}}

\usepackage{subcaption}
\usetikzlibrary{shapes,decorations,arrows,calc,arrows.meta,fit,positioning}
\tikzset{
    -Latex,auto,node distance =1 cm and 1 cm,semithick,
    state/.style ={circle, draw, minimum width = 1 cm},
    missingstate/.style ={circle, draw, minimum width = 1 cm, fill=lightgray},
    hiddenstate/.style ={circle, draw, minimum width = 1 cm, fill=darkgray, text=white},
    point/.style = {circle, draw, inner sep=0.04cm,fill,node contents={}},
    bidirected/.style={Latex-Latex,dashed},
    el/.style = {inner sep=2pt, align=left, sloped},
    normal/.style={-stealth', line width=1},
    double/.style={stealth'-stealth', line width=1},
}

\title{\huge A Unified Framework of Policy Learning for Contextual Bandit with Confounding Bias and Missing Observations}
\author[1]{Siyu Chen \thanks{Email: siyu.chen.sc3226@yale.edu}}

\author[1]{Yitan Wang \thanks{ Email: yitan.wang@yale.edu}}

\author[2]{Zhaoran Wang \thanks{Email: zhaoranwang@gmail.com}}

\author[1]{Zhuoran Yang \thanks{Email: zhuoran.yang@yale.edu}}

\affil[1]{
\small
\textit{Department of Statistics and Data Science, Yale University}}
\affil[2]{
\small
\textit{Department of Industrial Engineering and Management Sciences, Northwestern University}}
\date{}
\begin{document}
\maketitle
\vspace{-25pt}
\begin{abstract}
We study the offline contextual bandit problem, where we aim to acquire an optimal policy using observational data.
However, this data usually contains two deficiencies: (i) some variables that confound actions are not observed, and (ii) missing observations exist in the collected data.
Unobserved confounders lead to a confounding bias and missing observations cause bias and inefficiency problems.
To overcome these challenges and learn the optimal policy from the observed dataset, we present a new algorithm called Causal-Adjusted Pessimistic (CAP) policy learning, which forms the reward function as the solution of an integral equation system, builds a confidence set, and greedily takes action with pessimism.
With mild assumptions on the data, we develop an upper bound to the suboptimality of CAP for the offline contextual bandit problem.
\end{abstract}

\section{Introduction}
Contextual bandit is a mathematical framework that models the  decision-making problem under uncertainty. 
In specific, in a contextual bandit, the agent chooses an action 
based on an observation (also known as the context), and observes a random reward that depends on the observation and the action taken. 
Such a framework 
finds wide applications  in areas such as  healthcare (\cite{raghu2017continuous, prasad2017reinforcement, komorowski2018artificial}), robotics (\cite{pinto2016supersizing}), and computational advertising (\cite{bottou2013counterfactual}).
Typical online policy learning algorithms require many interactions between the agent and the environment.
However, in various applications of offline contextual bandit, e.g., autonomous driving (\cite{shalev2016safe}) and healthcare (\cite{gottesman2019guidelines}), collecting online generated data could take too much cost and be unethical.
On the other hand, there are many historically recorded dataset for tasks where policy learning could be applied.
For example, driving data generated by human (\cite{sun2020scalability}) and medical records (\cite{chakraborty2014dynamic}).
Therefore, in this work, we study the offline policy learning problem, aiming to learn an optimal policy from previously collected dataset and require no interactions with the environment.

To learn the optimal policy from an offline dataset in real-world practice, three challenges must be addressed: confounding effects, partially missing observations, and partial data coverage. 
Confounding effects arise because it is often impossible to conduct randomized controlled trials or collect all necessary covariates. \citep{pearl2009causality, hernan2010causal}.
For instance, medical data may intentionally omit a patient's health condition and medical history due to privacy concerns. \citep{brookhart2010confounding}.
In this example, the hidden information serves as the confounders. To address this, the power of side observations is used to adjust for unobserved confounders.

However, key information like side observations or context are often subject to partial missingness due to various reasons such as privacy concerns, attrition, or experimental errors. For example, lab test results that serve as side observations may be lost due to inability to conduct the tests or failure to store the results. Even worse, as the causes of missing observation problem vary, the missing pattern might be either at random or not at random \citep{rubin1976inference, yang2019causal}. When the partial missingness is at random, the full data distribution is identifiable and multiple methods can provide reasonable estimates \citep{qu2009propensity, seaman2014inverse}. On the other hand, missingness not at random is a more challenging issue, since the missingness mechanism can depend on other factors or even the missing value itself. In this work, we aim to handle the challenge of side observations missing not at random.

The last main challenge is the partial coverage of the actions in the offline dataset. the observational dataset is collected beyond the control of the learner, and the action space can usually be too large for sufficient exploration. This means that conditions on full coverage of the action space by the data set in order to learn a good interventional policy usually fails to hold in real world practice \citep{fujimoto2019off, fu2020d4rl}. In this paper, we therefore explore the question:
 \begin{center}
{\it Is it possible to design a provably efficient algorithm for offline learning confounded contextual \\ bandit problem with side observations and missing values under mild assumptions on the dataset?}
\end{center}

Our answer to this question is affirmative.
Fortunately, there are methods that can help us accurately estimate the value of a policy from offline data affected by unobserved confounders if we have access to some side observations.
Two typical examples of side observations we explore in this work are instrumental variables (IV) (\cite{baiocchi2014instrumental, wong2021integral, chen2011nonlinear}) and proxy variables (PV) (\cite{miao2018identifying}).
Informally, IVs are variables that affect the reward only through the action, while PVs serve as negative control for the confounding effects.
Therefore, we investigate the use of side observations to mitigate the confounding bias in this paper.

For the partially missing observation challenge, we consider the case where both the context of the bandit and the side observations collected in the offline dataset are subject to missingness not at random. 
Let us take the missingness of context $X$ for example. Let $R_X$ denote the binary missingness indicator for $X$. We assume that $X$ is totally missing if $R_X=0$. 
The key idea we adopt to overcome the missingness issue is using the \textit{distributional information} of the outcome to compensate for the missingness in the contexts or oberservations under certain completeness conditions.
Similar ideas can be found in \citep{yang2019causal, ding2014identifiability}, though they consider a different setting with missing elementary values in a vectorized random variable.
To address the challenges of confounding effects and non-random missing observations, we present a new approach that formulates the policy evaluation problem as solving an integral equation system (IES) with bridge functions. We demonstrate that the solution to the IES preserves the conditional average treatment effect (CATE), which enables us to further optimize the policy. In contrast to conventional causal methods that rely only on summary statistics such as the expectation of the outcome, our method leverages the full distributional information of the outcome.

Given the limitations of finite samples and incomplete coverage of the action set in our offline dataset, we incorporate the principle of pessimism into our approach for learning the optimal policy \citep{jin2021pessimism, buckman2020importance, xie2021bellman}.
In order to do pessimistic policy optimization, We must quantify the uncertainty in estimating the CATE from the IES.
To do so, we propose a Causal-Adjusted Pessimistic (CAP) policy optimization algorithm, which has two components: the uncertainty quantification step for estimating the CATE from the IES, and the policy optimization step based on the uncertainty quantification result.
In the CAP algorithm, the IES is solved by reducing the moment restriction equation system to a minimax problem via Fenchel duality.
After the uncertainty quantification step, the CAP algorithm takes greedy policy based on the confidence set constructed.



Our contribution can be summarized in three perspectives. 
\textbf{First,} we developed a general framework to model the contextual bandit problem with confounded offline dataset and missing observations not at random.
Under this framework, we derive a novel integral equation system (IES) for identification.
\textbf{Second, } 
We convert the IES to a minimax optimization problem, whose solution respects the CATE and the loss function of the minimax optimization problem paves the way for uncertainty quantification.
We emphasize that when the side observations are likely missing and the conditional moment restrictions form a system of equations rather than a single equation, it is non-trivial to select a proper way to do uncertainty quantification.
\textbf{Finally,} we propose a Causal-Adjusted Pessimistic (CAP) policy optimization algorithm and prove that our algorithms achieves fast statistical rate of sub-optimality for contextual bandit with side observations serving as instrumental variable and proxy variables.
To the best of our knowledge, this paper is the first one proposing a provably efficient algorithm for offline confounded contextual bandit with observations missing not at random.

\subsection{Related Works}
\paragraph{Offline reinforcement learning.}
Literatures on offline reinforcement learning, especially those on pessimistic algorithms, are related to our work (\cite{antos2007fitted, munos2008finite, chen2019information, liu2020provably, zanette2021exponential, xie2021bellman, yin2021towards, rashidinejad2021bridging, zhan2022offline, yin2022near, yan2022efficacy}).
The major difficulty of offline reinforcement learning is the distribution shift between the policy generating the collected data and the class of target policies.
To overcome the distribution shift problem, we have to incorporate pessimism \citep{jin2021pessimism, xie2021bellman, buckman2020importance, uehara2021pessimistic}.

\paragraph{Causal inference.}
A series of previous works in causal inference address adjusting the confounding bias via use of side observations like instrumental variables \citep{chen2011nonlinear, chen2021efficient, chen2016methods, bennett2019deep, bennett2023minimax, wong2021integral, athey2019estimating} or proxy variables \citep{bennett2021proximal, pearl2009causality, miao2018identifying, lee2021causal}.
Among these works, \cite{pearl2009causality, miao2018identifying, lee2021causal} are most relevant to this paper since the our framework covers side observations proposed in them.
In comparison, these works studied the identification of certain side observations.
As the task in this paper is to learn the optimal policy, we make further effort to construct confidence sets.
Additionally, a large volume of works discuss the missing observation problem in causal inference (\cite{rubin2004multiple, qu2009propensity, crowe2010comparison, mitra2011estimating, seaman2014inverse, yang2019causal}).
\cite{yang2019causal} is particularly related to our work as it discusses identification for the case in which confounders are missing not at random.
The model in \cite{yang2019causal} assumes that there is chance to observe some data of confounders, while we study a model assuming confounders are completely missing and the side observations are missing not at random.


\subsection{Preliminaries}
\paragraph{Notations.}
In this paper, we let $\Delta(\cA)$ denote distributions over $\cA$.
We denote the inner product by  $\langle \cdot, \cdot \rangle$.
For any function $f$, we let $\cO(f)$ denote $Cf$ and $\tilde{\cO}(f)$ to denote $\cO(f\cdot\text{poly}(\log f))$.
We use calligraphic symbols $\sF$ and $\vH$ to represent function classes.
We use $\overset{\cE}{\lesssim}$ and $\overset{\cE}{\gtrsim}$ to represent the inequality that holds on some event $\cE$.

\paragraph{Critical radius.} We define the localized empirical Rademacher complexity with respect to data set $\cD=\{x_i\}_{i=1}^n$ and function class $\sF:\cX\rightarrow [-c, c]$ as 
\begin{align*}
    \cR_\cD(\eta; \sF)=\EE\sbr{\sup_{f\in\sF, \nbr{f}_\cD\le \eta}\frac 1 n \sum_{i=1}^n\varepsilon_i f(x_i)}.
\end{align*}
The critical radius of $\sF$ on dataset $\cD$ is defined as any positive solution $\eta_\cD$ to $\cR_\cD(\eta;\sF)\le \eta^2/c$. Note that the critical radius $\eta_\cD$ is also a random quantity. 

\subsection{Roadmap}
In \S\ref{sec:problem formulation} we formalize the problem of policy learning for contextual bandit problem with confounding bias and missing observations.
In \S\ref{sec:CAP algorithm} we discuss the challenges of policy learning problem shown in \S\ref{sec:problem formulation} and show how these challenges motivate us to develop an algorithm framework named CAP policy learning to solve them.
In \S\ref{sec:Identification} and \S\ref{sec:Estimation} we expand details about the step of constructing confidence set in the CAP algorithm described in \S\ref{sec:CAP algorithm}.
The convergence results for the CAP algorithm are provided in \S\ref{sec: theoretical results}.
In \S\ref{sec:extended CCB-PV} we give convergence analysis of the CAP algorithm in an extended policy class.
Lastly, we show that the CAP algorithm could be applied to the linear Dynamic Treatment Regime (DTRs) problem in \S\ref{sec:DTR} and to the one-step linear Partially Observable Markov Decision Process (POMDP) in \S\ref{sec:POMDP}, both with the sub-optimality guaranteed to converge at a rate of $\tcO(T^{-1/2})$.



%
%
%

\section{Problem Formulation}\label{sec:problem formulation}
In this section, we formalize the contextual bandit problem with confounding bias and missing observations.
We describe the casual structure of confounded contextual bandit in \S\ref{sec:confounded contextual bandit}, the procedure of data collecting in \S\ref{sec:observational process}, and the performance metric in \S\ref{sec:interventional process}.

\subsection{Confounded Contextual Bandit}\label{sec:confounded contextual bandit}
In this paper we study the offline policy learning in confounded contextual bandit (CCB).
Each trial can be represented by a tuple of random variables
$$(U, X, A, Y, O), $$
where $U\in\cU$ is the confounder, $X\in\cX$ is the context, $A\in\cA$ is the treatment, $Y\in\mathbb{R}$ is the reward, and $O\in\cO$ denotes the side observations. 
We assume that variables for different trials are independent and identically distributed. 
In offline learning, the data is collected through the observational process, and a newly selected policy is carried out in the interventional process.

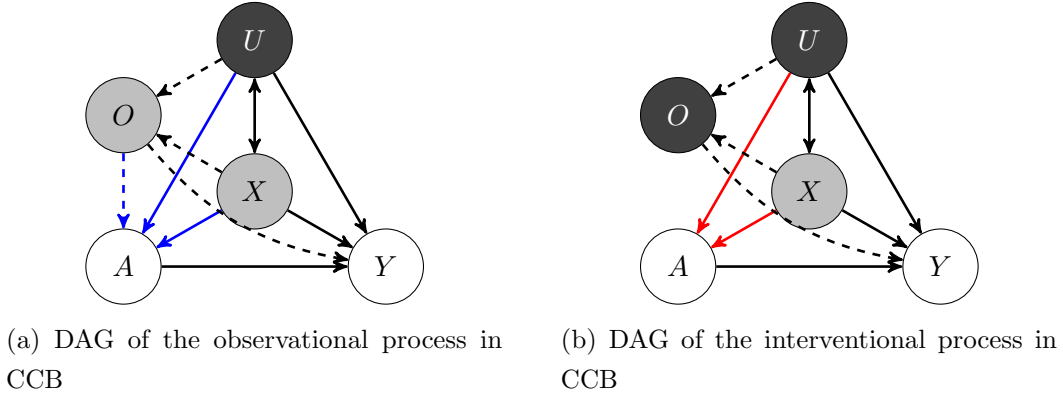
\begin{figure}[h]  
\centering 
  \begin{subfigure}[b]{0.4\linewidth}
  \centering
    \begin{tikzpicture}
        \node[hiddenstate] (U) {$U$};
        \node[missingstate] (X) [below=of U] {$X$};
        \node[state] (A) [left=of X, xshift=.27cm, yshift=-1cm] {$A$};
        \node[state] (Y) [right=of X, xshift=-.27cm, yshift=-1cm] {$Y$};
        
        \path[double] (U) edge (X);
        \path[blue, normal] (U) edge (A);
        \path[blue, normal] (X) edge (A);
        \path[normal] (A) edge (Y);
        \path[normal] (U) edge (Y);
        \path[normal] (X) edge (Y);

        \node[missingstate] (O) [above=of A] {$O$};
        \path[blue, dashed, normal] (O) edge (A);
        \path[dashed, normal] (U) edge (O);
        \path[dashed, normal] (X) edge (O);
        \path[dashed, normal] (O) edge[bend left=-20] (Y);
    \end{tikzpicture}%
    \caption{DAG of the observational process in CCB} \label{fig:CCB_behavior}
  \end{subfigure}\qquad
\begin{subfigure}[b]{0.4\linewidth}
\centering
  \begin{tikzpicture}  
      \node[hiddenstate] (U) {$U$};
        \node[missingstate] (X) [below=of U] {$X$};
        \node[state] (A) [left=of X, xshift=.27cm, yshift=-1cm] {$A$};
        \node[state] (Y) [right=of X, xshift=-.27cm, yshift=-1cm] {$Y$};
        
        \path[double] (U) edge (X);
        \path[red, normal] (U) edge (A);
        \path[red, normal] (X) edge (A);
        \path[normal] (A) edge (Y);
        \path[normal] (U) edge (Y);
        \path[normal] (X) edge (Y);

        \node[hiddenstate] (O) [above=of A] {$O$};
        \path[dashed, normal] (U) edge (O);
        \path[dashed, normal] (X) edge (O);
        \path[dashed, normal] (O) edge[bend left=-20] (Y);
\end{tikzpicture}
\caption{DAG of the interventional process in CCB} \label{fig:CCB_evaluation} 
\end{subfigure}
\caption{A DAG illustrating the observational and the interventional process in CCB. Here,  the white nodes represent the observed variables, the light gray nodes represent the variables missing not at random and the dark nodes represent unmeasured variables. A dashed line means the causal effect might either exist or not. }\label{fig:CCB}
\end{figure}

The observational process and the interventional process of a CCB with side observations are depicted in Figure \ref{fig:CCB}.
In each trial, there is an unmeasured confounder $U$ which has impacts on $O$, $A$, $X$, and $Y$.
As $U$ is not measured, the value of $U$ is not accessible in the data collected.
For example, sensitive information that is not allowed to reveal could be modeled by such unmeasured confounder. 
Since $U$ affects all of $O$, $A$, $X$, and $Y$, the confounder $U$ serves as a common cause for the model.
Context $X$ is coupled with confounder $U$, and side observations $O$ can be caused by both confounder $U$ and context $X$. 
A treatment $A$ is then selected following some policy $\pib(\cdot|U, X, O)$.
After the treatment is carried out, the environment generates a reward $Y$.
Suppose there are $T$ trials in total, then the \textbf{full dataset} $\tilde{\cD}$, which is distinguished from the dataset $\cD$ defined later, can be represented as
$$\tilde{\cD}=\{(u_t, x_t,a_t,y_t,o_t)\}_{t=1}^T.$$
Then by some missingness mechanism, records of $x_t$ and $o_t$ in some trials are possibly lost.
Additonally, recall that $u_t$ is also unmeasured and thus not included in the dataset.
The dataset collected for policy learning is defined as
$$\cD=\{(\check x_t, a_t, y_t, \check o_t)\}_{t=1}^T,$$
where $\check x_t$ either takes the value of $x_t$, if $x_t$ is not lost, or takes a special value of $\text{None}$, if $x_t$ is lost.
Similarly, $\check o_t$ either takes value of $r_t$ or $\text{None}$, depending on whether $o_t$ is lost.
More details about the observational process is presented in \S\ref{sec:observational process}.

Given the dataset $\cD$ collected in the observational process, the goal of policy learning in CCB is to build a new policy $\pi$, which is called the interventional policy.
The interventional policy $\pi$ is executed in the interventional process.
In the interventional process, the unmeasured confounder $U$ still has impacts on both the context $X$ and the reward $Y$.
As the confounder $U$ is unmeasured, the agent could only observe the context $X$ and must decide an action $A$ to take only depending on the context $X$.
The rules for the agent to make decision in the interventional process is modeled by the interventional policy $\pi: \cX\rightarrow\Delta(\cA)$.
The agent aims to learn an interventional policy $\pi$ that maximizes the expected reward.
We discuss more details about the interventional process and performance metric in \S\ref{sec:interventional process}.


\subsection{Observational Process}\label{sec:observational process}
The observational process describes how the offline dataset is collected, which is depicted in Figure \ref{fig:CCB_behavior}.
In the $t$-th trial of the observational process, the environment selects $(u_t, x_t, o_t)$ as a realization of $(U, X, O)$ according to the prior $\pr(u, x, o)$.
The agent then conducts a treatment according to the observational policy $\pib: \cX\times \cO \times \cU\rightarrow \Delta(\cA)$. We remark that it is very common for the observational policy to be confounded by $U$, which can be understood as the agent's natural predilections, e.g., the playing agent in the observational procedure has a preference for certain treatments due to some hidden causes encoded by $U$ \citep{bareinboim2015bandits}.
After the treatment is conducted, a reward $Y$ depending on $(U, X, O, A)$ is received by the agent. The joint distribution $\pob$ in the observational process is thereby given by
\begin{align}\label{def:pob}
    \pob(u, x, o, a, y) = \pr(u, x, o) \cdot  \pib(a\given u, x, o) \cdot \pr(y\given u, x, o, a).
\end{align}
Here we provide two typical examples of side observations in the observational process.
\begin{example}[Side observations as instrumental variable]\label{ex:IV}
In a confounded contextual bandit with instrumental variable (CCB-IV) shown in Figure \ref{fig:CCB-IV}, $O$ corresponds to instrumental variable $Z$, which is assumed to be independent of confounder $U$ and outcome $Y$. The observational policy is given by $\pib(a\given u, x, z)$.
\end{example}
\begin{example}[Side observations as proxy variables]\label{ex:PV}
In a confounded contextual bandit with proxy variables (CCB-PV) shown in Figure \ref{fig:CCB-PV}, $O$ corresponds to the negative controls $(Z, W)$. It is assumed that $W\indep A\given (U, X)$ holds for the outcome proxy $W$ and $Z\indep (Y, W)\given (A, X, U)$ holds for the treatment proxy $Z$. The observational policy is given by $\pib(a\given u, x, z)$.
\end{example}
Suppose there are $T$ trials in the observational process and the full dataset is denoted by $\tilde \cD = \{(u_t, x_t, a_t, y_t, o_t)\}_{t=1}^T$.
We emphasize that $\tilde\cD$ is not the dataset that will be used for policy learning due to the unmeasurement of the confounder $u_t$ and the missingness in both the contexts $x_t$ and the side observations $o_t$.
We formally discuss the missingness mechanism in the following paragraph.

\paragraph{Missingness Mechanism. }
In addition to the unmeasurement of confounder $U$, we assume that side observations $O$ and context $X$ in our model are subject to missingness, which extends the missingness mechanism in \cite{yang2019causal, yang2017nonparametric}  
where missingness was assumed to be not at random but independent of outcome $Y$.
We denote the observed dataset as $\cD = \{(\check x_t, a_t, y_t, \check o_t)\}_{t=1}^T$, where $\check x_t$ and $\check o_t$ denote the context and the side observations that we truly observe.  
Let random variables $R_X$ and $R_O$ denote the missingness indicators for $X$ and $O$, respectively.
Let $r_{X, t}$ and $r_{O, t}$ denote realizations of $R_X$ and $R_O$ in the $t$-th trial.
When $r_{X,t}=1$, the record of $x_t$ is not missing and $r_{X,t}=0$ indicates that the record of $x_t$ is lost.
We introduce a special dummy value $\text{None}$ to represent a missing record.
So $\check x_t$ takes values in $\{x_t, \text{None}\}$ by the rule of
\begin{equation*}
    \check x_t = \left\{
    \begin{aligned}
    & x_t & \quad \text{if } r_{X,t} = 1 \\
    & \text{None} & \quad \text{if } r_{X,t} = 0
    \end{aligned}
    \right.
\end{equation*}
In contrast to assuming the observations to be missing randomly \citep{rubin2004multiple, qu2009propensity, crowe2010comparison, mitra2011estimating, seaman2014inverse}, in this paper we study a more general and challenging setting in which the missingness is not at random, i.e., $R_X$ and $R_O$ are not independent of the model $(U,X,A,Y,O)$.
As $R_X$ and $R_O$ could be dependent of $(U,X,A,Y,O)$, our results covers the case of malicious adversarial missing.
We revisit the previous two examples to illustrate the missingness.
\begin{example}[Example \ref{ex:IV} revisited]
In the CCB-IV, we allow $R_X$ to be caused by $(Z, X, A)$ and $R_Z$ to be caused by $(Z, X)$.
\end{example}
\begin{example}[Example \ref{ex:PV} revisited]
In the CCB-PV, we allow $R_X$ to be caused by $X$, $R_Z$ to be caused by $(Z, U, A, X)$, and $R_W$ to be caused by $(W, X, A)$.
\end{example}
Identifying the causal effect in the presence of missingness is nontrivial because the missingness interferes with the structure of the observational dataset.
For instance, when conditioning on $R_X=1$ in the CCB-IV, the instrumental variable $Z$ might no longer be independent of the confounder $U$, which leads to failure of conventional identification approaches.
More details on the difficulties brought by the missingness mechanism as well as the method we use to address the missingness issue will be provided case by case in \S\ref{sec:Identification}.

\subsection{Interventional Process} \label{sec:interventional process}
In the interventional process, an interventional policy is carried out after the model is learned from the offline dataset.
The interventional process is different from the observational process in the following three aspects: 
\begin{description}
\item[(i)] side observations $O$ appearing in the dataset are unmeasurable while context $X$ is fully measurable in the interventional process;
\item[(ii)] the agent follows an interventional policy $\pie:\cX\rightarrow \Delta(\cA)$ which is independent of $U$ and $O$ since they are unable to measure in the interventional process;
\item[(iii)] context $X$ follows a new marginal distribution $\tilde p(x)$ in the interventional process.
\end{description}
Aspect (i) indicates that only the context is revealed to the agent in the interventional process. Therefore, the interventional policy is context-dependent, as is stated in (ii).
We remark that (iii) can be understood through the idea of a marginal distribution shift in $X$ between the observational group and the interventional group, which is very common in real-world practice.
For example, when studying the effect of recommended ads' type ($A$) on the clicking rates ($Y$) with users' age ($X$) serving as the context, we might have an interventional group whose age distribution differs from the observational group.
Another example is the in-context learning paradigm,  where the task specification procedure can be viewed as "conditioning" the model on a certain context presented by the input texts/token \citep{brown2020language,radford2019language}.
Following (i)-(iii), the joint distribution $\pin{\pie}$ of random variables in the interventional process is given by
\begin{align}\label{def:pin}
    \pin{\pie}(u, x, o, a, y) = \tpr(x) \pr(u, o\given x) \pie(a\given x) \pr(y\given u, x, o, a).
\end{align}
Diagrammatic explanations of the interventional process are given in Figure \ref{fig:CCB_evaluation}. 
We see that the DAG of the interventional model is given by substituting the blue incoming edges to treatment $A$ encoded by $\pib$ in the observational model with the orange incoming edges encoded by $\pie$, while the remaining part of the DAG remains unchanged except for the marginal distribution of $X$.

\paragraph{Reward function and policy optimization.}
In the interventional process, the average reward $v^\pie$ is defined as
\begin{align}
    v^\pie = \EE_{\pin{\pie}}\sbr{Y},\label{def:v}
\end{align}
where $\EE_{\pin{\pie}}$ corresponds to the expectation taken with respect to $\pin{\pie}$ defined in \eqref{def:pin}.
Our target is to find  $\piestar\in\Pie:\cX\rightarrow \Delta(\cX)$ that optimizes the average reward, 
\begin{align*}
    \piestar= \arg\max_{\pie\in\Pie} v^\pie.
\end{align*}
Correspondingly, we define the performance metric as the following sub-optimality, 
\begin{align}
    \text{SubOpt}(\pie) = v^\piestar - v^\pie.\label{def: SubOpt}
\end{align}

In summary, our goal is to design a learning algorithm that returns a policy $\hat \pi$ based on the offline dataset $\cD$ collected in the observational process. Here the dataset is subject to unmeasured confounder and missingness. 



\section{CAP Algorithm}\label{sec:CAP algorithm}
In this section, we first investigate the main challenges of such an offline bandit problem, including the issue of confounding and missing data and also the spurious correlation that arises in the decomposition of the sub-optimality in \S\ref{sec:challenges}.
We then put forward an algorithm framework named Causal-Adjusted Pessimistic (CAP) policy learning in the face of such challenges in \S\ref{sec:algorithm outline}.
\subsection{Challenges in the Offline Setting }\label{sec:challenges}
The offline learning problem in the CCB boils down to the following two questions:
(i) how to evaluate the average reward given an interventional policy; (ii) how to efficiently find an interventional policy that maximizes the average reward.
When trying to answer these two questions, we encounter two major challenges: (i) confounded and missing data; (ii) spurious correlation in the sub-optimality. 
We briefly discuss where these challenges stem from and what technologies we use to overcome these challenges in this subsection.
\paragraph{Challenges in average reward evaluation: confounded and missing data.}
The key to the problem of evaluating the average reward \eqref{def:v} in the interventional process is to learn the conditional average treatment effect (CATE) defined as
\begin{align*}
    \gstar(x, a)=\EEob\sbr{Y\given X=x, \doopt(A=a)}, 
\end{align*}
where $\EEob$ is an abbreviation for $\EE_{\pob}$.
Here, the do-calculus $\doopt(A=a)$ in the condition means that the expectation is taken with respect to the distribution obtained by deleting $\pib(a\given u, x, o)$ from the product decomposition of $\pob$ in \eqref{def:pob} and restricting $A=a$. 
Learning the CATE is important since the average reward is related to the CATE by
\begin{align}\label{eq:v to CATE}
    v^\pie = \EE_{\pin{\pie}}\sbr{\gstar(X, A)}.
\end{align}
In the presence of confounding bias \citep{vanderweele2008causal, jager2008confounding}, learning the CATE needs tools borrowed from causal inference.
To control for the confounding bias, a typical way is to exploit side observations $O$ in the offline data \citep{lipsitch2010negative, singh2020kernel}.
Instances of controlling the confounding bias using side observations are presented in Examples \ref{ex:IV} and \ref{ex:PV} where instrumental variable (IV) \citep{cragg1993testing, arellano1995another,  newey2003instrumental} or proxy variables (PV) \citep{tchetgen2020introduction, ying2021proximal} are introduced for negative controls.

However, our problem is still challenging given the fact that the missingness bias is coupled with the confounding bias. 
Note that identification with outcome-independent missingness is rather trivial in the unconfounded contextual bandit setting with tuple $(X, A, Y, R_X)$ where $R_X$ is caused by $(X, A)$.
The simplest way is to use the dataset without missingness, i.e., conditioning on $R_X=1$ and estimate $\EEob\sbr{Y\given X=x, A=a, R_X=1}$. 
Such a method is valid since we have $R_X\indep Y\given (X, A)$ without confounders.
However, in the confounded contextual bandit setting, the causal effect is identified with the aid of side observations, and some model assumptions related to these side observations are broken when simply conditioning on $(R_X, R_O)=\ind$.
Take the CCB-IV case for instance. The IV independence assumption $Z\indep U\given X$ is broken by conditioning on $R_X=1$ since $R_X$ also depends on confounded action $A$. 
As we will show in \S\ref{sec:Identification}, the CATE learning problem is addressed by solving a novel integral equation system (IES), in which the integral equations are coupled together, and the CATE is obtained as the solution to the IES.


\paragraph{Challenges in policy optimization: spurious correlation.}
We discuss the second question on how to efficiently optimize the interventional policy.
Let $g$ denote an estimation of the CATE and $\gstar$ denote the exact $\CATE$ thereafter.
Following \eqref{eq:v to CATE}, we define the average reward function corresponding to $g$ and $\pie$ as
\begin{align}
    v(g, \pie) = \EE_{\pin{\pie}}\sbr{g(X, A)}.\label{def:v approx}
\end{align}
It is okay if we simply take a greedy policy $\hat \pi$ that maximizes $v(g, \pie)$. However, such a greedy policy can sometimes be misleading.
A little calculation of the sub-optimality helps gain intuition. 
We let $\tpr(x)=\ind(x=x_0)$ for brevity. By definition of the sub-optimality in \eqref{def: SubOpt} and the fact that $\hat \pie$ is greedy with respect to $g$, we have
\begin{align}
    \SubOpt(\hat \pi) \le \underbrace{\langle \gstar(x_0, \cdot)-g(x_0, \cdot), \piestar(\cdot\given x_0) \rangle}_{\text{(i)}} + \underbrace{\langle g(x_0,\cdot) - \gstar(x_0, \cdot), \hat\pie(\cdot\given x_0) \rangle}_{\text{(ii)}}. \label{eq:spurious subopt}
\end{align}
Note that $\piestar$ in term (i) is intrinsic to the bandit model and does not depend on $g$. In contrast, $\hat \pie$ in term (ii) is coupled with the estimated $g$, which yields the spurious correlation \citep{jin2021pessimism} and makes term (ii) hard to control.
Bounding term (ii) usually needs strong assumptions on the "uniform coverage" of the dataset $\cD$ as in the existing bandit and RL literature \citep{brandfonbrener2020bandit, tennenholtz2021bandits, laroche2019safe}, which occasionally fails to hold in practice.

Instead, we adopt the technique of uncertainty quantification and pessimism to cope with the spurious correlation challenges. Similar techniques have been applied to other problems in the existing literature \citep{jin2021pessimism, uehara2021pessimistic, zhan2022offline, rashidinejad2021bridging}. Our work successfully integrates such techniques with the confounded and missing data scenarios. Specifically, we first construct a confidence set $\CICATE$ for the estimated $g$ based on the offline data such that $\gstar\in\CICATE$ holds with high probability.
If the policy is optimized with respect to $g\in\CICATE$ that minimizes $v(g, \cdot)$, it follows that $v(g, \hat \pie)\le v(\gstar, \hat \pie)$, and the spurious correlation in term (ii) vanishes. Then, the estimated policy is given by
\begin{align*}
    \piepessi = \arg\sup_{\pie\in\Pie} \inf_{g\in\CICATE} v(g, \pie).
\end{align*}
Moreover, it is also shown that pessimism can promote exploration \citep{auer2008near, azar2017minimax} and help weaken the assumptions on the concentrability coefficient or the data coverage \citep{buckman2020importance}.

\subsection{Algorithm Outline}\label{sec:algorithm outline}
Now that we have answered the two questions raised in the last subsection by (i) identifying the CATE from an integral equation system (IES); (ii) optimizing the policy with a pessimistic estimator $g$ selected from some confidence set $\CICATE$.
What remains to clarify is how to construct the confidence set $\CICATE$ based on the IES.
As we will show in  \S\ref{sec:Estimation}, learning the CATE from the IES can be alternatively done by minimizing some empirical loss function $\cL_\cD(\vh)$ on the hypothesis class $\vH$, where $\vh$ is an estimated solution to the IES.
We are then inspired to construct the confidence set as a level set of $\vH$ with respect to metric $\cL_\cD(\cdot)$ and a threshold $e_\cD$. 
The whole procedure is summarized in the following Causal-Adjusted Pessimistic (CAP) policy learning algorithm.

\begin{algorithm}
\caption{Causal-Adjusted Pessimistic (CAP) policy learning}\label{alg:meta}
\small
\begin{algorithmic}
\REQUIRE dataset $\cD = \{\check x_t, a_t, y_t, \check o_t\}_{t=1}^T$ from the observational process, hypothesis space $\vH$, policy class $\Pie$, threshold $e_\cD$.
\STATE (i) Construct confidence set $\CICATE(e_\cD)$ as the level set of $\vH$ with respect to metric $\cL_\cD(\cdot)$ and threshold $e_\cD$.
\STATE (ii) $\piepessi=\arg\sup_{\pie\in\Pie}\inf_{g\in \CICATE(e_\cD)} v(g, \pie)$.
\ENSURE $\piepessi$.
\end{algorithmic}
\end{algorithm}
The IESs for identifying the CATE in both the CCB-IV and the CCB-PV settings are formulated in \S\ref{sec:Identification}, and a united form is presented with use of a linear operator $\cT$. 
Based on such a united form, the loss function $\cL_\cD(\cdot)$ and the confidence set $\CICATE$ is constructed using the technique of minimax estimator. More details about constructing $\CICATE$ are provided in \S\ref{sec:Estimation}. 

\section{Causal Identification of CATE}\label{sec:Identification}
In this section, we show how to identify the CATE for CCB-IV and CCB-PV with missingness. Under certain completeness assumptions, the CATE is learnable through solving an integral equation system (IES). We also give explanations for the IES in the matrix form and compare the IES to standard identification equations without missingness to highlight how we address the missingness issue. A united form for the IES with a linear operator $\cT$ is provided in \S\ref{sec:united form}.

\subsection{Identification in Confounded Contextual Bandit with Instrumental Variable}

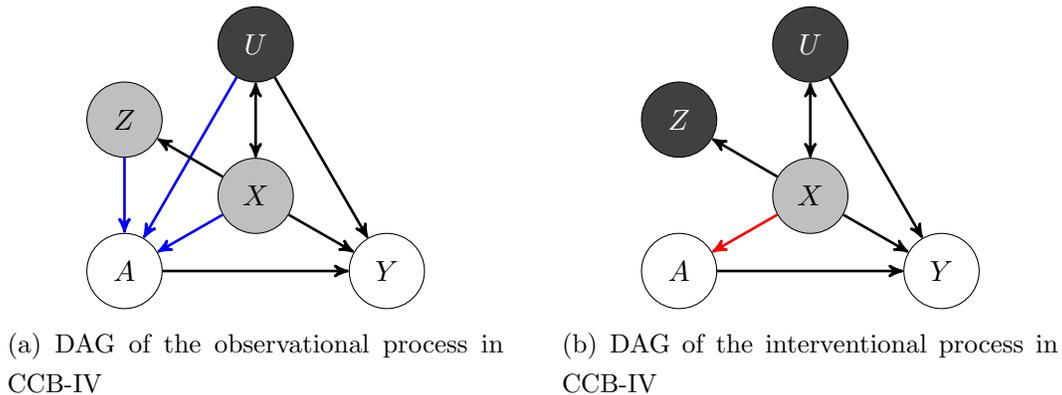
\begin{figure}[h]  
\centering 
  \begin{subfigure}[b]{0.4\linewidth}
  \centering
    \begin{tikzpicture}
        \node[hiddenstate] (U) {$U$};
        \node[missingstate] (X) [below=of U] {$X$};
        \node[state] (A) [left=of X, xshift=.27cm, yshift=-1cm] {$A$};
        \node[state] (Y) [right=of X, xshift=-.27cm, yshift=-1cm] {$Y$};
        
        \path[double] (U) edge (X);
        \path[blue, normal] (U) edge (A);
        \path[blue, normal] (X) edge (A);
        \path[normal] (A) edge (Y);
        \path[normal] (U) edge (Y);
        \path[normal] (X) edge (Y);

        \node[missingstate] (Z) [above=of A] {$Z$};
        \path[blue, normal] (Z) edge (A);
        \path[normal] (X) edge (Z);
    \end{tikzpicture}%
    \caption{DAG of the observational process in CCB-IV}\label{fig:IV behavior}
  \end{subfigure}\qquad
\begin{subfigure}[b]{0.4\linewidth}
\centering
  \begin{tikzpicture}  
        \node[hiddenstate] (U) {$U$};
        \node[missingstate] (X) [below=of U] {$X$};
        \node[state] (A) [left=of X, xshift=.27cm, yshift=-1cm] {$A$};
        \node[state] (Y) [right=of X, xshift=-.27cm, yshift=-1cm] {$Y$};
        
        \path[double] (U) edge (X);
        \path[red, normal] (X) edge (A);
        \path[normal] (A) edge (Y);
        \path[normal] (U) edge (Y);
        \path[normal] (X) edge (Y);

        \node[hiddenstate] (Z) [above=of A] {$Z$};
        \path[normal] (X) edge (Z);
\end{tikzpicture}
\caption{DAG of the interventional process in CCB-IV}\label{fig:IV evaluation}
\end{subfigure}
\caption{A DAG illustrating the introduction of side observations ($O=Z$) in CCB-IV and the difference between the observational model and the interventional model in CCB-IV. Here,  the white nodes represent observed variables, the light grey nodes represent the variables with missingness not at random and the dark nodes represent unmeasured variables. A line with arrows at both ends means that the causal effect going in each way is allowed and therefore, the direction is not specified.}\label{fig:CCB-IV}
\end{figure}


Instrumental variable (IV) regression is a method in causal statistics for estimating the confounded causal effect of treatment $A$ on outcome $Y$. Researchers in
economics employ IV to overcome issues of strategic interaction, e.g., supply cost shifters ($Z$) only influence sales ($Y$) via price ($A$), thereby identifying counterfactual demand even though prices are confounded by supply and demand market forces \citep{wright1928tariff, blundell2012measuring}.

Our model for confounded contextual bandit with instrumental variable (CCB-IV) is illustrated in Figure \ref{fig:CCB-IV}. In contrast to the standard IV model without context, we assume the IV to depend on the context $X$, for the reason that the IV usually appears as a recommendation of the treatment given by an advisor based on the current context.
The model assumptions for CCB-IV are summarized as follows.
\begin{assumption}[Model assumptions for observational process in CCB-IV]\label{asp:CCB-IV}
We assume that the following assumptions hold for the observational process of the CCB-IV.
\begin{itemize}
    \item[(i)] (Structured reward). $Y=f(A, X)+\epsilon$ where $\epsilon \indep A\given (X, U)$;
    \item[(ii)] (IV completeness). $\EEob[\sigma(X, A)\given Z=z, R_Z=1] = 0$ holds for all $z\in\cZ$ if and only if $\sigma(X, A)=0$ holds almost surely;
    \item[(iii)] (IV independence). For the IV, we assume that $Z\indep (U, \epsilon)\given X$;
    \item[(iv)] (Unconfounded and outcome-independent missingness). 
    We allow $R_X$ to be caused by $(Z, X, A)$ and $R_Z$ to be caused by $(Z, X)$.
\end{itemize}
\end{assumption}
The model assumption in (i) can be viewed as a generalization of the semi-parametric contextual bandits, whose outcome is given by $Y(a)=\langle \theta, X_a\rangle+g(X)+\epsilon$ when selecting treatment $a$ \citep{krishnamurthy2018semiparametric}.
The context $X=(X_{a_1}, \cdots, X_{a_{|\cA|}})$ is a tuple and each element corresponds to a feature for an action.
Moreover, $\epsilon$ can be viewed as a treatment-independent noise.
The IV completeness assumption in (ii) ensures that different IV ($Z$) generates enough variation in $(X, A)$ and the IV independence assumption in (iii) ensures that the IV is not confounded and is independent of the noise.
Combining (i) and (iii) we see that the IV is also outcome-independent, i.e., $Z\indep Y\given (A,U,X)$.
We remark that assumptions (i)-(iii) are standard in the IV literature \citep{baiocchi2014instrumental, newey2003instrumental, singh2019kernel, chen2016methods, chen2011nonlinear}, and an IV satisfying these assumptions is referred to as a valid IV. 
Assumption (iv) shows that the missingness is unconfounded and outcome-independent, since $R_X$ and $R_Z$ are neither caused nor have a direct effect on $U$ and $Y$.
We remark that the missingness issue cannot be addressed trivially by only using the dataset subject to $(R_X, R_Z)=\ind$. Note that we have $(U, Z)\rightarrow A$ and $A\rightarrow R_X$ in Figure \ref{fig:IV behavior}. Conditioning on $R_X=\ind$ will therefore create an edge between $Z$ and $U$ and break the IV independence assumption.
Fortunately, we have the following theorem to identify the CATE in the CCB-IV model with missingness.
\begin{theorem}[IES for CATE identification in CCB-IV]\label{thm:IV identification}
Suppose that Assumption \ref{asp:CCB-IV} holds.
If there exist functions $h_1:\cY\times\cA\times\cZ\rightarrow \RR$ and $g:\cA\times\cX\rightarrow \RR$ satisfying,
\begin{align}
    &\EEob\sbr{h_1(Y, A, Z) - Y\given Z=z, R_Z=1} = 0, \label{eq:IV bridge 1}\\
    &\EEob\sbr{g(X, A) - h_1(Y, A, Z)\given A=a, X=x, Z=z, (R_Z, R_X)=\ind}=0 \label{eq:IV bridge 2}, 
\end{align}
it follows that $g(x, a)\overset{\text{a.s.}}{=}\gstar(x, a)$ where $\gstar(x, a)$ is the CATE.
\begin{proof}
See \S\ref{pro:IV identification} for a detailed proof.
\end{proof}
\end{theorem}
\paragraph{An understanding of Theorem \ref{thm:IV identification} in the matrix form.}
We give a matrix explanation of the method we use to overcome the confounding and missingness issue in Theorem \ref{thm:IV identification}. 
We first study what happens if there is no missingness issue but just confounding effect.
Suppose $R_Z\equiv R_X\equiv 1$, a simple Combination of \eqref{eq:IV bridge 1} and \eqref{eq:IV bridge 2} gives the following identification equation, 
\begin{align}
    \EEob\sbr{Y\given Z=z}
    =\EEob\sbr{g^*(A, X)\given Z=z},\label{eq:IV standard id}
\end{align}
which corresponds to the standard identification in the IV model \citep[Proposition 3.1]{liao2021instrumental}. Since \eqref{eq:IV standard id} is learnable from the dataset without missingness, we can thus overcome the confounding issue and recover the CATE with the distributional information encoded in the side observation $Z$.

For the missingness issue, we remark that additionally conditioning on $R_Z=1$ on both sides of \eqref{eq:IV standard id} still recovers the exact CATE, i.e., 
\begin{align}\label{eq:iv matrix id with R_Z}
\EEob\sbr{Y\given Z=z, R_Z=1}
    =\EEob\sbr{g^*(A, X)\given Z=z, R_Z=1},
\end{align}
which is proved in \S\ref{pro:IV identification}.
Thus, we just need to focus on the missingness in $X$.
The difficulty is that we cannot simply evaluate the right-hand side of \eqref{eq:IV standard id} based on the observed data since $\pob(x, a\given z, R_Z=1)\neq \pob(x, a\given z, (R_X, R_Z)=\ind)$. 
To address the problem of missingness in $X$, we have the following observation
\begin{align}
    P(Y, a\given z, R_Z=1) = P(Y\given X, a, z, (R_Z, R_X)=\ind) \cdot  P(X, a\given z, R_Z=1), \label{eq:IV chain rule}
\end{align}
where $P(Y\given X)=\{\pob(y_i\given x_j)\}_{ij}$ denotes a matrix of size $|\cY|\times |\cX|$ whose element in row $i$ and column $j$ is $\pob(y_i\given x_j)$. In the following, we use capital $P$ to denote the matrix formed by matrixing the mass function. 
We remark that \eqref{eq:IV chain rule} is a direct result following the chain rule and the fact that $R_X$ is outcome-independent, i.e., $R_X\indep (Y, R_Z)\given (A, Z, X)$.
By assuming $\text{rank}(P(Y\given X, a, z, (R_X, R_Z)=\ind))=|\cX|$, the Moore-Penrose inverse exists and we have $P(X, a\given z, R_Z=1)=P(Y\given X, a, z, (R_Z, R_X)\allowbreak =\ind)^\dagger P(Y, a\given z, R_Z=1)$.
Now we can rewrite \eqref{eq:iv matrix id with R_Z} as
\begin{align}
    \EEob\sbr{Y\given Z=z, R_Z=1} &=\sum_{a\in\cA}\underbrace{g(a, X)P(Y\given X, a, z, (R_X, R_Z)=\ind)^\dagger}_{\displaystyle{=: h_1(Y, a, z)}} P(Y, a\given z, R_Z=1),\label{eq:IV matrix id}
\end{align}
where $g(X, a)=\{g(x_i, a)\}_{i}$ and $P(Y, a\given z, R_Z=1)=\{p(y_i, a\given z, R_Z=1)\}_i$ are column vectors and $h_1(Y, a, z)=\{h_1(y_i, a, z)_i\}$ is a column vector defined by the under-brace in \eqref{eq:IV matrix id}.
The benefit of introducing $h_1$ is that $h_1$ can be directly learned from the observed dataset.
As a matter of fact, the definition of $h_1$ leads to \eqref{eq:IV bridge 2} and plugging $h_1$ into \eqref{eq:IV matrix id} gives \eqref{eq:IV bridge 1}.
For the bridge function $h_1$ to exist, we need $\text{rank}(P(Y\given X, a, z))=|\cX|$, which implies that the conditional distribution of $Y$ should be informative enough to recover the missing distribution of $X$.
So to overcome the missingness issue, we additionally exploit the \emph{distributional information} of the outcome rather than merely using the \emph{average}. 
In the continuous setting, the equivalent condition for $h_1$ to exist can be expressed as follows.
\begin{remark}[Condition for the existence of a solution to the IES in Theorem \ref{thm:IV identification}]\label{rmk:IV existence}
A solution $h=(h_1, g)$ to the IES in Theorem \ref{thm:IV identification} exists if and only if there exists a solution $h_1$ to the following equation, 
\begin{align}
    \EEob\sbr{\gstar(A, X)-h_1(Y, A, Z)\given A=a, X=x, Z=z, R_Z=1}=0, \quad \forall (a, x, z)\in \cA\times\cX\times\cZ, \label{cond:iv bridge exist}
\end{align}
where $\gstar$ is the exact CATE. We leave the proof for \S\ref{pro:rmk IV existence}.
\end{remark}
Such a condition corresponds to the study of the linear inverse problem. It follows from Picard's theorem that certain completeness conditions are required \citep[Theorem 2.41]{miao2018identifying, carrasco2007linear}. See \S\ref{app:linear inverse} for more details.

\paragraph{Integral equation system.} 
In the following, we assume that the condition in Remark \ref{rmk:IV existence} holds. 
We remark that \eqref{eq:IV bridge 1} and \eqref{eq:IV bridge 2} form an integral equation system (IES), meaning that separately solving \eqref{eq:IV bridge 1} or \eqref{eq:IV bridge 2} alone would not  give the correct answer. The reason is that not all $h_1$ satisfying \eqref{eq:IV bridge 1} respect \eqref{eq:IV bridge 2}.
To illustrate the point, let us consider a special tabular case. Suppose $\epsilon=0$ and $y=f(x, a)$ where $f$ is invertible with respect to $x$ for any fixed $a$. We thus have $g(x, a)=f(x,a)$ as the CATE and the independent condition $Y\indep Z\given (X, A)$. It follows from \eqref{eq:IV bridge 2} that
\begin{align*}
   h_1(Y, a, z)\given_{Y=y}=g(a, X)P(Y\given X, a, (R_X, R_Z)=\ind)^{-1}\given_{Y=y} = g(a, f^{-1}(y, a))=y,
\end{align*}
is the \textit{unique} solution for $h_1$.
However, solving \eqref{eq:IV bridge 1} alone might give the following solution,
\begin{align*}
    h_1(y, a, z)=\EEob\sbr{Y\given Z=z, R_Z=1}.
\end{align*}
Apparently, such a solution does not respect the solution $h_1(y, a, z)=y$ given by \eqref{eq:IV bridge 2}.
Therefore, we see that \eqref{eq:IV bridge 1} and \eqref{eq:IV bridge 2} are coupled together. What matters about the IES is that we have to construct the confidence set for $h_1$ and $g$ as a whole instead of using a nested structure. Details for quantifying the uncertainty that arises from the solving the IES empirically are defered to \S\ref{sec:Estimation}.

\subsection{Identification for Confounded Contextual Bandit with Proximal Variable}

\begin{figure}[h]  
\centering 
  \begin{subfigure}[b]{0.4\linewidth}
  \centering
    \begin{tikzpicture}
        \node[hiddenstate] (U) {$U$};
        \node[missingstate] (X) [below=of U] {$X$};
        \node[state] (A) [left=of X, xshift=.27cm, yshift=-1cm] {$A$};
        \node[state] (Y) [right=of X, xshift=-.27cm, yshift=-1cm] {$Y$};
        
        \path[double] (U) edge (X);
        \path[blue, normal] (U) edge (A);
        \path[blue, normal] (X) edge (A);
        \path[normal] (A) edge (Y);
        \path[normal] (U) edge (Y);
        \path[normal] (X) edge (Y);

        \node[missingstate] (Z) [above=of A] {$Z$};
        \node[missingstate] (W) [above=of Y] {$W$};
        \path[normal] (A) edge (Z);
        \path[dashed, double] (X) edge (Z);
        \path[dashed, normal] (X) edge (W);
        \path[double] (U) edge (Z);
        \path[normal] (U) edge (W);
        \path[normal] (W) edge (Y);
    \end{tikzpicture}
    \caption{DAG of the observational process in CCB-PV}\label{fig:PV behavior}
  \end{subfigure}\qquad
\begin{subfigure}[b]{0.4\linewidth}
\centering
  \begin{tikzpicture}  
        \node[hiddenstate] (U) {$U$};
        \node[missingstate] (X) [below=of U] {$X$};
        \node[state] (A) [left=of X, xshift=.27cm, yshift=-1cm] {$A$};
        \node[state] (Y) [right=of X, xshift=-.27cm, yshift=-1cm] {$Y$};
        
        \path[double] (U) edge (X);
        \path[red, normal] (X) edge (A);
        \path[normal] (A) edge (Y);
        \path[normal] (U) edge (Y);
        \path[normal] (X) edge (Y);

        \node[hiddenstate] (Z) [above=of A] {$Z$};
        \node[hiddenstate] (W) [above=of Y] {$W$};
        \path[normal] (A) edge (Z);
        \path[dashed, double] (X) edge (Z);
        \path[dashed, normal] (X) edge (W);
        \path[double] (U) edge (Z);
        \path[normal] (U) edge (W);
        \path[normal] (W) edge (Y);
\end{tikzpicture}
\caption{DAG of the interventional process in CCB-PV}\label{fig:PV evaluation}
\end{subfigure}
\caption{A DAG illustrating the introduction of side observations ($O=(Z, W)$) in CCB-PV and the difference between the observational model and the interventional model in CCB-PV. Here,  the white nodes represent observed variables, the light grey nodes represent the variables with missingness not at random and the dark nodes represent unmeasured variables. A dashed line between two variables means they can either have explicit causal effect or not. A line with arrows at both ends means that the causal effect going in each way is allowed and therefore, the direction is not specified. }\label{fig:CCB-PV}
\end{figure}
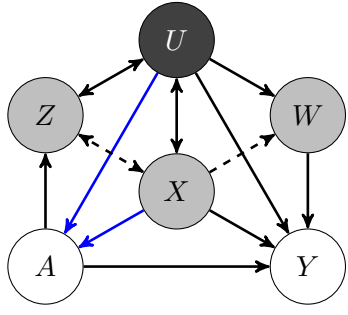
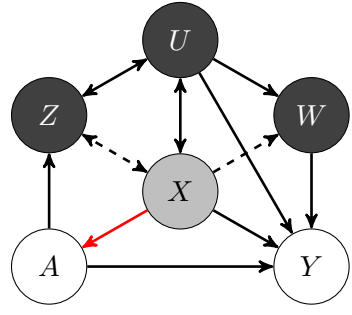

The idea behind CCB-PV is to identify the causal effect using two auxiliary side observations $Z$ and $W$ as negative controls to check for spurious relationships in the existence of unobserved confounder \citep{singh2020kernel, miao2018confounding}.
The model is depicted in Figure \ref{fig:CCB-PV}. 
We present the model assumptions as follows.

\begin{assumption}[Model assumptions for the observational process in CCB-PV]\label{asp:CCB-PV}
We assume that for the CCB-PV with outcome independent missingness, the following assumptions hold for the observational process,
\begin{itemize}
    \item[(i)] (PV completeness). 
    For any $a\in\cA, x\in\cX$,  $\EEob\sbr{\sigma(U)\given X=x, A=a, Z=z, R_Z=1}=0$ holds for any $z\in\cZ$ if and only if $\sigma(u)\overset{\text{a.s.}}{=} 0$ holds.
    \item[(ii)] (PV independence). $W\indep A\given (U, X)$ and $Z\indep (Y, W)\given (A, X,U)$.
    \item[(iii)] (Unconfounded and outcome-independent missingness). We assume that $R_W$ is caused by $(W, X, A)$, $R_X$ is caused by $X$, and $R_Z$ is caused by $(Z, U, A, X)$.
\end{itemize}
\end{assumption}
We remark that (i) and (ii) in Assumption \ref{asp:CCB-PV} are standard for identification in the PV setting \citep{miao2018confounding, miao2018identifying, nair2021spectral, cui2020semiparametric, bennett2021proximal}. Here, (iii) is the unconfounded and outcome-independent missingness assumption, with an exception that $Z$ is allowed to be confounded. 
We remark that the missingness issue cannot be addressed trivially by conditioning on $(R_Z, R_X, R_W)=\ind$, since the PV independence assumption $W\indep A\given (U, X)$ no longer holds when conditioning on $R_W=1$. Moreover, conditioning on $R_Z=1$ also yields a distribution shift in $U$, rendering a bias in identifying the CATE. 
An example of POMDP with such a missingness mechanism is given in \S\ref{sec:POMDP}. 
Now, we provide the identification formula for the CCB-PV as follows.
\begin{theorem}[IES for CATE identification in CCB-PV]\label{thm:PV ID}
Suppose Assumption \ref{asp:CCB-PV} holds.
If there exist bridge functions $h_1:\cY\times\cA\times\cX\times\cZ\rightarrow\RR$, $h_2:\cA\times\cW\times\cX\rightarrow \RR$, $h_3:\cY\times\cA\times\cX\times\cA\rightarrow\RR$ and $g:\cX\times\cA\rightarrow \RR$ satisfying,
\begin{align}
    &\EEob\sbr{h_1(Y, A, X, Z) - Y\given A=a, X=x, Z=z, (R_X, R_Z)=\ind} = 0,\label{eq:PV ID 1}\\
    &\EEob\sbr{h_2(A, W, X) - h_1(Y, A, X, Z)\given (A, W, X, Z)=(a, w, x, z), (R_W, R_X, R_Z)=\ind}=0, \label{eq:PV ID 2}\\
    &\EEob\sbr{h_3(Y, A, X, a')-h_2(a', W, X)\given A=a, W=w, X=x, (R_W, R_X)=\ind}=0 \label{eq:PV ID 3}, \\
    &\EEob\sbr{g(X, a') - h_3(Y, A, X, a')\given X=x, R_X=1}=0, \label{eq:PV ID 4}
\end{align}
for any $(x, a, z, w, a')\in\cX\times\cA\times\cZ\times\cW\times\cA$,
it follows that $g(x, a)\overset{\text{a.s.}}{=}\gstar(x, a)$ where $\gstar(x, a)$ is the CATE.
\begin{proof}
See \S\ref{pro:PV ID} for a detailed proof. 
\end{proof}
\end{theorem}
The matrix explanation for the CCB-PV is similar to the CCB-IV case. See \S\ref{app:PV details} for more details. We give the following conditions on the existence of a solution to the above-mentioned IES.
\begin{remark}[Conditions for existence of a solution to the IES in Theorem \ref{thm:PV ID}]\label{rmk:PV existence}
A solution $h=(h_1, h_2, h_3, g)$ to the IES in Theorem \ref{thm:PV ID} exists if and only if the following three conditions are satisfied:
\begin{itemize}
    \item[(i)] There exists a solution $h_2$ to $\EEob\sbr{h_2(A, W, X)-Y\given A=a, X=x, U=u}=0$;
    \item[(ii)] For any solution $h_2$ in (i), there exists a solution $h_1$ to \eqref{eq:PV ID 2} and a solution $h_3$ to \eqref{eq:PV ID 3}.
\end{itemize}
The proof is given in \S\ref{pro:rmk PV existence}.
\end{remark}
Conditions (i)-(ii) also require certain property of completeness, as is discussed in the CCB-IV case. See \S\ref{app:linear inverse} for more details.
Note that (i) is a standard condition in proximal causal learning \citep{miao2018identifying, miao2018confounding, cui2020semiparametric}. In a discrete setting, condition (i) means $\text{rank}(P(W\given U, x, a))=|U|$, showing that $W$ should couple enough information from the unmeasured confounder.
For $h_1$ and $h_3$ to exist, we just need $\text{rank}(P(Y\given W, a, x, z))=|\cW|$ and $\text{rank}(P(Y\given W, a, x))=|\cW|$, implying that $Y$ couples sufficient information of the missing variable $W$. 
Analogue to the CCB-IV case, we remark that equations in the IES for CCB-PV are coupled and should be solved at the same time. 


\paragraph{Pseudo random variable $A'$.}
Here in the CCB-PV setting, we encounter an issue concerning $a'$ that appears in \eqref{eq:PV ID 3} and \eqref{eq:PV ID 4}. Note that \eqref{eq:PV ID 3} and \eqref{eq:PV ID 4} should hold point-wise with respect to $a'$. However, if we treat each $a'\in\cA$ separately, we need to solve $\abr{\cA}$ equations, and the problem becomes intractable as $|\cA|$ grows larger, or we have a continuous treatment space. To overcome such a difficulty, we propose to treat $a'$ as a realization of a pseudo random variable $A'$, which is independent of the CCB-PV model and uniformly distributed across the action space. 
Therefore, the joint distribution $\mu$ including $A'$ is given by
\begin{align*}
    \mu(y, a, w, x, z, u, r_W, r_X, r_Z, a')=\pob(y, a, w, x, z, u, r_W, r_X, r_Z) u(a').
\end{align*}
Thereby, \eqref{eq:PV ID 3} and \eqref{eq:PV ID 4} can be written as, 
\begin{align}
    &\EEob\sbr{h_3(Y, A, X,  A')-h_2(A', W, X)\given A=a, W=w, X=x, A'=a', (R_W, R_X)=\ind}=0 \label{eq:PV ID 5}, \\
    &\EEob\sbr{h_4(X, A') - h_3(Y, A, X, A')\given X=x, A'=a', R_X=1}=0. \label{eq:PV ID 6}
\end{align}
Here, by letting $A'\sim u(a)$, we try to learn an estimator that yields small error in each $a'$. Moreover, the dataset $\cD$ can be easily adjusted by adding an element $a_t'$ which is uniformly selected from $\cA$ and added to each sample, i.e., $\cD=\{\check x_t, a_t, y_t, \check w_t, \check z_t, a'_t\}_{t=1}^T$. By treating $A'$ as a pseudo random variable and $a'$ as its realization, we transform \eqref{eq:PV ID 3} and \eqref{eq:PV ID 4} into conditional moment equations, which facilitates our analysis of the IES in the sequel. 

\subsection{United Form for the IES}\label{sec:united form}
We summarize the IES discussed in both the CCB-IV and CCB-PV cases into the following united form
\begin{align}\label{eq:united form}
    \EE_{\mu_k}[\alpha_k(\vh(\vX), \vY_k)\given \vZ_k]=0, \quad \forall k\in\cbr{1, \cdots, K}, 
\end{align}
where 
\begin{description}
    \item[$K$] denotes the total number of equations in the IES,
    \item[$\vh$] represents the vector of bridge functions to learn, i.e., $ \vh(\vX)= (h_1(\vX_1),\cdots, h_{K-1}(\vX_{K-1}), g(\vX_K))$ where $X_k$ is the random variable vector that $h_k$ depends on and $X$ is a union of $\{X_k\}_{k=1}^K$, 
    \item[$\alpha_k$] is the linear function that is taken expectation with in the $k$-th equation of the IES, which depends on $h(X)$ and random variable vector $Y_k$, 
    \item[$\vZ_k$] is the random variable vector that is conditioned on in the $k$-th equation of the IES,
    \item[$\mu_k$] denotes the joint distribution for $(\vX, \vY_k, \vZ_k)$ and $\EE_{\mu_k}$ is the expectation taken with respect to $\mu_k$, 
\end{description}
We give an example to illustrate the united form in \eqref{eq:united form}.
Following the IES of CCB-PV in Theorem \ref{thm:PV ID}, we have $K=4$, $X=\bigcup_{k=1}^K \vX_k = \{Y, A, X, Z, W, A'\}$, $Y_1=Y$, $Z_1=\{A, X, Z\}$, $\mu_1(x, a, y, z)=\pob(x, a, y, z\given (R_X, R_Z)=\ind)$, and $\alpha_1(\vh(\vX), \vY_1)=h_1(Y, A, X, Z)-Y$.  Note that the joint distribution $\mu_k$ for the variables in each equation can be different. For example, the joint distribution for \eqref{eq:PV ID 1} is conditioned on $(R_X, R_Z)=\ind$ while the joint distribution for \eqref{eq:PV ID 2} is conditioned on $(R_W, R_X, R_Z)=\ind$. 
Moreover, let $\cD_k$ denote the subset data of $\cD$ corresponding to the missingness condition in the $k$-th equation of the IES, e.g., in the CCB-PV case we have $\cD_1=\{(\check x_t, a_t, y_t, \check w_t, \check z_t, a'_t): (r_{X, t}, r_{Z, t})=\ind\}$, which is a subset of $\cD=\{(\check x_t, a_t, y_t, \check w_t, \check z_t, a'_t)\}_{t=1}^T$ and $\cD_1\sim\mu_1$.
Let $\cF(\vZ_k)$ denotes the functional space on $\vZ_k$.
Since $\alpha_k$ is linear, we can define a linear operator $\cT_k: \vH\rightarrow \cF(\vZ_k)$ for the united form of the IES in \eqref{eq:united form} as
\begin{align}
    \cT_k \vh (\cdot) :=\EE_{\mu_k}\sbr{\alpha_k(\vh(X), \vY_k)\given \vZ_k=\cdot}. \label{def:cT}
\end{align}
By letting $\vT\vh(z)=(\cT_1 h (z_1), \cdots, \cT_K h(z_K))$, the IES in \eqref{eq:united form} can be alternatively expressed as 
\begin{align}
    \vT\vh(z) = 0.\label{eq:operator=0}
\end{align}
Note that the IES \eqref{eq:operator=0} comprises a series of conditional moment equations, which is hard to solve with offline data. In the next section, we propose to transform the conditional moment equations into unconditional moment estimators.  

\section{Estimation}\label{sec:Estimation}
In this section, we use the method of minimax estimation to transform the conditional moment restrictions in the IES into an unconditional moment minimax estimator with respect to $\cL_\cD(\cdot)$.
We then build the confidence set for the CATE as the level set in $\vH$ under metric $\cL_\cD(\cdot)$ and threshold $e_\cD$. Based on the confidence set, we integrate pessimism in policy optimization.
\paragraph{RMSE and unconditional moment criteria.}
We let $\vH=\cH_1\times\cdots\times\cH_{K-1}\times\cG$ be the hypothesis space for $\vh$.
Following the idea of projected residual mean squared error (RMSE) minimization \citep{dikkala2020minimax}, our estimation target is good generalization performance subject to the following RMSE, 
\begin{align}\label{def:RMSE}
    \nbr{\vT\vh}^2_{\mu, 2}:= \sum_{k=1}^K \EE_{\mu_k}\sbr{\cT_k \vh(\vZ_k)}^2 = \sum_{k=1}^K \EE_{\mu_k}\sbr{\rbr{\EE_{\mu_k}\sbr{\alpha_k(\vh(X), \vY_k)\given \vZ_k}}^2}, 
\end{align}
where $\vT\vh=(\cT_1\vh, \cdots, \cT_K\vh)$ and $\mu=(\mu_1, \cdots, \mu_K)$. Note that solving the conditional moment equations \eqref{eq:operator=0} corresponds to finding the $\vh\in\vH$ that minimizes the RMSE.
However, learning the causal relationship with conditional moment restrictions is a challenging task. It has been investigated in the existing literature how to transform conditional moment conditions into unconditional moment conditions, e.g., methods of importance weighting using conditional density ratio \citep{kato2021learning} or using linear sieve estimator \citep{ai2003efficient} where the estimator has a rate of $n^{-1/4}$.
Inspired by the method of minimax estimation \citep{dikkala2020minimax, duan2021risk, uehara2021finite} with fast rate of $n^{-1/2}$, we propose to approximate \eqref{def:RMSE} with an unconditional minimax estimator. 
Specifically, we introduce a test function $\theta_k:\vZ_k\rightarrow \RR$ for each linear operator $T_k$. The function class for $\theta_k$ is $\Theta_k$, which we refer to as the dual function class.
The unconditional moment loss function that is used to replace the RMSE in \eqref{def:RMSE} is then given by
\begin{align}\label{def:L}
    \cL(\vh) = \sum_{k=1}^K \cL_k(\vh), \text{ where } \cL_k(\vh)=\sup_{\theta_k\in\Theta_k}\EE_{\mu_k}\sbr{\alpha_k(\vh(X), \vY_k) \theta_k(\vZ_k)}-\frac 1 2 \norm{\theta_k}_{\mu_k, 2}^2.
\end{align}
By assuming the test function class $\Theta_k$ to be star-shaped, we always have $\cL_k(\vh)\ge 0$ (otherwise by letting $\theta_k=0$ there is a conflict with the fact that $\cL_k(\vh)$ takes the supremum over $\Theta_k$).
Note that as long as $\cT_k\vh\in\Theta_k$, the loss function $\cL(\cdot)$ is equivalent to the RMSE $\nbr{\cT(\cdot)}^2_{\mu, 2}$, which can be verified by the property of Fenchel duality (see \S\ref{lem:Fenchel} for a detailed proof). Therefore, if we have $\cT_k\vh\in\Theta_k$, we see that any solution to \eqref{eq:operator=0} is a minimizer to the unconditional moment loss function, i.e., $\vhstar=\arg\inf_{\vh} \cL(\vh)$.
In line with \eqref{def:L}, 
we define the empirical loss function on the dataset $\cD$ as follows,
\begin{align*}
    \cL_\cD(\vh) = \sum_{k=1}^K \cL_{k, \cD}(\vh), \text{ where } \cL_{k,\cD}(\vh)=\sup_{\theta_k\in\Theta_k}\EE_{\cD_k}\sbr{\alpha_k(\vh(X), \vY_k) \theta_k(\vZ_k)}-\frac 1 2 \norm{\theta_k}_{\cD_k, 2}^2, 
\end{align*}
where $\cD_k$ is a subset of $\cD$ and we have $\cD_k\sim \mu_k$, as is discussed in \S\ref{sec:united form}.
We also have $\cL_{k, \cD}(\vh)\ge 0$ following the same argument that $\cL_{k, \cD}$ takes the supremum over the dual function class.
Now we are ready to build the confidence set $\CIH$ as the level set with respect to the metric $\cL_\cD(\cdot)$ and threshold $e_\cD$ as follows,
\begin{align}\label{def:CI}
    \CIH(e_\cD) = \cbr{\vh\in\vH:  \vcL_{\cD}(\vh) \le  \inf_{\vh\in\vH}\vcL_{\cD}(\vh) + e_\cD}.
\end{align}
Correspondingly, the confidence set for the CATE is given by
\begin{align*}
    \CICATE(e_\cD) = \cbr{g\in\cG: \exists\vh\in\CIH(e_\cD), \text{ s.t., } g=\vh^{(K)}}, 
\end{align*}
where $\vh^{(K)}$ is the last element of $\vh$. 
Therefore, we have $g=h^{(K)}$ by noting that the last element of $\vh$ is the estimated CATE following both Theorems \ref{thm:IV identification} and \ref{thm:PV ID}.
We remark that building the confidence set is a way to address the aleatoric uncertainty stemming from the data generating process, as will be shown in Theorem \ref{thm:Fast rate} that $\CIH(e_\cD)$ can capture an estimator $\vhHstar$ with small realizability error. On the other hand, we can eliminate the spurious correlation in \eqref{eq:spurious subopt} with pessimism on such a confidence set, i.e., greedily selecting the policy that optimizes the pessimistic average reward function with $g\in\CICATE(e_\cD)$,
\begin{align*}
    \piepessi = \arg\sup_{\pie\in\Pie} \inf_{g\in\CICATE(e_\cD)} v(g, \pie),  
\end{align*}
where $v$ is the average reward function defined in \eqref{def:v approx}.
Here, we denote by $\gpessi{\pie}=\arg\inf_{g\in\CICATE(e_\cD)}v(g, \pie)$ the estimated pessimistic CATE under the interventional policy $\pie$.  
By plugging in the definition of $\cL_\cD(\cdot)$ and $\CICATE(\cdot)$ in  Algorithm \ref{alg:meta}, we obtain the complete Causal-Adjusted Pessimistic (CAP) policy learning algorithm.

\section{Theoretical Results} \label{sec: theoretical results}
Let $\vhstar$ denote the exact solution to \eqref{eq:united form}.
We allow the model to be misspecified, i.e., the exact solution might not be fully captured by the hypothesis space $\vH$.
To characterize the approximation error, we pose the following assumption on the realizability of the hypothesis class $\vH$.
\begin{assumption}[Realizability of hypothesis class]\label{asp:Realizability}
Let $\vareH>0$ be the minimal positive value such that there exists $\vhHstar=\{\hHstark{1}, \cdots, \hHstark{K-1}, \gHstar\}\in\vH$ satisfying,
\begin{itemize}
    \item[(i)] $\nbr{\vT\vhHstar}_{ \mu, 2} \le \vareH$, where $\nbr{\vT\vhHstar}_{\mu, 2}$  is the RMSE defined in \eqref{def:RMSE}.
    \item[(ii)] $\sup_{v\in\cV} \nbr{\gHstar-\gstar}_{v, 2}\le \vareH$, where $\cV=\{v: v(x, a)=\tpr(x)\pie(a\given x), \forall \pie\in\Pie\}$.
\end{itemize}
\end{assumption}
Here, (i) characterizes the approximation error of $\vhHstar$ under the metric $\nbr{\cT(\cdot)}_{\mu, 2}$ where $\mu$ represents the distribution in the dataset, and the approximation error in (ii) is the supremum over all the possible measure that is realizable by the policy class $\Pie$.
We remark that $\vhHstar$ can be softly viewed as the projection of $\vhstar$ onto the hypothesis space $\vH$ with approximation error $\vareH$.
We then pose the following assumption on the compatibility of the test function class $\Theta_k$.
\begin{assumption}[Compatibility of test function class]\label{asp:compatibility}
Suppose that for any $\vh\in\vH$ and for any $k\in\{1, \cdots, K\}$, it holds that
$\inf_{\theta_k\in\Theta_k} \nbr{\theta_k - \cT_k\vh}_{\mu_k, 2}\le \vareTheta$.
\end{assumption}
We give an example where the test function class has full compatibility. Following the discussion in \cite{dikkala2020minimax}, we consider a case where $\Theta_k$ lies in a Reproducing Kernel Hilbert space (RKHS) $\HH_{K_{\theta_k}}$ with RKHS kernel $K_{\theta_k}: \cZ_k\times\cZ_k\rightarrow \RR$. If $\alpha_k$ lies in another RKHS space $\HH_{K_{\alpha_k}}$ and the conditional density function $\pr(\vX, \vY_k\given \vZ_k)$ satisfies $\pr(\vX, \vY_k\given \cdot)\in \HH_{K_{\Theta_k}}$, we then have $\cT_k\vh\in \HH_{K_{\Theta_k}}$, which means that the dual function class has full compatibility. In addition, we pose the following assumption on the regularity of function classes $\vH$, $\vTheta$ and linear function $\alpha_k$.
\begin{assumption}[Regularity]\label{asp:regularity}
We assume that $\alpha_k$ is $L_{\alpha, 1}$-Lipschitz continuous with respect to $h_j$ and $Y_k$ for all  $j,k\in\{1, \cdots, K\}$.
We assume that the support of $Y_k$ is bounded, i.e., $\nbr{\text{supp}(Y_k)}_\infty\le L_Y$.
Moreover, we assume that $\sup_{h\in\vH}\norm{h}_\infty\le L_{h}$ and $\sup_{\theta\in\vTheta}\norm{\theta}_\infty\le L_{\theta}$.
\end{assumption}
We justify the regularity assumption by the examples of CCB-IV and CCB-PV.
Following Theorems \ref{thm:IV identification} and \ref{thm:PV ID}, 
we see that the continuity of $\alpha_k$ is apparent.
The regularity of $\vH$ and $\vTheta$ is easy to satisfy by choosing bounded function classes.
With bounded reward, it is straightforward that the linear function $\alpha_k(\vh, \vY_k)$ is globally bounded. Specifically, we can assume that  $\nbr{\alpha_k}_\infty\le L_\alpha$.

To characterize the properties of the confidence set Under these assumptions, we first define an event $\cE$ as
\begin{align}
    \cE &= \Big\{\abr{\EE_{\cD_k}\sbr{\alpha_k(\vh, \cY_k) \theta_k(\cZ_k)} - \EE_{\mu_k}\sbr{\alpha_k(\vh, \cY_k) \theta_k(\cZ_k)}} \le \eta_k\rbr{L_\alpha\nbr{\theta_k}_{\mu_k, 2}+\eta_k},\nend
    &\quad \abr{\norm{\theta_k}^2_{\cD_k, 2}-\norm{\theta_k}^2_{\mu_k, 2}}\le \frac 1 2\rbr{\norm{\theta_k}^2_{\mu_k, 2}+\eta_k^2}, 
    \forall \vh\in \vH, \forall \theta_k\in\Theta_k, \forall k\in\{1, \cdots, K\}\Big\}.\label{def:cE}
\end{align}
where $\eta_k$ bounds the maximal critic radius for function classes $\cQ_k$ and $\Theta_k$ with respect to $\xi\in(0, 1)$. Here, we define function class $\cQ_k$ as
\begin{align*}
    \cQ_k=\cbr{\alpha_k(\vh(\vX_k), \vY_k)\theta_k: \forall \vh\in\vH, \theta_k\in\Theta_k}.
\end{align*}
See \S\ref{app:critical radius} for calculation of the critical radius in the case of linear function class. 
We let $\eta^2=\sum_{k=1}^K \eta_k^2$ for simplicity. Now we give the following theorem, which shows that event $\cE$ holds with high probability and the confidence set built for uncertainty quantification enjoys some good properties.
\begin{theorem}[Uncertainty Quantification]\label{thm:Fast rate}
Suppose that Assumptions \ref{asp:Realizability}, \ref{asp:compatibility} and \ref{asp:regularity}  hold. 
Event $\cE$ holds with probability at least $1-2K\xi$ and the confidence set enjoys the following properties on $\cE$, 
\begin{itemize}
    \item[(i).] For the $\vhHstar$ satisfying Assumption \ref{asp:Realizability}, it holds that $\cL_\cD(\vhHstar) \le 2\vareH^2 + \rbr{2 L_\alpha^2+5 /4}\eta^2$. Moreover, if we set $e_\cD>2\vareH^2 + \rbr{2 L_\alpha^2+ 5/ 4}\eta^2$, it holds that $\gHstar\in\CICATE(e_\cD)$.
    \item[(ii).] For all $\vh\in\CIH(e_\cD)$, we have,
\begin{align*}
    \sup_{k\in\{1, \cdots, K\}}\nbr{\cT_k\vh}_{\mu_k, 2}\overset{\cE}{\lesssim} \cO(\vareTheta) + \cO(\vareH) + \cO\rbr{\sqrt{e_\cD}} +  \cO\rbr{\eta}.
\end{align*}
\end{itemize}
\begin{proof}
See \S\ref{pro:Fast rate} for a detailed proof.
\end{proof}
\end{theorem}
Theorem \ref{thm:Fast rate} shows that event $\cE$ holds with a high probability.
We see from (i) that it is theoretically guaranteed that $\vhHstar$ lies within the confidence set by properly setting the threshold $e_\cD$. As will be shown shortly after, such a property is vital for the use of pessimism.
Property (ii) in Theorem \ref{thm:Fast rate} shows that the RMSE for any $\vh\in\CIH(e_\cD)$ is well controlled on event $\cE$.
Now we are ready to present the convergence results for the sub-optimality defined in \eqref{def: SubOpt}. We give the following theorem on the sub-optimality for the CCB-IV.
\begin{theorem}[Convergence of sub-optimality for CCB-IV]\label{thm:IV subopt}
Suppose that Assumption 
\ref{asp:CCB-IV} for the CCB-IV model and the conditions in Remark \ref{rmk:IV existence} hold.
Suppose that Assumptions
\ref{asp:Realizability},  \ref{asp:compatibility}, and 
\ref{asp:regularity} for function classes $\vH$ and $\Theta$ hold.
The threshold $e_\cD$ for the confidence set is set to $e_{\cD}>2\vareH^2 + (2 L_\alpha^2+5/ 4)\eta^2$.
For the marginal distribution of context $\tpr$ in the interventional process and the optimal interventional policy $\piestar$, suppose that there exists $b_1:\cZ\rightarrow \RR$ satisfying 
\begin{align}
    \EEob\sbr{b_1(Z)\given A=a, X=x, R_Z=1} = \frac{\tpr(x)\piestar(a\given x)}{\pob(x, a\given R_Z=1)}.\label{cond:CCB-IV b1 exist}
\end{align}
The sub-optimality corresponding to $\piepessi$ for the CCB-IV is bounded on event $\cE$ with probability at least $1-4\xi$ by
\begin{align*}
    \SubOpt(\piepessi) \overset{\cE}{\lesssim} \sum_{k=1}^2 \nbr{b_k}_{\mu_k, 2}\cdot \big(\cO(\vareTheta) + \cO(\vareH) + \cO\rbr{\sqrt{e_\cD}} +  \cO\rbr{\eta}\big),
\end{align*}
where $b_2:\cA\times\cX\times\cZ\rightarrow \RR$ is defined as,
\begin{gather}
    b_2(a, x, z)=b_1(z)\frac{\pob(a, x, z\given R_Z=1)}{\pob(a, x, z\given(R_X, R_Z)=\ind)}.\label{eq:b2 IV}
\end{gather}
\begin{proof}
See \S\ref{proof: pessimism} and \S\ref{proof: subopt of IV} for a detailed proof.
\end{proof}
\end{theorem}
Similarly, the convergence of sub-optimality for CCB-PV is given by
\begin{theorem}[Convergence of sub-optimality for  CCB-PV]\label{thm:PV subopt}
    Suppose that Assumption 
\ref{asp:CCB-PV} for the CCB-PV model and the conditions in Remark \ref{rmk:PV existence} hold.
Suppose that Assumptions
\ref{asp:Realizability}, \ref{asp:compatibility}, and
\ref{asp:regularity} for function classes $\vH$ and $\Theta$ hold.
The threshold $e_\cD$ for the confidence set is set to $e_{\cD}>2\vareH^2 + (2 L_\alpha^2+5/ 4)\eta^2$.
For the marginal context distribution  $\tpr$ in the interventional process and the optimal interventional policy $\piestar$, suppose that there exists $b_1:\cX\times\cA\times\cZ\rightarrow\RR$ satisfying
    \begin{align}
        \EEob\sbr{b_1(X, A, Z)\given X=x, U=u, A=a, R_Z=\ind} = \frac{\pob(u\given x)\tpr(x)\piestar(a\given x)}{\pob(u, x, a\given  R_Z=1)}.\label{cond:CCB-PV b1 exist}
    \end{align}
    The sub-optimality corresponding to $\piepessi$ for the CCB-PV is bounded with probability at least $1-8\xi$ by
    \begin{align*}
        \SubOpt(\piepessi) \lesssim \sum_{k=1}^4 \nbr{b_k}_{\mu_k, 2}\cdot \big(\cO(\vareTheta) + \cO(\vareH) + \cO\rbr{\sqrt{e_\cD}} +  \cO\rbr{\eta}\big),
    \end{align*}
    where $b_2:\cA\times\cW\times\cX\times\cZ$, $b_3:\cW\times\cX\times\cA\times\cA'$ and $b_4:\cX\times\cA'\rightarrow\RR$ are defined as
    \begin{gather}
        b_2(a, w, x, z) = b_1(x, a, z)\frac{\pob(a, w, x\given (R_X, R_Z)=\ind)}{\pob(a, w, x\given (R_W, R_X, R_Z)=\ind)}.\nend
        b_3(w, x, a, a') = \frac{\tpr(x)\piestar(a'\given x)\pob(a, w\given x, R_X=1)}{u(a')\pob(x, a, w\given (R_W, R_X)=\ind)}, \quad
        b_4(x, a') = \frac{\tpr(x)\piestar(a'\given x)}{\pob(x\given R_X=1)u(a')}.\nonumber
    \end{gather}
\begin{proof}
See \S\ref{proof: pessimism} and \S\ref{proof: subopt of PV} for a detailed proof.
\end{proof}
\end{theorem}
\paragraph{Remarks on the existence of $b_1$.} We remark that the existence of $b_1$ in \eqref{cond:CCB-IV b1 exist} and \eqref{cond:CCB-PV b1 exist} are also related to the linear inverse problem,  as is discussed in \S\ref{app:linear inverse}. In the discrete setting, \eqref{cond:CCB-IV b1 exist} is automatically satisfied if the distribution shift ratio on the right-hand side is globally bounded. This is because we already have $\text{rank}(P(Z\given (A,X), R_Z=1))\ge |\cA|\times|\cX|$ by the IV completeness assumption and thus the Moore-Penrose inverse  $P(Z\given (A,X), R_Z=1)^+$ exists. Similarly, \eqref{cond:CCB-PV b1 exist} is also automatically satisfied in the discrete setting following the PV completeness assumption.

\paragraph{Significance of the main theorems.} 
Theorem \ref{thm:IV subopt} and \ref{thm:PV subopt} establish the convergence of the sub-optimality for the CAP algorithm in the offline confounded contextual bandit with missingness. 
Such results are achieved by using the minimax estimator, building a confidence set for the CATE, and integrating pessimism in the policy optimization step.
Our theories deal with model misspecification in the hypothesis space and the dual function class. Moreover, the sub-optimality does not rely on the “uniform coverage” of the observational dataset, e.g., uniformly lower bounded densities of visitation measures \citep{yang2020off, liao2020batch, duan2020minimax}.
This is because the distribution shift ratio $b_k$ only depends on the optimal policy $\piestar$, which is intrinsic to the bandit model, instead of the whole policy class $\Pie$. Therefore, we only require the dataset to “cover” certain distributions induced by $\piestar$.

\paragraph{Implications of the main theorems.} 
When it holds that $\eta\sim\tcO(n^{-1/2})$ (see \S\ref{app:critical radius} for a calculation of the critical radius of the linear function class) and $\vareH=\vareTheta=0$, by setting $e_\cD\sim\tcO(n^{-1})$, Theorem \ref{thm:IV subopt} and \ref{thm:PV subopt} indicates that $\SubOpt(\piepessi)\sim\tcO(n^{-1/2})$, which corresponds to a “fast statistical rate” for minimax estimation \citep{uehara2021finite}.
However, given the fact that the distribution shift ratio $b_k$ also stems from the missingness issue, we require some “compliance” of the data distribution with missingness in order for $b_k$ to be bounded. For instance, we require $\pob(a, x, z\given  R_Z=1)=\cO(\pob(a, x, z\given (R_X, R_Z)=\ind))$ for $b_2$ in \eqref{eq:b2 IV} to be bounded. Such a condition is reasonable if we consider a counterexample where $\pob(x_0\given R_Z)>0$ but $\pob(x_0\given (R_X, R_Z)=\ind)=0$ for $x_0\in \cX$, meaning that $x_0$ is totally missing from the observed dataset $\cD$. Therefore, there is no way to learn the CATE corresponding to $x_0$. 
On the other hand, following the discussion in \S\ref{app:critical radius} on the critical radius of a linear function class, we have $\eta_k\sim\cO(\sqrt{\log T_k/T_k})$ where $T_k=|\cD_k|$. Recall that $\cD_k$ may have different size since the conditional moment equations in the IES are subject to different missingness conditions. 
Therefore, we also require $T_k$ being of the same order of $T$ for $\SubOpt(\piepessi)$ to enjoy a fast statistical rate.

\section{Extended Policy Class for CCB-PV}\label{sec:extended CCB-PV}
\paragraph{Motivation.} 
In the previous discussion, since we assume that the side observations are not accessible in the interventional process, we restrict our interventional policy to the class $\pie:\cX\rightarrow \Delta(\cA)$. 
In this section, we discuss an extension to the setting with accessible side observations in the interventional process.
Note that we have $Z\indep Y\given (X, A)$ in the CCB-IV, meaning that including the side observation $Z$ adds no additional information to the outcome and therefore the policy class $\pie:\cX\rightarrow \Delta(\cA)$ is good enough. 
In the CCB-PV setting, however, it is possible to improve the performance by allowing the interventional policy to also depend on the side observations. 
Specifically, we consider an extension of the CCB-PV setting where the interventional policy class is given by $\pie\in\Pie:\cX\times\cW\rightarrow \Delta(\cA)$. 
We have to redefine the CATE and average reward function for the extended policy class by
\begin{align}
    &\CATE(a, x, w) = \EEin\sbr{Y\given X=x, W=x, \doopt(a)}, \nonumber\\
    &v^\pi(x) = \EE_{\pin{\pie}} \sbr{\CATE(A, X, W)\given X=x},\label{def:extended v}
\end{align}
where $\pin{\pie}$ is given  by plugging  $\pie=(a\given x, w)$ into the joint distribution of the interventional process in \eqref{def:pin}. 
Following the model in Assumption \ref{asp:CCB-PV}, we provide the identification IES as follows.
\begin{theorem}[IES for CCB-PV with extended policy]\label{thm:CCB-PV ID extension} 
Suppose Assumption \ref{asp:CCB-PV} holds. For any interventional policy $\pie:\cX\times\cW\rightarrow\Delta(\cA)$, if there exist bridge functions $h_1:\cY\times\cA\times\cX\times\cZ\rightarrow\RR$, $h_2:\cA\times\cW\times\cX\rightarrow \RR$, $h_3:\cY\times\cA\times\cX\times\cA\rightarrow\RR$ and $g:\cX\rightarrow \RR$ satisfying,
\begin{align}
    &\EEob\sbr{h_1(Y, A, X, Z)\given a,x,z, (R_X, R_Z)=\ind} = 0,
    \nend
    &\EEob[h_2(A, W, X) - h_1(Y, A, X, Z)-Y\pie(A\given X, W)
    \given a, w, x, z, (R_W, R_X, R_Z)=\ind]=0, 
    \nend
    &\EEob\sbr{h_3(Y, A, X)-\sum_{a'\in\cA}h_2(a', W, X)\given a, w, x, (R_W, R_X)=\ind}=0,
    \label{eq:PV ID extension 3} 
    \\
    &\EEob\sbr{g(X) - h_3(Y, A, X)\given x, R_X=1}=0, 
    \label{eq:PV ID extension 4}
\end{align}
it follows that $ v^\pie(x)\aseq g(x)$ where $v^\pie$ is the average reward.
\begin{proof}
See \S\ref{pro:CCB-PV ID extension} for a detailed proof.
\end{proof}
\end{theorem}
\paragraph{Existence of the solution. }
The conditions for existence of a solution to such an IES is similar to Remark \ref{rmk:PV existence}, except that the first condition is adjusted by assuming there exists a solution $h_2$ to $\EEob\sbr{h_2(A, W, X)-Y\pie(A\given W, X)\given A=a, X=x, U=u}$ and the rest two conditions are just the same.
\paragraph{A comparison to Theorem \ref{thm:PV ID}.}
The differences between these two versions of identification formula are in three folds: (i) The identification equations in Theorem \ref{thm:CCB-PV ID extension} are policy specific while those in Theorem \ref{thm:PV ID} hold for any interventional policy; (ii) there is no need for introducing a pseudo variable $A'$ here since $a'$ is already marginalized in \eqref{eq:PV ID extension 3}; (iii) $g$ corresponds to the average reward instead of the CATE.

\paragraph{Algorithm.} 
Note that the linear function $\alpha$, the operator $\cT$ and the loss function $\cL_\cD$ should depend on policy $\pie$ and we denote them by $\alpha^\pie$, $\cT^\pie$ and $\cL_\cD^\pie$, respectively.
The confidence set is built for each $\pie$ by
\begin{align*}
    \CICATE^\pie = \cbr{g\in\cG: \exists \vh\in\vH \text{, s.t., } g=\vh^{(K)} \text{ and }\cL_\cD^\pie(\vh)\le \inf_{\vh\in\vH}\cL_\cD^\pie(\vh)+e_\cD}, 
\end{align*}
Therefore, the estimated policy with pessimism is given by 
\begin{align}
    \piepessi = \arg\sup_{\pie\in\Pie} \inf_{g\in\CICATE^\pie(e_\cD)} v(g), \quad \text{where } v(g) = \int_{\cX} g(x) \tpr(x)\rd x.\label{def:extended piepessi}
\end{align}
Before we give the main theorem, we restate the realizability assumption (Assumption \ref{asp:Realizability}) as follows.
\begin{assumption}[Realizability of hypothesis class for extended CCB-PV]\label{asp:extended realizability}
Let $\vareH>0$ be the minimal positive value such that there exists $\vhHpie=\{\hHpiek{1}, \cdots, \hHpiek{K-1}, \gHpie\}\in\vH$ satisfying,
\begin{itemize}
    \item[(i)] $\sup_{\pie\in\Pie}\nbr{\vT^\pie\vhHpie}_{\mu, 2} \le \vareH$
    \item[(ii)] $\sup_{\pie\in\Pie}\nbr{\gHpie-g^\pie}_{\tpr, 2}\le \vareH$, where $g^\pie$ is the exact solution to the identification equations in Theorem \ref{thm:CCB-PV ID extension}.
\end{itemize}
\end{assumption}
The compatibility assumption can be easily adjusted by assuming that $\inf_{\theta_k\in\Theta_k} \nbr{\theta_k - \cT_k^\pie\vh}_{\mu_k, 2}\le \vareTheta$ for any $\pie\in \Pie$. The regularity assumption remains the same, except that we also assume  $\nbr{\pie}_\infty$ to be bounded in order to have $\alpha^\pie_k$ globally bounded.
Now we provide the following theorem to characterize the convergence of sub-optimality for the CCB-PV with extended interventional policy. 

\begin{theorem}[Convergence of sub-optimality of CCB-PV with extended policy class]\label{thm:extended PV subopt}
Suppose that Assumptions \ref{asp:CCB-PV}, \ref{asp:extended realizability}, \ref{asp:compatibility},
\ref{asp:regularity} hold and the solution to the IES in Theorem \ref{thm:CCB-PV ID extension} exists.
let $e_{\cD}>2\vareH^2 + (2 L_\alpha^2+5/ 4)\eta^2$, where $\eta=\sum_{k=1}^K\eta_k^2$ and $\eta_k$ bounds the maximal critic radius for function classes $\tilde\cQ_k=\cbr{\alpha_k^\pie(\vh(\vX_k), \vY_k)\theta_k: \forall \vh\in\vH, \theta_k\in\Theta_k, \pie\in\Pi}$ and $\Theta_k$. Suppose that for any $x\in\cX, u\in\cU$ and $a\in\cA$ there exists $b_1:\cX\times\cA\times\cZ\rightarrow\RR$ satisfying
\begin{align}
    \EEob\sbr{b_1(X, A, Z)\given x, u, a, R_Z=1} = \frac{\pob(u\given x)\tpr(x)}{\pob(u, x, a\given R_Z=1)}.\label{cond:b1 PV extended}
\end{align}
The sub-optimality corresponding to $\piepessi$ for the CCB-PV with extended policy class is bounded with probability at least $1-8\xi$ by
    \begin{align*}
        \SubOpt(\piepessi) \lesssim \sum_{k=1}^4 \nbr{b_k}_{\mu_k, 2} \cdot \rbr{O(\vareTheta) + O(\vareH) + O\rbr{\sqrt{e_\cD}} +  O\rbr{\eta}}, 
    \end{align*}
    where $b_2:\cA\times\cW\times\cX\times\cZ\rightarrow\RR$, $b_3:\cW\times\cX\times\cA\rightarrow\RR$ and $b_4:\cX\rightarrow\RR$ are defined by
    \begin{align*}
        &b_2(a, w, x, z) = b_1(x, a, z)\frac{\pob(a, w, x\given (R_X, R_Z)=\ind)}{\pob(a, w, x\given (R_W, R_X, R_Z)=\ind)}, \\
        &b_3(w, x, a) = \frac{\tpr(x)\pob(a, w\given x, R_X=1)}{\pob(x, a, w\given (R_W, R_X)=\ind)}, \\
        &b_4(x) = \frac{\tpr(x)}{\pob(x\given R_X=1)}.
    \end{align*}
\begin{proof}
See \S\ref{proof:extended PV Subopt} for a detailed proof.
\end{proof}
\end{theorem}
The arguments are similar except that each action should be “uniformly covered” in the observational process if we want $b_1$ to be bounded by \eqref{cond:b1 PV extended}, which implies that $|\cA|$ should be finite or a bounded set. Moreover, $\eta_k$ bounds the maximal critical radius for the function class $\tilde\cQ_k$, which also bounds the critical radius of the policy class $\Pie$. In \S\ref{app:POMDP critical radius}, such a critical radius is calculated with linear function class assumptions for the one-step POMDP.

\section{Application of CCB-IV: Linear Dynamic Treatment Regimes} \label{sec:DTR}

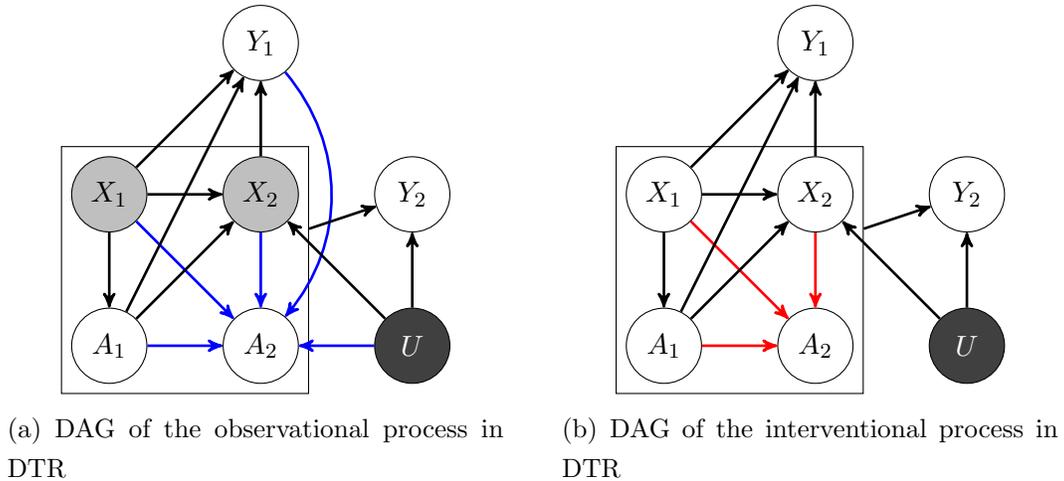
\begin{figure}[h]  
\centering 
  \begin{subfigure}[b]{0.4\linewidth}
  \centering
    \begin{tikzpicture}
        \node[missingstate] (X1) {$X_1$};
        \node[missingstate] (X2) [right=of X1] {$X_2$};
        \node[state] (A1) [below=of X1] {$A_1$};
        \node[state] (A2) [below=of X2] {$A_2$};
        \node[state] (Y1) [above=of X2] {$Y_1$};
        \node[state] (Y2) [right=of X2] {$Y_2$};
        \node[hiddenstate] (U) [below=of Y2] {$U$};
        \node[draw, fit=(X1) (X2) (A1) (A2)] (H) {};

        \path[normal] (X1) edge (A1);
        \path[normal, blue] (X1) edge (A2);
        \path[normal] (X1) edge (X2);
        \path[normal] (X1) edge (Y1);
        \path[normal, blue] (X2) edge (A2);
        \path[normal] (X2) edge (Y1);
        \path[normal] (A1) edge (Y1);
        \path[normal] (A1) edge (X2);
        \path[normal, blue] (A1) edge (A2);
        \path[normal, blue] (Y1) edge[bend left=40] (A2);
        \path[normal] (U) edge (X2);
        \path[normal] (U) edge (Y2);
        \path[normal, blue] (U) edge (A2);
        \path[normal] (H) edge (Y2);
    \end{tikzpicture}
    \caption{DAG of the observational process in DTR}
  \end{subfigure}\qquad
\begin{subfigure}[b]{0.4\linewidth}
\centering
  \begin{tikzpicture}  
        \node[state] (X1) {$X_1$};
        \node[state] (X2) [right=of X1] {$X_2$};
        \node[state] (A1) [below=of X1] {$A_1$};
        \node[state] (A2) [below=of X2] {$A_2$};
        \node[state] (Y1) [above=of X2] {$Y_1$};
        \node[state] (Y2) [right=of X2] {$Y_2$};
        \node[hiddenstate] (U) [below=of Y2] {$U$};
        \node[draw, fit=(X1) (X2) (A1) (A2)] (H) {};

        \path[normal] (X1) edge (A1);
        \path[normal, red] (X1) edge (A2);
        \path[normal] (X1) edge (X2);
        \path[normal] (X1) edge (Y1);
        \path[normal, red] (X2) edge (A2);
        \path[normal] (X2) edge (Y1);
        \path[normal] (A1) edge (Y1);
        \path[normal] (A1) edge (X2);
        \path[normal, red] (A1) edge (A2);
        \path[normal] (U) edge (X2);
        \path[normal] (U) edge (Y2);
        \path[normal] (H) edge (Y2);
\end{tikzpicture}
\caption{DAG of the interventional process in DTR}
\end{subfigure}
\caption{A DAG illustrating the DTR model. Note that $Y_2$ depends on the whole trajectory $H$}\label{fig:DTR}
\end{figure}

\paragraph{Background.} Dynamic treatment regimes (DTRs) is an extension of the individualized treatment rules (ITRs) to multi-steps. Estimating optimal policy can be challenging with unmeasured confounders in the observational dataset
\cite{chen2021estimating, qi2021proximal, NEURIPS2019_8252831b, singh2022automatic}.
We consider a DTRs with two steps, which is graphically represented in Figure \ref{fig:DTR}. 

In the observational process, at stage $i\in\{1, 2\}$, the treatment $A_i$ is selected based on the current state $X_i$ and the historical information $\{(X_j, A_j, Y_j)\}_{j=1}^{i-1}$. Then the state transits to $X_{i+1}$ and a reward $Y_i$ is generated. At the second state, $X_2$, $A_2$ and $Y_2$ are confounded by an unmeasured confounder $U$. Since the first step is not influenced by $U$, we have $Y_1\indep U\given (X_1, X_2, A_1)$. Therefore, we see that the first state reward $Y_1$ serves as an instrumental variable to $A_2$. Here, we provide table \ref{tab:DTRs} to illustrate the mapping from this two-stage DTRs to the CCB-IV model. 
\begin{table}[]
    \centering
    \begin{tabular}{  c  c c c} 
  \hline
  Variable Type & Two-step DTRs & Observability in $\cD$ & Correspondence to CCB-IV\\ 
  \hline
  confounder    &   $U$                 & unobservable  &   $U$\\
  context       &   $(X_1, X_2, A_1)$   & partially missing  & $X$ \\
  treatment     &   $A_2$               & observable & $A$\\
  outcome       &   $Y_2$               & observable & $Y$\\
  IV            &   $Y_1$               & observable & $Z$\\
  \hline
\end{tabular}
    \caption{Mapping of variables from the two-state DTRs to the CCB-IV.}
    \label{tab:DTRs}
\end{table}
We assume that $Y_2=f(X_1, X_2, A_1, A_2)+\varepsilon$ where $\varepsilon\indep (A_1, A_2)\given (U, X_1, X_2)$ and that $Y_1$ and $Y_2$ are fully observed, which satisfies the structured reward assumption. 
For $Y_1$ to function as an IV, we require that $Y_1$ is complete over $(X_1, X_2, A_1, A_2)$. 
Additionally, in the observational process, the missingness indicator $R_{X_1}$ is caused by $(X_1, A_1)$ and the missingness indicator $R_{X_2}$ is caused by $(X_1, A_1, X_2, A_2)$. 
Therefore, the model assumptions for CCB-IV are satisfied. 
By Theorem \ref{thm:IV identification}, the CATE $g(x_1, a_1, x_2, a_2)=\EEob\sbr{Y_2\given x_1, a_1, x_2, \doopt(a_2)}$ is identified by
\begin{align}
    &\EEob\sbr{h_1(A, Y) - Y_2\given Y_1=y_1}=0, \label{eq:DTR ID 1}\\
    &\EEob\sbr{g(X, A)-h_1(A, Y) \given (X, A, Y_1)=(x, a, y_1), (R_{X_1}, R_{X_2})=\ind}=0.\label{eq:DTR ID 2}
\end{align}
We assume that the solutions $h_1$ and $g$ always exist, which requires certain completeness conditions for $Y_2$ to restore the missingness in $X_1$ and $X_2$.
Here, we let $X=(X_1, X_2)$, $A=(A_1, A_2)$ and $Y=(Y_1, Y_2)$ in the remaining part of the section for DTRs example. 
Note that $Y_2\indep \pib_1\given (X_1, A_1, X_2)$ in the observational settings, since $A_1$ is not confounded. We thereby have $g(x, a)=\EEob\sbr{Y_2\given x, \doopt(a)}$.
Our optimization target thereby corresponds to maximizing the average reward function on $\pie=(\pi_1, \pi_2)\in\Pi$,
\begin{align*}
    v(g, \pie)=&\int_{\cX\times\cA} g(x,a) \tpr(x_1) \pob(x_2\given x_1, a_1)   \pi_1(a_1\given x_1) \pi_2(a_2\given x_1, a_1, x_2)
    \rd x\rd a,
\end{align*}
where we assume that $\pob(x_2\given x_1, a_1)$ is already known for brevity, although a little extension of our framework is capable of dealing with unknown $\pob(x_2\given x_1, a_1)$ by learning from data. Moreover, we only consider $Y_2$ as the reward. We remark that, since the first stage is not confounded, $\EEob\sbr{Y_1\given x_1, \doopt(a_1), x_2}$ can also be easily learned and integrated into the average reward. Now we pose the following assumptions on the linearity of the DTRs model.

\paragraph{Linear function class.} We make the following assumptions to ensure the existence of the bridge functions and the linearity of our DTRs model.
\begin{assumption}[Existence of linear bridge functions]\label{asp:DTRs linear 1}
We assume that a solution $\vhstar=(\hstark{1}, \gstar)$ exists to the IES given by \eqref{eq:DTR ID 1} and \eqref{eq:DTR ID 2}. Furthermore, we assume that $\hstark{1}, \gstar$ fall into the following function classes,
\begin{align*}
    &\cH_1=\{h_1\given h_1(\cdot)=w_1^\top \phi_1(\cdot), \nbr{w_1}_2\le C_1, \nbr{\phi_1(\cdot)}_2\le 1\}, \\
    &\cG=\{g\given g(\cdot)=w_2^\top \phi_2(\cdot), \nbr{w_2}_2\le C_2, \nbr{\phi_2(\cdot)}_2\le 1\}, 
\end{align*}
where $\phi_1:\cA\times\cY\rightarrow \RR^{m_1}$ and $\phi_2:\cX\times\cA\rightarrow \RR^{m_2}$.
Moreover, we assume that $\hstark{1}=(w_1^*)^\top \phi_1$ and $\gstar=(w_2^*)^\top\phi_2$.
\end{assumption}
Assumption \ref{asp:DTRs linear 1} assumes the existence and linearity of $\vh$.
We remark that it suffices for (i) to hold if $\EEob\sbr{Y_2\given y_1}$ is captured by the linear kernel $\EEob\sbr{\phi_2(X, A)\given y_1}$, and it suffices for assumption (ii) to hold if $\gstar(x, a)$ is captured by the linear kernel $\EEob\sbr{\phi_1(A, Y)\given y_1, x, a}$ for any $y_1\in\cY_1$.
In addition, Assumption \ref{asp:DTRs linear 1} also suggests that by using $\cH=\cH_1\times \cG$ as the hypothesis class, we have no realizability error, i.e., $\vareH=0$.
For the linear kernel $\phi_1$ and $\phi_2$, we continue to assume that their conditional expectation also falls into some linear spaces. 
\begin{assumption}[Linearity of dual function class]\label{asp:DTRs linear 2}
We assume that the conditional expectations of kernel $\phi_1$ and $\phi_2$ with respect to \eqref{eq:DTR ID 1}  and \eqref{eq:DTR ID 2} satisfy,
\begin{itemize}
\item[(i)] $\EEob\sbr{\phi_1(A, Y)\given y_1} = W_1 \psi_1(y_1)$, where $\psi_1:\cY_1\rightarrow \RR^{d_1}$, $W_1\in\RR^{m_1\times d_1}$.
\item[(ii)] $\EEob\sbr{\phi_1(A, Y)\given (x, a, y_1), (R_{X_1}, R_{X_2})=\ind}=W_2\psi_2(x, a, y_1)$, where $\psi_2:\cX_1\times\cA_1\times\cX_2\times\cA_2\times\cY_1\rightarrow \RR^{d_2}$, $W_2\in\RR^{m_1\times d_2}$.
\item[(iii)] $\EEob\sbr{\phi_2(X, A)\given (x, a, y_1), (R_{X_1}, R_{X_2})=\ind}=W_3\psi_2(x, a, y_1)$, where $W_3\in\RR^{m_2\times d_2}$.
\end{itemize}
\end{assumption}
In Assumption \ref{asp:DTRs linear 2}, we remark that if the operator $T:\cF(\cA\times\cY)\rightarrow \cF(\cY_1)$ defined as $Tf(y_1) =\EEob\sbr{f(A, Y)\given y_1}$ is captured by the kernel $\psi_1(y_1)$, it suffices for $W_1$ to exists. Similarly, it suffices for (ii), (iii) to hold if the corresponding operators are captured by the linear kernel $\psi_2(x, a, y_1)$.
Following Assumption \ref{asp:DTRs linear 2}, it holds for the linear operator $\cT$ that
\begin{align*}
    &\cT_1 \vh(y_1) = (w_1-w_1^*)^\top W_1 \psi_1(y_1), \\
    &\cT_2 \vh(x, a, y_1) = \rbr{(w_2-w_2^*)^\top W_3-(w_1-w_1^*)^\top W_2}\psi_2(x, a, y_1), 
\end{align*}
which suggests that $\cT_1\vh$ and $\cT_2\vh$ fall into the following linear function classes, 
\begin{align*}
    \Theta_k=\{\theta_k\given \theta_k(\cdot)= \beta_k^\top \psi_k(\cdot), \beta_k\in\RR^{d_k}, \nbr{\beta_k}\le D_k, \nbr{\psi_k(\cdot)}_2\le 1\}, 
\end{align*}
where we have $D_1>2C_1 \nbr{W_1}_F$ and $D_2> 2 (C_1\nbr{W_2}_F+C_2\nbr{W_3}_F)$.
By further letting $\vTheta=\Theta_1\times \Theta_2$, the dual function space has full compatibility, i.e., $\vareH$ and $\vareTheta$ are both zero. By Theorem \ref{thm:IV subopt}, we have the following corollary to establish the convergence of the sub-optimality for the two-step linear DTRs.
\begin{corollary}[Convergence of sub-optimality for linear DTRs]
Suppose that Assumptions \ref{asp:DTRs linear 1} and \ref{asp:DTRs linear 2} hold. Let $e_{\cD}>(2 L_\alpha^2+5/ 4)\eta^2$ where $\eta=\sum_{k=1}^2\eta_k^2$ and $\eta_k$ bounds the critic radius for function class $\cQ_k=\{\alpha_k(\vh, \cdot)\theta_k: \vh\in\vH, \theta_k\in\Theta_k\}$. Assume that for any $x\in\cX$ and $a\in\cA$, there exists $b_1:\cY_1\rightarrow \RR$ satisfying
\begin{align*}
    \EEob\sbr{b_1(Y_1)\given x, a} = \frac{\tpr(x_1)\pob(x_2\given x_1, a_1)\piestar_1(a_1\given x_1)\piestar_2(a_2\given x_1, a_1, x_2)}{\pob(x_1, a_1, x_2, a_2)}.
\end{align*}
The sub-optimality of $\piepessi$ for the two-step DTRs is bounded with probability at least $1-4\xi$ by
\begin{align*}
    \SubOpt(\piepessi) \lesssim  \rbr{\nbr{b_1}_{\mu_1, 2}+\nbr{b_2}_{\mu_2, 2}}\cdot \rbr{O\rbr{\sqrt{e_\cD}} +  O\rbr{\eta}},
\end{align*}
where $b_2:\cX\times\cA\times\cY_1\rightarrow \RR$ is defined by
\begin{gather*}
    b_2(x,a, y_1)=\frac{b_1(y_1)\pob(x_1, a_1, x_2, a_2, y_1)}{\pob(x_1, a_1, x_2, a_2, y_1\given(R_{X_1}, R_{X_2})=\ind)}.
\end{gather*}
\end{corollary}
As is proved in \S\ref{app:DTRs critical radius}, the critical radiuses are of order $\eta_1=\cO(\sqrt{(m_1+d_1)\log T/T})$ and $\eta_2=\cO(\sqrt{\max\{m_1+m_2+d_2\}\log T_2/T_2})$, where $T$ corresponds to the size of the whole dataset $\cD$ and $T_2$ corresponds to the size of the dataset satisfying $(R_{X_1}, R_{X_2})=\ind$. Such a result shows that the convergence rate is of the order $\cO(\sqrt{\log T_2/T_2})$ if we choose $e_\cD=\cO(\log T_2/T_2)$.
Note that $T_2$ is the total number of samples that are subject to no missingness, which requires that there should be a fixed proportion of samples on which we have fully observed contexts and side-observations to guarantee the fast convergence rate.

\section{Application of CCB-PV with Extended Policy Class: One-step Linear Partially Observable Markov Decision Process}\label{sec:POMDP}
 

\paragraph{Background.}
We consider a one-step Partially Observable Markov Decision Process (POMDP) following the example in \citep{shi2021minimax, uehara2021finite}. Here, the term "one-step" means that we only care about the policy and reward at the first step, but the environment is allowed to transit into the following steps. 
The POMDP starts with a pre-observation $O^-$ and the environment transits into state $S$. An observation $O$ is generated according to $S$, and the agent in the observational process takes an action $A$ according to $\pib:\cS\rightarrow \Delta(\cA)$. After the action is conducted, a reward $Y_0$ is received and the environment transits into the following state $S^+$ with observation $O^+$. 
Note that $Y$ is allowed to depend on $O$.
In the interventional process, since the agent gains no access to the hidden state, its policy can only depend on the observations.
We consider the extended interventional policy class discussed in \S\ref{sec:extended CCB-PV}, i.e.,  $\pie:\cO^-\times\cO\rightarrow\Delta(\cA)$ by viewing $O^-$ as the context ($X$ in CCB-PV) and $O$ as the outcome proxy ($W$ in CCB-PV).
Note that such a policy also captures the case where the policy only depends on $O$.

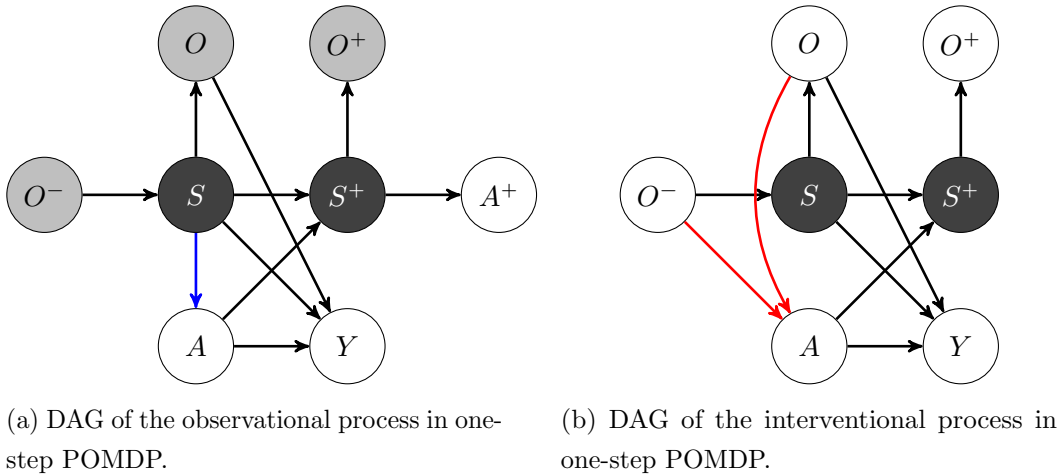
\begin{figure}[h]  
\centering 
  \begin{subfigure}[b]{0.4\linewidth}
  \centering
    \begin{tikzpicture}
        \node[hiddenstate] (S) [right=of O] {$S$};
        \node[missingstate] (O) [above=of S] {$O$};
        \node[missingstate] (O-) [left=of S] {$O^-$};
        \node[state] (A) [below=of S] {$A$};
        \node[hiddenstate] (S+) [right=of S] {$S^+$};
        \node[state] (Y) [below=of S+] {$Y$};
        \node[missingstate] (O+) [above=of S+] {$O^+$};
        \node[state] (A+) [right=of S+] {$A^+$};
        \path[normal] (O-) edge (S);
        \path[normal] (O) edge (Y);
        \path[normal] (S) edge (O);
        \path[normal, blue] (S) edge (A);
        \path[normal] (S) edge (S+);
        \path[normal] (S) edge (Y);
        \path[normal] (A) edge (S+);
        \path[normal] (A) edge (Y);
        \path[normal] (S+) edge (O+);
        \path[normal] (S+) edge (A+);
    \end{tikzpicture}
    \caption{DAG of the observational process in one-step POMDP.}
  \end{subfigure}\qquad
\begin{subfigure}[b]{0.4\linewidth}
\centering
  \begin{tikzpicture}  
        \node[hiddenstate] (S) [right=of O] {$S$};
        \node[state] (O) [above=of S] {$O$};
        \node[state] (O-) [left=of S] {$O^-$};
        \node[state] (A) [below=of S] {$A$};
        \node[hiddenstate] (S+) [right=of S] {$S^+$};
        \node[state] (Y) [below=of S+] {$Y$};
        \node[state] (O+) [above=of S+] {$O^+$};
        \path[normal] (O-) edge (S);
        \path[normal] (O) edge (Y);
        \path[normal] (S) edge (O);
        \path[normal] (S) edge (S+);
        \path[normal] (S) edge (Y);
        \path[normal] (A) edge (S+);
        \path[normal] (A) edge (Y);
        \path[normal] (S+) edge (O+);
        \path[normal, red] (O) edge[bend left=-30] (A);
        \path[normal, red] (O-) edge (A);
\end{tikzpicture}
\caption{DAG of the interventional process in one-step POMDP.}
\end{subfigure}
\caption{A DAG illustration of the one-step POMDP model.}
\end{figure}

\paragraph{Missingness.} Very similar to the DTRs example, we assume that $R_{O}$ is caused by $O$ and $A$ and $R_{O^+}$ is caused by $O^+$ and $A^+$ in the observational dataset. Note that the pre-observation $O^-$ is exogenous to the model, it is thereby reasonable to assume that $R_{O^-}$ only depends on $O^-$. A tricky part is that following the observational policy, we have that $A^+\sim \pib(a^+\given s^+)$ and $S^+\sim \pr(s^+\given s, a)$. It thus turns out that $R_{O^+}$ is alternatively caused by $(O^+, S, A)$. 

\paragraph{Mapping to CCB-PV.} We provide a mapping from this one-step POMDP to the CCB-PV in Table \ref{tab:POMDP}. 
It is easy to verify that the assumption of PV independence and the assumption of unconfounded and outcome-independent missingness in Assumption \ref{asp:CCB-PV} both hold for this one-step POMDP. The PV complete assumption corresponds to assuming that $O^+$ is complete over $S$, i.e., for any $a\in\cA$, $o^-\in\cO^-$, $\EEob\sbr{\sigma(S)\given o^-, a, o^+, R_{O^+}=1}=0$ holds for any $o^+\in\cO^+$ if and only if $\sigma(S^+)\overset{\text{a.s.}}{=} 0$ holds. Such an assumption suggests this should be a non-degenerate MDP, i.e., $O^+$ still contains sufficient information of the hidden state of the previous step.
Then following Theorem \ref{thm:CCB-PV ID extension}, we have the $v^\pie(x)$ identified by
\begin{align}
    &\EEob\sbr{h_1(Y, A, O^-, O^+) \given a, o^-, o^+, (R_{O^-}, R_{O^+})=\ind}=0, \label{eq:POMDP ID 1}\\
    &\EEob\sbr{h_2(A, O, O^-) - h_1(Y, A, O^-, O^+)-Y\pie(A\given O^-, O)\given a, o, o^-, o^+, (R_{O}, R_{O^-}, R_{O^+})=\ind}, 
    \nend
    &\EEob\sbr{h_3(Y, A, O^-)-\sum_{a'\in\cA} h_2(a', O, O^-)\given a, o, o^-, (R_{O}, R_{O^-})=\ind}, 
    \nend
    &\EEob\sbr{g^\pie(O^-)-h_3(Y, A, O^-)\given o^-, R_{O^-}=1}=0, \label{eq:POMDP ID 4}
\end{align}
if the bridge functions exist.
\begin{table}[]
    \centering
    \begin{tabular}{  c  c c c} 
  \hline
  Variable Type & One-step POMDP  & Observability in $\cD$ & Correspondence to CCB-PV\\ 
  \hline
  confounder        &   $S$ & unobservable        & $U$    \\
  context           &   $O^-$ & partially missing   & $X$    \\
  treatment         &   $A$ & observable          & $A$    \\
  outcome           &   $Y$ & observable          & $Y$    \\
  treatment proxy   &   $O^+$ & partially missing   & $Z$    \\
  outcome proxy     &   $O$ & partially missing   & $W$    \\
  \hline
\end{tabular}
    \caption{Mapping of variables from the one-step POMDP to the CCB-PV.}
    \label{tab:POMDP}
\end{table}

\paragraph{Linear function class.} Similar to the linear DTRs example, we characterize the existence of the bridge functions and the linearity of the one-step POMDP model. We assume that the interventional policy falls into some linear function class. Specifically, we let $\Pie$ be a subset of the following linear function class,
\begin{align}
\Pie = \cbr{\pie \bigg | \pie(a\given o, o^-)=\frac{\exp\rbr{w_0^\top\phi_0(a, o, o^-)}}{\sum_{a'\in\cA}\exp\rbr{w_0^\top\phi_0(a', o, o^-)}}, w_0\in\RR^{m_2}, \nbr{w_0}_2\le C_0,  \nbr{\phi_0(\cdot)}_2\le 1}.\label{def:linear policy}
\end{align}
\begin{assumption}[Existence of linear bridge function] \label{asp:POMDP linear 1}
We assume that for any $\pie\in\Pie$, there exists $\vh^{\pie, *}=(h_1^{\pie, *}, h_2^{\pie, *}, h_3^{\pie, *}, g^{\pie, *})$ as a solution to the IES \eqref{eq:POMDP ID 1}-\eqref{eq:POMDP ID 4}.
In addition, we assume that $h_1^{\pie, *}, h_2^{\pie, *}, h_3^{\pie, *}$ fall into the following linear function classes,
\begin{gather*}
    \cH_k=\{h_k\given h_k(\cdot)= w_k^\top\phi_k(\cdot), w_k\in\RR^{m_k}, \nbr{w_k}_2\le C_k, \nbr{\phi_k(\cdot)}\le 1\}, \quad k=1, 2, 3, 
\end{gather*}
with $h_k^{\pie, *}=(w_k^{\pie, *})^\top \phi_k$.
\end{assumption}

Assumption \ref{asp:POMDP linear 1} assumes the bridge functions to exist and fall into some linear function classes.
Now for the corresponding kernels $\phi_1$, $\phi_2$ and $\phi_3$, we assume their conditional moments are captured by kernel series $\psi_1, \psi_2, \psi_3, \psi_4$. 
\begin{assumption}[Linearity of the dual function class]\label{asp:POMDP linear 2}
We assume that the kernel $\phi_1$, $\phi_2$, $\phi_3$, $\psi_4$ satisfies
\begin{itemize}
    \item[(i)] $\EEob\sbr{\phi_1(Y, A, O^-, O^+)\given a, o^-, o^+, R_{o^+}=1} = W_1 \psi_1(a, o^-, o^+)$ where $\psi_1:\cA\times\cO^-\times\cO^+\rightarrow \RR^{d_1}$, $W_1\in \RR^{m_1\times d_1}$, and $\nbr{\psi_1(\cdot)}_2\le 1$.
    \item[(ii)] $\EEob\sbr{\phi_1(Y, A, O^-, O^+)\given a, o, o^-, o^+, R_{O^+}=1}=W_2\psi_2(a, o, o^-, o^+)$ and $\EEob[\phi_2(A, O, O^-) \given a, o, \allowbreak o^-, o^+, R_{O^+}]=W_3\psi_2(a, o, o^-, o^+)$ where $\psi_2:\cA\times\cO\times\cO^-\times\cO^+\rightarrow \RR^{d_2}$, $W_2\in\RR^{m_1\times d_2}$, $W_3\in\RR^{m_2\times d_2}$, and $\nbr{\psi_2(\cdot)}_2\le 1$.
    \item[(iii)] $\sum_{a'\in\cA}\phi_2(a', O, O^-)=W_4\psi_3(a, o, o^-)$ for any $a\in\cA$ and $\EEob\sbr{\phi_3(Y, A, O^-)\given a, o, o^-}=W_5\allowbreak \psi_3(a, o, o^-)$, where $\psi_3:\cA\times\cO\times\cO^-\rightarrow \RR^{d_3}$, $W_4\in\RR^{m_2\times d_3}$, $W_5\in \RR^{m_3\times d_3}$, and $\nbr{\psi_3(\cdot)}_2\le 1$.
    \item[(iv)] $\EEob\sbr{\phi_3(Y, A, O^-)\given o^-}=W_6 \psi_4(o^-)$ where $\psi_4:\cO^-\rightarrow \RR^{m_4}$, $W_5\in\RR^{m_3\times m_4}$, and $\nbr{\psi_4(\cdot)}_2\le 1$.
\end{itemize}
\end{assumption}
Consider a linear operator $T:\cF(\cA, \cO, \cO^+)\rightarrow \cF(\cA, \cO^-, \cO^+)$ defined as $Tf(a, o^-, o^+)=\EEob\sbr{f(A, O, O^-)\given a, o^-, o^+, R_{o^+}=1}$.
Condition (i) of Assumption \ref{asp:POMDP linear 2} indicates that the operator $T$ is captured by the kernel $\psi_1(a, o^-, o^+)$.
The arguments for conditions (ii)-(iv) are similar. 
Using condition (iv) of Assumption \ref{asp:POMDP linear 2} in conditional moment equation \eqref{eq:POMDP ID 4}, it holds for the CATE $g^{\pie, *}$  that,
\begin{align*}
    g^{\pie, *}(o^-)=\EEob\sbr{\h{3}{\pie, *}(Y, A, O^-)\given o^-} = (w_3^{\pie, *} )^\top W_6 \psi_4(o^-), 
\end{align*}
which implies that $g^{\pie, *}$ lies in the linear space $\cG=\{w_4\in\RR^{m_4}:\cO^-\rightarrow w_4^\top\psi_4(\cdot), \nbr{w_4}\le C_3\nbr{W_6}_F, \nbr{\psi_4(\cdot)}_2\le 1\}$.
Therefore, by letting $\vH=\cH_1\times \cH_2\times \cH_3\times \cG$ be the hypothesis space, we have the realizability error $\vareH$ equal to zero.
Combining Assumptions \ref{asp:POMDP linear 1} and \ref{asp:POMDP linear 2}, it further holds for the linear operator $\cT^\pie$ that,
\begin{align}
    &\cT^\pie_1\vh(a,o^-, o^+)=(w_1-w_1^{\pie, *})^\top W_1 \psi_1(a, o^-, o^+), \nend
    &\cT^\pie_2\vh(a, o, o^-, o^+)=\rbr{(w_2-w_2^{\pie, *})^\top W_3 - (w_1-w_1^{\pie, *})^\top W_2}\psi_2(a, o, o^-, o^+), \nend
    &\cT^\pie_3\vh(a, o, o^-) = \rbr{(w_3-w_3^{\pie, *})^\top W_5- (w_2-w_2^{\pie, *})^\top W_4} \psi_3(a, o, o^-), \nend
    &\cT^\pie_4\vh(o^-) = \rbr{(w_4-w_4^{\pie, *})^\top - (w_3-w_3^{\pie, *})^\top W_6}\psi_4(o^-).\nonumber
\end{align}
Therefore, $\cT^\pie_k\vh$ falls into the following linear function class
\begin{align*}
 \Theta_k=\{\theta_k\given \theta_k(\cdot)=\beta_k^\top \psi_k(\cdot), \beta_k\in\RR^{d_k}, \nbr{\beta_k}\le D_k, \nbr{\psi_k(\cdot)}_2\le 1\}, \quad k=1, 2,3, 4, 
\end{align*}
where we require $D_1>2C_1 \nbr{W_1}_F$, $D_2> 2 (C_1\nbr{W_2}_F+C_2\nbr{W_3}_F)$, $D_3> 2 (C_3\nbr{W_5}_F+C_2\nbr{W_4}_F)$, and $D_4> 2 (C_4+C_3\nbr{W_6}_F)$.
Using $\vH=\cH_1\times \cH_2\times \cH_3\times \cG$ as the hypothesis space and $\Theta=\Theta_1\times \Theta_2\times \Theta_3\times \Theta_4$ as the dual function class,  we have the following corollary for the convergence of the sub-optimality for the one-step linear POMDP by Theorem \ref{thm:CCB-PV ID extension}.
\begin{corollary}
Suppose that Assumptions \ref{asp:POMDP linear 1} and \ref{asp:POMDP linear 2} hold. Let $e_{\cD}>(2 L_\alpha^2+5/ 4)\eta^2$ where $\eta=\sum_{k=1}^2\eta_k^2$ and $\eta_k$ bounds the critic radius for the function class $\cQ_k=\{\alpha_k^\pie(\vh, \cdot)\theta_k: \vh\in\vH, \theta_k\in\Theta_k, \pie\in\Pie\}$. Suppose that for any $o^-\in\cO^-$, $a\in\cA$ and $o^+\in\cO^+$ there exists $b_1:\cO^-\times\cA\times\cO^+\rightarrow\RR$ satisfying
\begin{align*}
    \EEob\sbr{b_1(O^-, A, O^+)\given o^-, s, a, R_{O^+}=1} = \frac{\pob(s\given o^-)\tpr(o^-)}{\pob(s, o^-, a\given  R_{o^+}=1)}.
\end{align*}
The sub-optimality corresponding to $\piepessi$ for the CCB-PV is bounded with probability at least $1-2K\xi$ by
    \begin{align*}
        \SubOpt(\piepessi) \lesssim \sum_{k=1}^4 \nbr{b_k}_{\mu_k, 2} \cdot \rbr{ O\rbr{\sqrt{e_\cD}} +  O\rbr{\eta}}, 
    \end{align*}
    where $b_2:\cA\times\cO\times\cO^-\rightarrow \RR$, $b_3:\cO\times\cO^-\times\cA\rightarrow\RR$ and $b_4:\cO^-\rightarrow\RR$ characterize the distribution shift and are defined by
    \begin{align*}
        &b_2(a, o, o^-) = b_1(o^-, a, o^+)\frac{\pob(a, o, o^-\given (R_{O^-}, R_{O^+})=\ind)}{\pob(a, o, o^-\given (R_{O}, R_{O^-}, R_{O^+})=\ind)}, \\
        &b_3(o, o^-, a) = \frac{\tpr(o^-)\pob(a, o\given o^-, R_{O^-}=1)}{\pob(o^-, a, o\given (R_{O}, R_{O^-})=\ind)}, \\
        &b_4(o^-) = \frac{\tpr(o^-)}{\pob(o^-\given R_{O^-}=1)}.
    \end{align*}
\end{corollary}
The critical radius is calculated in \S\ref{app:POMDP critical radius}. The result can be summarized as $\eta=\cO(|\cA|\sqrt{\log T_2/T_2})$, where $T_2$ corresponds the total number of samples that are subject to no missingness in the observations $(O, O^-, O^+)$. Therefore, we establish the convergence of the sub-optimality for the one-step linear POMDP.

\paragraph{Discussion of RKHS Space.} 
We remark that a similar result can also be established for other function classes, e.g., the RKHS space.
Following Proposition 6.3 in \cite{duan2021risk}, the critical radius for a RKHS space $\cF$ with kernel $K$ and bounded norm $\nbr{f}_\cK\le C$ is given by,
\begin{align*}
    \eta=2\min_{j\in\NN} \cbr{\frac j T + C\sqrt{\frac{2}{T}\sum_{i=j+1}^\infty \lambda_i^\cF}}, 
\end{align*}
where $\lambda_i^\cF$ corresponds to the eigenvalues of the kernel $K$. If the eigenvalues decay exponentially with high probability, we can also obtain a fast convergence rate of order $\cO(\sqrt{1/T})$.


\section{Conclusion} 
In this work, we propose a provably efficient algorithm CAP for offline confounded contextual bandit with missing observations.
The essential idea of the CAP algorithm is to (i) form an integration equation system (IES) for identification of the CATE; (ii) reformulate the IES as an unconditional moment minimax estimator and yield the confidence set from the estimator; (iii) Such an uncertainty quantification makes it valid to construct a policy taking greedy pessimistic action.
To the best of our knowledge, CAP is the first pessimism-based algorithm solving the offline confounded contextual bandit with missing observations.

\newpage
\bibliographystyle{ims}
\bibliography{reference}

\newpage 
\appendix

\section{A Matrix Explanation for the IES of the CCB-PV}\label{app:PV details} 
The matrix explanation for CCB-PV is just the same as CCB-IV. 
Without missingness in $W$, it can be easily verified that \eqref{eq:PV ID 1}-\eqref{eq:PV ID 4} give a reduced integral equation system, 
\begin{gather}
    \EEob\sbr{Y-h_2(A, W, X)\given A=a, X=x, Z=z, (R_Z, R_X)=\ind} = 0, \label{eq:PV matrix id 1}\\
    g(x, a') = \EEob\sbr{h_2(a', W, X)\given X=x, R_X=1},\label{eq:PV matrix id 2}
\end{gather}
which is consistent with the standard identification equations for the PV model \citep{miao2018confounding, miao2018identifying} by also ignoring the conditions for $R_X$ and $R_Z$. 
Now, we show how we get around the missingness of $W$. We can rewrite \eqref{eq:PV matrix id 1} as
\begin{align}
    &\EEob\sbr{Y\given A=a, X=x, Z=z, (R_Z, R_X)=\ind}\nend
    &= \underbrace{h_2(A, W, X) P(Y\given W, a, x, z, (R_Z, R_X, R_W)=\ind)^\dagger}_{h_1(Y, a, x, z)} P(Y\given a, x, z, (R_X, R_Z)=\ind)
    ,  \label{eq:PV matrix 1}
\end{align}
where the equality holds by noting that
\begin{align}
    P(Y\given a, x, z, (R_X, R_Z)=\ind)=P(Y\given W, a, x, z, (R_Z, R_X, R_W)=\ind)P(W\given a, x, z, (R_Z, R_X)=\ind), \label{eq:PV matrix 3}
\end{align}
since $R_W$ is only caused by $(W, X, A)$.
We also rewrite \eqref{eq:PV matrix id 2} as
\begin{align}
    g(x, a') &= \sum_{a\in\cA}\underbrace{h_2(a', W, x) P(Y\given W, a, x, (R_W, R_X)=\ind)^\dagger }_{h_3(Y, a, x, a')}P(Y, a\given x, R_x=1)
    ,  \label{eq:PV matrix 2}
\end{align}
where the equality holds by noting that
\begin{align}
    P(Y, a\given x, R_X=1)=P(Y\given W, a, x, (R_W, R_X)=\ind) P(W, a\given x, R_X=1).\label{eq:PV matrix 4}
\end{align}
Here, \eqref{eq:PV matrix 3} and \eqref{eq:PV matrix 4} hold by the chain rule and noting that $R_W\indep (Y, Z, R_Z, R_X)\given (A, X, W)$.
Similar to the matrix explanation for the CCB-IV case, we show that \eqref{eq:PV matrix 1} with the introduction of bridge function $h_1$ gives \eqref{eq:PV ID 1} and \eqref{eq:PV ID 2}  while \eqref{eq:PV matrix 2} with the introduction of bridge function $h_3$ gives \eqref{eq:PV ID 3} and \eqref{eq:PV ID 4}.
\section{Proof of the Identification for CCB-IV and CCB-PV}
\subsection{Proof of Theorem \ref{thm:IV identification}}\label{pro:IV identification}
Under the assumption that there exists $h_1$ and $g$ satisfying \eqref{eq:IV bridge 1} and \eqref{eq:IV bridge 2}, we prove the conclusion of Theorem \ref{thm:IV identification} that $g$ is a recovery of the CATE of the CCB-IV almost surely in this subsection.
\begin{proof}
We first prove the following two equality, $\EEob\sbr{Y\given Z, R_Z=1}=\EEob\sbr{g(A, X)\given Z, R_Z=1}$ and $\EEob\sbr{Y\given Z, R_Z=1}= \EEob[\tilde f(A, X)\given Z, R_Z=1]$, where $\tilde f(a, x)$ is by our construction and is shown to be the exact CATE. Then, with the model assumption for CCB-IV, we can prove that $g$ recovers the CATE almost surely.
We start with $\EEob\sbr{Y\given Z, R_Z=1}$ and it holds that 
\begin{align}
    \EEob\sbr{Y\given Z, R_Z=1} &= \EEob\sbr{\EEob\sbr{Y\given X, Z}\given Z, R_Z=1} \nend
    &=\EEob\sbr{f(A, X) + \EEob\sbr{\EEob\sbr{\epsilon\given X, U, Z}\given X, Z} \given Z, R_Z=1}\nend
    &=\EEob\sbr{f(A, X) + \EEob\sbr{\EEob\sbr{\epsilon\given X, U}\given X} \given Z, R_Z=1},\label{eq:1 iv id pro}
\end{align}
where the first equality holds by noting that $R_Z\indep Y\given (X, Z)$, the second equality holds by noting that $Y=f(A, X)+\epsilon$, and the third equality holds by noting that $\epsilon\indep Z\given (X, U)$ and $Z\indep U\given X$.
Let $\tilde f(a, x)=f(a, x) + \EEob\sbr{\epsilon\given X=x}$. We continue with \eqref{eq:1 iv id pro} and obtain
\begin{align}
    \EEob\sbr{Y\given Z, R_Z=1}
    &=\EEob\sbr{f(A, X) + \EEob\sbr{\epsilon\given X} \given Z, R_Z=1}\nend
    &=\EEob\sbr{\tilde f(A, X)\given Z, R_Z=1}.\label{eq:2 iv id pro}
\end{align}
On the other hand, it also holds for $\EEob\sbr{Y\given Z, R_Z=1}$ that
\begin{align}
    \EEob\sbr{Y\given Z, R_Z=1}
    & = \EEob\sbr{h_1(Y, A, Z)\given Z, R_Z=1}\nend
    &= \EEob\sbr{\EEob\sbr{h_1(Y, A, Z)\given A, X, Z, (R_Z, R_X)=\ind}\given Z, R_Z=1},\label{eq:3 iv id pro}
\end{align}
where the first equality holds by \eqref{eq:IV bridge 1} and the second equality holds by noting that $R_X\indep (Y, R_Z)\given (A, X, Z)$.
Plugging \eqref{eq:IV bridge 2} into \eqref{eq:3 iv id pro}, it follows that
\begin{align}
    \EEob\sbr{Y\given Z, R_Z=1}
    &=\EEob\sbr{\EEob\sbr{g(A, X)\given A, X, Z, (R_Z, R_X)=\ind}\given Z, R_Z=1}\nend
    &=\EEob\sbr{g(A, X)\given Z, R_Z=1}. \label{eq:4 iv id pro}
\end{align}
Combining \eqref{eq:2 iv id pro} and \eqref{eq:4 iv id pro}, we arrive at
\begin{align*}
    \EEob\sbr{g(A, X)-\tilde f(A, X)\given Z, R_Z=1}=0. 
\end{align*}
By the IV completeness assumption ((i) in Assumption \ref{asp:CCB-IV}), it holds that
\begin{align}
    \tilde f(A, X) \overset{\text{a.s.}}{=} g(A, X).\label{eq:5 iv id pro}
\end{align}
Lastly, it remains to characterize the relationship between $\tilde f$ and the exact CATE. 
In the CCB-IV, the exact CATE $\gstar$ is given by
\begin{align*}
    \gstar(x, a) &= \EEin\sbr{Y\given X=x, \doopt(a)} \nend
    & =\EEob\sbr{\EEob\sbr{Y\given U, X, A=a}\given X=x},
\end{align*}
where the first equality holds by the definition of CATE, the second equality holds by noting that $Y\indep Z\given (A, U, X)$ and the definition of do-calculus. Recalling that $Y=f(A, X) + \epsilon$, it follows that
\begin{align*}
    \gstar(x, a)&= \EEob\sbr{\EEob\sbr{f(A, X) + \epsilon\given U, X, A=a}\given X=x}\nend
    & = \EEob\sbr{f(a, X) +\EEob\sbr{ \epsilon\given U, X}\given X=x} \nend
    &=f(a, x) + \EEob\sbr{\epsilon\given X=x},
\end{align*}
where the second equality holds by noting that $\epsilon\indep A\given (X, U)$. Recall the definition that $\tilde f(a, x)=f(a, x)+\EEob\sbr{\epsilon\given X=x}$, by \eqref{eq:5 iv id pro} we finally obtain
\begin{align}
    \gstar(x, a)= \tilde f(a, x)\overset{\text{a.s.}}{=} g(A, X),\label{eq:7 iv id pro}
\end{align}
which finishes the proof of Theorem \ref{thm:IV identification} that if $h_1$ and $g$ exists, $g$ recovers the CATE almost surely.
\end{proof}

\subsection{Proof of Remark \ref{rmk:IV existence}}\label{pro:rmk IV existence}
Our proof is given in two folds: (i) we give a proof of the “if” part in Remark \ref{rmk:IV existence}; (ii) we give a proof of the “only if“ part in Remark \ref{rmk:IV existence}.
\paragraph{“If" part.}We first prove the “if” part that it suffices for $h_1$ and $g$ to exist if there exists a solution $h_1$ to the following equation
\begin{align}
    \EEob\sbr{\gstar(A, X)-h_1(Y, A, Z)\given A, X, Z, R_Z=1}=0, \label{eq:6 iv id pro}
\end{align}
where $\gstar$ is the exact CATE. 
\begin{proof}
Let $\tilde g(a, x)=\gstar(a, x)$ and $\tilde h_1$ be the solution to \eqref{eq:6 iv id pro}. We just need to check that $\tilde g$ and $\tilde h_1$ satisfies \eqref{eq:IV bridge 1} and \eqref{eq:IV bridge 2}. Note that \eqref{eq:IV bridge 2} holds directly by \eqref{eq:6 iv id pro}. For \eqref{eq:IV bridge 1}, we see that
\begin{align}
    &\EEob\sbr{\tilde h_1(Y, A, Z)\given Z, R_Z=1}\nend
    &= \EEob\sbr{\EEob\sbr{\tilde h_1(Y, A, Z)\given A, X, Z, (R_Z, R_X)=\ind}\given Z, R_Z=1}\nend
    &=\EEob\sbr{\EEob\sbr{\gstar(A, X)\given A, X, Z, (R_Z, R_X)=\ind}\given Z, R_Z=1}\nend
    &=\EEob\sbr{\gstar(A, X)\given Z, R_Z=1},\label{eq:8 iv id pro}
\end{align}
where the first equality holds by noting that $R_X\indep (Y, R_Z)\given (A, X, Z)$ and the second equality holds by \eqref{eq:6 iv id pro}. Note that in the first part, we have already proved that $\EEob\sbr{Y\given Z, R_Z=1}= \EEob[\tilde f(A, X)\given Z, R_Z=1]$ with $\tilde f$ defined as $\tilde f(x, a)=f(x, a)+\EEob\sbr{\epsilon\given X=x}$  (see \eqref{eq:2 iv id pro}) and that $\gstar(x, a)=\tilde f(a, x)$ in \eqref{eq:7 iv id pro}. We remark that these two properties hold without any assumption on the existence of the bridge functions. Therefore, it holds for \eqref{eq:8 iv id pro} that
\begin{align*}
    \EEob\sbr{\tilde h_1(Y, A, Z)-Y\given Z, R_Z=1}=0,
\end{align*}
which justifies that $\tilde h$ and $\tilde g$ satisfy \eqref{eq:IV bridge 1} and \eqref{eq:IV bridge 2} and serve as a solution. 
\end{proof}

\paragraph{“Only if" part.}
We prove that any solution to the IES in Theorem \ref{thm:IV identification} must satisfy \eqref{cond:iv bridge exist}.
\begin{proof}
The proof is direct, following the fact that $g\aseq\gstar$ if $\vh=(h_1, g)$ satisfies the IES. By \eqref{eq:IV bridge 2} we have
\begin{align*}
    \EEob\sbr{\gstar(X, A) - h_1(Y, A, Z)\given A=a, X=x, Z=z, (R_Z, R_X)=\ind}=0.
\end{align*}
Noting that $R_X\indep (R_Z, Y)\given A, X, Z$, we thus have
\begin{align*}
    \EEob\sbr{\gstar(X, A) - h_1(Y, A, Z)\given A=a, X=x, Z=z, R_Z=1}=0.
\end{align*}
Thus, we complete the proof of Remark \ref{rmk:IV existence}.
\end{proof}

\subsection{Proof of Theorem \ref{thm:PV ID}}\label{pro:PV ID}
Under the assumption that there exists $h_1$, $h_2$, $h_3$ and $g$ satisfying \eqref{eq:PV ID 1}-\eqref{eq:PV ID 4}, we prove that $g$ is a recovery of the CATE of the CCB-PV in this subsection. 
\begin{proof}
We remark that $h_2$ corresponds to the value bridge function for identifying a PV model and $h_1$ and $h_3$ are additional bridge functions to deal with the missingness issue. In the following part, we first show that $\EEob\sbr{h_2(A, W, X)-Y\given U, A, X}\aseq 0$ and then prove that $g(x, a)=\gstar(x,a)$.
We start with the conditional expectation $\EEob\sbr{Y\given A, X, Z, (R_X, R_Z)=\ind}$.
\begin{align}
    &\EEob\sbr{Y\given A, X, Z, (R_X, R_Z)=\ind} \nend
    & = \EEob\sbr{h_1(Y, A, X, Z)\given A, X, Z, (R_X, R_Z)=\ind} \nend
    & = \EEob\sbr{\EEob\sbr{h_1(Y, A, X, Z)\given A, W, X, Z, (R_X, R_Z, R_W)=\ind}\given A, X, Z, (R_X, R_Z)=\ind}\nend
    &= \EEob\sbr{h_2(A, W, X)\given A, X, Z, (R_X, R_Z)=\ind}, \label{eq:1 pv id pro}
\end{align}
where the first equality holds by \eqref{eq:PV ID 1}, the second equality holds by noting that $R_W\indep (Z, Y, R_Z, R_X)\given (W, X, A)$, and the last equality holds by \eqref{eq:PV ID 2}.
We can rewrite \eqref{eq:1 pv id pro} by additionally conditioning on the confounder $U$,
\begin{align*}
    0 &= \EEob\sbr{Y-h_2(A, W, X)\given A, X, Z,(R_X, R_Z)=\ind}\nend
    &= \EEob\sbr{\EEob\sbr{Y-h_2(A, W, X)\given U, A, X, Z,  (R_X, R_Z)=\ind}\given A, X, Z, (R_X, R_Z)=\ind} \nend
    & = \EEob\sbr{\EEob\sbr{Y-h_2(A, W, X)\given U, A, X}\given A, X, Z,  R_Z=\ind},
\end{align*}
where the last equality holds by noting that $(R_X, R_Z, Z)\indep (Y, W, R_X)\given (U, A, X)$ and that $R_X\indep U\given (A, X, Z)$.
By (i) of Assumption \ref{asp:CCB-PV} (i) on the PV completeness, we have that
\begin{align}
    \EEob\sbr{Y-h_2(A, W, X)\given U, A, X}\aseq 0. \label{eq:3 pv id pro}
\end{align}
Now, it remains to show $g=\gstar$.
For $g$, we have
\begin{align}
    g(X, a')
    & = \EEob\sbr{h_3(Y, A, X;a')\given X, R_X=1}\nend 
    &= \EEob\sbr{\EEob\sbr{h_3(Y, A, X; a')\given A, W, X, (R_W, R_X)=\ind}\given X, R_X=1}\nend
    & = \EEob\sbr{\EEob\sbr{h_2(a', W, X)\given A, X, W, (R_W, R_X)=\ind}\given X, R_X=1}, \label{eq:2 pv id pro}
\end{align}
where the first equality holds by \eqref{eq:PV ID 4}, the second equality holds by noting that $R_W\indep (R_X, Y)\given (A, W, X)$, and the third equality holds by \eqref{eq:PV ID 3}. We continue with \eqref{eq:2 pv id pro}, 
\begin{align}
    g(X, a')
    & = \EEob\sbr{h_2(a', W, X)\given X}\nend
    & = \EEob\sbr{\EEob\sbr{h_2(A, W, X)\given U, A=a', X}\given X} \nend
    &\aseq\EEob\sbr{\EEob\sbr{Y\given U, A=a', X}\given X},\label{eq:4 pv id pro}
\end{align}
where the first equality holds by noting that $R_X\indep W\given X$, the second equality holds by noting that $W\indep A\given (U, X)$, and the last equality holds by \eqref{eq:3 pv id pro}.
By definition of the CATE, we have
\begin{align}
    \gstar(x, a)
    &=\EEob\sbr{Y\given X=x, \doopt(a)}\nend
    &=\EEob\sbr{\EEob\sbr{Y\given X, U, A=a}\given X=x}.\label{eq:5 pv id pro}
\end{align}
Combining \eqref{eq:4 pv id pro} and \eqref{eq:5 pv id pro}, we conclude with 
\begin{align*}
    g(x, a)\aseq \gstar(x, a),
\end{align*}
which finishes the proof of Theorem \ref{thm:PV ID} that if $h_1, h_2, h_3$ and $g$ exist, $g$ recovers the CATE almost surely. 
\end{proof}

\subsection{Proof of Remark \ref{rmk:PV existence}}\label{pro:rmk PV existence}
We give a proof of the “if” and “only if” part in Remark \ref{rmk:PV existence} in this subsection. 
\paragraph{“If" part.} In this part we prove that it suffices for $h_1, h_2, h_3$ and $g$ to exist if the following conditions hold,
\begin{itemize}
    \item[(i)] There exists a solution $h_2$ to $\EEob\sbr{h_2(A, W, X)-Y\given A=a, X=x, U=u}=0$;
    \item[(ii)] For any solution $h_2$ in (i), there exists a solution $h_1$ to \eqref{eq:PV ID 2}. 
    \item[(iii)] For any solution $h_2$ in (i), there exists a solution $h_3$ to \eqref{eq:PV ID 3}.
\end{itemize}
\begin{proof}
Let $\tilde h_2$ be a solution to $\EEob\sbr{h_2(A, W, X)-Y\given A=a, X=x, U=u}$ following condition (i).
Let $\tilde h_1$ be a solution to \eqref{eq:PV ID 2} with $h_2$ substituted by $\tilde h_2$ by condition (ii) and let $\tilde h_3$ be a solution to \eqref{eq:PV ID 3} with $h_2$ substituted by $\tilde h_2$ by condition (iii).
Moreover, we let $\tilde g=\gstar$.
Therefore, we just need to verify that $\eqref{eq:PV ID 1}$ and \eqref{eq:PV ID 4} holds for $\tilde h_1$, $\tilde h_3$, and $\tilde g$.
For \eqref{eq:PV ID 1}, it holds that
\begin{align*}
   & \EEob\sbr{\tilde h_1(Y, A, X, Z)\given A, X, Z, (R_X, R_Z)} \nend
   &= \EEob\sbr{\EEob\sbr{\tilde h_1(Y, A, X, Z)\given A, W, X, Z, (R_W, R_X, R_Z)=\ind}\given A, X, Z, (R_X, R_Z)}\nend
   &=\EEob\sbr{\tilde h_2(A, W, X)\given A, X, Z, (R_X, R_Z)=\ind}\nend
   &=\EEob\sbr{\EEob\sbr{\tilde h_2(A,W, X)\given A, X, U}\given A, X, Z, (R_X, R_Z)=\ind},
\end{align*}
where the first equality holds by noting that $R_W\indep (R_X, R_Z, Y, Z)\given (A, W, X)$, the second equality holds by noting that $\tilde h_2$ and $\tilde h_1$ satisfy \eqref{eq:PV ID 2}, and the last equality holds by noting that $W\indep (Z, R_X, R_Z)\given (A, X, U)$. Following condition (i), we thus have
\begin{align*}
    & \EEob\sbr{\tilde h_1(Y, A, X, Z)\given A, X, Z, (R_X, R_Z)}\nend
    &=\EEob\sbr{\EEob\sbr{Y\given A, X, U}\given A, X, Z, (R_X, R_Z)=\ind}\nend
    &=\EEob\sbr{\EEob\sbr{Y\given A, X, U, Z, (R_X, R_Z)=\ind}\given A, X, Z, (R_X, R_Z)=\ind}\nend
    &=\EEob\sbr{Y\given A, X, Z, (R_X, R_Z)=\ind},
\end{align*}
where the second equality holds by noting that $Y\indep (Z, R_X, R_Z)\given (A, X, U)$. Therefore, we verify that $\tilde h_1$ satisfies \eqref{eq:PV ID 1}. It remains to check for \eqref{eq:PV ID 4}. We have for $\tilde h_3$ that
\begin{align}
    &\EEob\sbr{\tilde h_3(Y, A, X, a')\given X, R_X=1}\nend
    &=\EEob\sbr{\EEob\sbr{\tilde h_3(Y, A, X, a')\given A, W, X, (R_W, R_X)=\ind}\given X, R_X=1}\nend
    &=\EEob\sbr{\EEob\sbr{\tilde h_2(a', W, X)\given A, W, X, (R_W, R_X)=\ind}\given X, R_X=1}\nend
    &=\EEob\sbr{\tilde h_2(a', W, X)\given X},\label{eq:6 pv id pro}
\end{align}
where the first equality holds by noting that $R_W\indep (R_X, Y)\given (A, W, X)$, the second equality holds by noting that $\tilde h_2$ and $\tilde h_3$ satisfy \eqref{eq:PV ID 3}, and the last equality holds by noting that $R_X\indep W\given X$. Continuing with \eqref{eq:6 pv id pro}, we have
\begin{align}
    &\EEob\sbr{\tilde h_3(Y, A, X, a')\given X}\nend
    &=\EEob\sbr{\EEob\sbr{\tilde h_2(A, W, X)\given A=a', X, U}\given X}\nend
    &=\EEob\sbr{\EEob\sbr{Y\given A=a', X, U}\given X}\nend
    &=\EEob\sbr{Y\given X, \doopt(a')}, \label{eq:7 pv id pro}
\end{align}
where the first equality holds by noting that $A\indep W\given (X, U)$, the second equality holds by condition (i), and the last equality holds by the definition of do-calculus. Note that the right-hand side of \eqref{eq:7 pv id pro} corresponds to the definition of CATE $\gstar$. Therefore, we verify that $\tilde h_3$ and $\tilde g$ satisfy \eqref{eq:PV ID 4}. The proof in this part suggests that following conditions (i)-(iii), $\tilde h_1, \tilde h_2, \tilde h_3$ and $\tilde g$ are solution to \eqref{eq:PV ID 1}-\eqref{eq:PV ID 4}, i.e., conditions (i)-(iii) are sufficient for a solution to exist. 
\end{proof}

\paragraph{“Only if" part.}
We give a proof that any solution to the IES in Theorem \ref{thm:PV ID} must satisfy the conditions in Remark \ref{rmk:PV existence}.
\begin{proof}
The “only if” part is direct if we plug \eqref{eq:PV ID 2} into \eqref{eq:PV ID 1} and obtain,
\begin{align*}
    &\EEob\sbr{Y\given A, X, Z, (R_X, R_Z)=\ind}\nend
    &=\EEob\sbr{\EEob\sbr{h_1(Y, A, X, Z)\given A, W, X, Z, (R_X, R_Z)=\ind}\given A, X, Z, (R_X, R_Z)=\ind}\nend
    &=\EEob\sbr{\EEob\sbr{h_1(Y, A, X, Z)\given A, W, X, Z, (R_X, R_Z, R_W)=\ind}\given A, X, Z, (R_X, R_Z)=\ind}\nend
    &=\EEob\sbr{h_2(A, W, X)\given A, X, Z, (R_X, R_Z)=\ind},
\end{align*}
where the second inequality holds by noting that $R_W\indep (Y, Z, R_X, R_Z)\given (A, W, X)$.
Moreover, by noting that $(W, Y)\indep (Z, R_X, R_Z)\given (U,A, X)$, it follows that
\begin{align*}
    &\EEob\sbr{Y-h_2(A, W, X)\given A, X, Z, (R_X, R_Z)=\ind}\nend
    &=\EEob\sbr{\EEob\sbr{Y-h_2(A, W, X)\given U, A, X}\given A, X, Z, (R_X, R_Z)=\ind}\nend
    &=\EEob\sbr{\EEob\sbr{Y-h_2(A, W, X)\given U, A, X}\given A, X, Z, R_Z=1}, 
\end{align*}
where the last inequality holds by noting that $R_X$ is only caused by $X$.
Following the PV completeness condition, we thereby have,
\begin{align*}
    \EEob\sbr{Y-h_2(A, W, X)\given U, A, X}\aseq 0, 
\end{align*}
which corresponds to the first condition. The remaining two conditions hold directly by \eqref{eq:PV ID 2} and \eqref{eq:PV ID 3}. Hence, we complete the proof of Remark \ref{rmk:PV existence}.
\end{proof}

\subsection{Proof of Theorem \ref{thm:CCB-PV ID extension}}\label{pro:CCB-PV ID extension}
Under the assumption that the bridge functions $h_1, h_2, h_3$ and $g$ exist, we prove that $g$ is a recovery of the average reward $v^\pie$.
\begin{proof}
Our proof is separated  into two steps.  (i) First,  we prove that $h_2$ satisfies $$\EEob\sbr{h_2(A, W, X)-Y\pie(A\given X, W)\given U, A, X}\aseq 0. $$ (ii) Then in the second step,  we prove that $g\aseq v^\pie. $
From \eqref{eq:PV ID 2} we have
\begin{align}
    &\EEob\sbr{h_1(Y, A, X, Z)\given A, X, Z, (R_X, R_Z)=\ind}\nend
    &=\EEob\sbr{\EEob\sbr{h_1(Y, A, X, Z)\given A, W, X, Z, (R_X, R_Z)=\ind}\given A, X,Z, (R_X, R_Z)=\ind}\nend
    &=\EEob\sbr{\EEob\sbr{h_1(Y, A, X, Z)\given A, W, X, Z, (R_X, R_Z, R_W)=\ind}\given A, X,Z, (R_X, R_Z)=\ind}\nend
    &=\EEob\sbr{h_2(A, W, X)-Y\pie(A\given X, W)\given A, X, Z, (R_X, R_Z)=\ind},\label{eq:PV extend 1}
\end{align}
where the second equality holds by noting that $R_W\indep (Y, Z, R_X, R_Z)\given (A, W, X)$.
By noting that $(R_X, R_Z, Z)\indep (W, Y)\given (A, X, U)$, we further have
\begin{align}
    &\EEob\sbr{h_2(A, W, X)-Y\pie(A\given X, W)\given A, X, Z, (R_X, R_Z)=\ind}\nend
    &=\EEob\sbr{\EEob\sbr{h_2(A, W, X)-Y\pie(A\given X, W)\given A, X, U}\given A, X, Z, (R_X, R_Z)=\ind}. \label{eq:PV extend 2}
\end{align}
Following \eqref{eq:PV ID 1} and combining \eqref{eq:PV extend 1} and \eqref{eq:PV extend 2}, it follows that
\begin{align}
    \EEob\sbr{\EEob\sbr{h_2(A, W, X)-Y\pie(A\given X, W)\given A, X, U}\given A, X, Z, R_Z=1} = 0,\label{eq:PV extend 3}
\end{align}
where the equality holds by noting that $R_X$ is only caused by $X$.
By the PV completeness assumption, \eqref{eq:PV extend 3} implies that
\begin{align}
    \EEob\sbr{h_2(A, W, X)-Y\pie(A\given X, W)\given A, X, U}\aseq 0.\label{eq:PV extend 4}
\end{align}
Here we finish the first step. 

In the following, we prove $g\aseq v^\pie. $ From \eqref{eq:PV ID 4}, we have
\begin{align}
    g(X)&=\EEob\sbr{h_3(Y, A, X)\given X, R_X=1}\nend
    &=\EEob\sbr{\EEob\sbr{h_3(Y, A, X)\given A, W, X, (R_X, R_W)=1}\given X, R_X=1}\nend
    &=\EEob\sbr{\EEob\sbr{\sum_{a'\in\cA}h_2(a', W, X)\given A, W, X, (R_X, R_W)=\ind}\given X, R_X=1},\label{eq:PV extend 5}
\end{align}
where the second equality holds by noting that $R_W\indep (R_X, Y)\given (A, W, X)$ and the last equality holds by \eqref{eq:PV ID 3}.
We can rewrite \eqref{eq:PV extend 5} as
\begin{align}
    g(X)&=\EEob\sbr{\sum_{a'\in\cA}h_2(a', W, X)\given X, R_X=1}\nend
    &=\EEob\sbr{\sum_{a'\in\cA}\EEob\sbr{h_2(a', W, X)\given A, X, U}\given X, R_X=1}\nend
    &=\EEob\sbr{\sum_{a'\in\cA}\EEob\sbr{h_2(A, W, X)\given A=a', X, U}\given X, R_X=1}, \label{eq:PV extend 6}
\end{align}
where the second equality holds by $R_X\indep W\given (A, X, U)$ and the last equality holds by noting that $W\indep A \given (X, U)$.
Plugging \eqref{eq:PV extend 4} into \eqref{eq:PV extend 6}, it follows that
\begin{align*}
    g(x)&\aseq\EEob\sbr{\sum_{a'\in\cA} \EEob\sbr{Y\pie(A\given X, W)\given A=a', X, U}\given X=x, R_X=1}\nend
    &=\EEob\sbr{\sum_{a'\in\cA} \EEob\sbr{Y\given A=a', X, U, W}\pie(a'\given X, W)\given X=x, R_X=1}\nend
    &=v^\pie(x),
\end{align*}
where the second equality holds by noting that $A\indep W\given (X, U)$ and that $R_X$ is only caused by $X$.
The last equality holds by \eqref{def:extended v}, which completes the proof of Theorem \ref{thm:CCB-PV ID extension}.
\end{proof}
\section{Proof of the Main Results}
\subsection{Proof of Theorem \ref{thm:Fast rate}}\label{pro:Fast rate}
In this section, we prove that event $\cE$ holds with probability at least $1-4K\xi$ by Assumption \ref{asp:regularity} in Part \RNum{1}. Then, with Assumptions \ref{asp:Realizability} and \ref{asp:compatibility}, we prove that $\gHstar\in\CI_\cD(e_\cD)$ by showing the upper bound for $\cL_\cD(\vhHstar)$ in Part \RNum{2} and show the upper bound for the projected error $\nbr{\cT(\cdot)}$ for any $\vh\in\CIH(e_\cD)$ in Part \RNum{3}.

\paragraph{Part \RNum{1}. } We prove that event $\cE$ holds with probability at least $1-2K\xi$ with the help of the following two technical lemmas. 
\begin{lemma}[Lemma 11 in \cite{foster2019orthogonal}]\label{lem:1st order localized consistency}
Assume $\sup_{f\in\cF}\|f\|_\infty \le c$ and $f^*\in\cF$. Let $\eta$ be the constant such that
$$\cR_n(\eta;\text{star}(\cF-f^*))\le\eta^2/c$$
Additionally, we assume loss function $l(\cdot,\cdot)$ is $L$-Lipschitz in the first argument. Then with probability at least $1-\delta$ for all $f\in\cF$,
$$|(\EE_n[l(f(x),z)]-\EE_n[l(f^*(x),z)])-(\EE[l(f(x),z)]-\EE[l(f^*(x),z)])| \le L\eta_n(\|f-f^*\|_2+\eta_n),$$
where $\eta_n=\eta+c_0\sqrt{\log(c_1/\delta)/n}$ and $c_0$, $c_1$ are universal constants.
\end{lemma}

\begin{lemma}[Theorem 14.1 in \cite{wainwright2019high}]\label{lem:2nd order localized consistency}
Let $\cG$ be a star-shaped and $b$-uniformly bounded function class and $\eta_n$ be any positive solution of $R_s(\eta;\cG)\le \eta^2/b$. Then for any $t\ge\eta_n + c_0\sqrt{\log(c_1/\delta)/n}$, we have
$$|\|g\|_n^2-\|g\|_2^2| \le \frac{1}{2}\|g\|_2^2+\frac{1}{2}t^2, \quad \forall g\in\cG$$
with probability at least $1-\delta$.
\end{lemma}

In Lemma \ref{lem:1st order localized consistency}, we substitute $f$ for $\alpha_k(\vh, \cY_k)\theta_k(\cZ_k)$ and $f^*$ for a zero-like function. We simply let $l(f(x), z)=f(x)$. For all $\vh\in\CI_{\vH, \cD}(e_\cD)\subseteq\vH$ and $\theta_k\in\Theta_k$, it holds with probability at least $1-\xi$ that 
\begin{align}
    &\abr{\EE_{\cD_k}\sbr{\alpha_k(\vh, \cY_k) \theta_k(\cZ_k)} -\EE_{\cD_k}\sbr{\alpha_k(\vh, \cY_k) \theta_k(\cZ_k)}} \nend
    &\le \eta_k\rbr{\nbr{\alpha_k(\vh, \cY_k)\theta_k(\cZ_k)}_{\mu_k, 2}+\eta_k}\nend
    &\le \eta_k\rbr{L_\alpha\nbr{\theta_k}_{\mu_k, 2}+\eta_k}, \label{eq:2 fr}
\end{align}
where $\eta_k$ bounds the critical radius for $\alpha_k(\vh, \vY_k)\theta_k(\vZ_k)$. Following Lemma \ref{lem:2nd order localized consistency} and substituting $g$ for $\theta_k$, it holds also  with probability at least $1-\xi$ that 
\begin{align}
    \abr{\norm{\theta_k}^2_{\cD_k, 2}-\norm{\theta_k}^2_{\mu_k, 2}}\le \frac 1 2\rbr{\norm{\theta_k}^2_{\mu_k, 2}+\eta_k^2}, \label{eq:1 fr}
\end{align}
where $\eta_k$ also bounds the critical radius for $\theta_k$.
Recall the definition of $\cE$,
\begin{align*}
    \cE &= \Big\{\abr{\EE_{\cD_k}\sbr{\alpha_k(\vh, \cY_k) \theta_k(\cZ_k)} - \EE_{\mu_k}\sbr{\alpha_k(\vh, \cY_k) \theta_k(\cZ_k)}} \le \eta_k\rbr{L_\alpha\nbr{\theta_k}_{\mu_k, 2}+\eta_k},\nend
    &\quad \abr{\norm{\theta_k}^2_{\cD_k, 2}-\norm{\theta_k}^2_{\mu_k, 2}}\le \frac 1 2\rbr{\norm{\theta_k}^2_{\mu_k, 2}+\eta_k^2}, 
    \forall \vh\in \vH, \forall \theta_k\in\Theta_k, \forall k\in\{1, \cdots, K\}\Big\}.
\end{align*}
Combining \eqref{eq:1 fr} and \eqref{eq:2 fr} and taking a union bound over $k\in\{1, \cdots, K\}$, it is straight forward that $\cE$ holds with probability at least $1-2K\xi$.

\paragraph{Part \RNum{2}.}
We prove that $\gHstar\in\CI_\cD(e_\cD)$ if $e_\cD>2\vareH^2+(2L_\alpha^2+5/4)\eta^2$ by showing the upper bound for $\cL_\cD(\vhHstar)$ in this part.
For the empirical loss function with respect to $\vhHstar$, it holds on $\cE$ that
\begin{align*}
    \cL_{k, \cD}(\vhHstar)&=\sup_{\theta_k\in\Theta_k}\EE_{\cD_k}\sbr{\alpha_k(\vhHstar, \cY_k) \theta_k(\cZ_k)}-\frac 1 2 \norm{\theta_k}_{\cD_k, 2}^2\nend
    &\le \sup_{\theta_k\in\Theta_k} \Big\{\abr{\EE_{\cD_k}\sbr{\alpha_k(\vhHstar, \cY_k) \theta_k(\cZ_k)} - \EE_{\mu_k}\sbr{\alpha_k(\vhHstar, \cY_k) \theta_k(\cZ_k)}}\nend
    &\quad + \frac 1 2 \abr{\norm{\theta_k}^2_{\cD_k, 2}-\norm{\theta_k}^2_{\mu_k, 2}}
    + \EE_{\mu_k}\sbr{\alpha_k(\vhHstar, \cY_k) \theta_k(\cZ_k)} - \frac 1 2 \norm{\theta_k}^2_{\mu_k, 2}\Big\}\nend
    &\overset{\cE}{\lesssim} \sup_{\theta_k\in\Theta_k} \Big\{\eta_k\rbr{L_\alpha\norm{\theta_k}_{\mu_k, 2}+\eta_k} + \frac 1 4\rbr{\norm{\theta_k}^2_{\mu_k, 2}+\eta_k^2} \nend 
    &\quad + \EE_{\mu_k}\sbr{\alpha_k(\vhHstar, \cY_k) \theta_k(\cZ_k)} - \frac 1 2 \norm{\theta_k}^2_{\mu_k, 2}\Big\}, 
\end{align*}
where the last inequality holds by the definition of $\cE$. Let $\cL^{\lambda}_k(\cdot)=\sup_{\theta_k\in\Theta_k}\EE_{\mu_k}\sbr{\alpha_k(\cdot, \cY_k) \theta_k(\cZ_k)} - \lambda \norm{\theta_k}^2_{\mu_k, 2}$. It then holds for $\cL_{k, \cD}(\vhHstar)$ that
\begin{align}
   \cL_{k, \cD}(\vhHstar) \le \cL^{1/8}_k(\vhHstar) - \inf_{\theta_k\in\Theta_k}\rbr{\frac 1 8   \norm{\theta_k}^2_{\mu_k, 2} - L_\alpha\eta_k\norm{\theta_k}_{\mu_k, 2}} + \frac 5 4\eta_k^2,\label{eq:3 fr}
\end{align}
We further let $\theta_k^\lambda(\cdot;\vh)=\arg\sup_{\theta_k\in\Theta_k}\cL^\lambda_k(\vh)$.
To bridge $\cL_k^{1/8}$ to $\cL_k^{1/2}$, we first study the scaling property of $\cL_k^\lambda$.
For $\cL^{\lambda_1}_k(\vh)$ and $\cL^{ \lambda_2}_k(\vh)$ where $0<\lambda_1\le \lambda_2$, it holds that
\begin{align}
    \cL^{\lambda_2}_k(\vh) &=\sup_{\theta_k\in\Theta_k}\EE_{\mu_k}\sbr{\alpha_k(\vh, \cY_k) \theta_k(\cZ_k)}-\lambda_2 \norm{\theta_k}_{\mu_k, 2}^2\nend
    &=\frac{\lambda_2}{\lambda_1}\cdot \sup_{\theta_k\in\Theta_k}\cbr{\EE_{\mu_k}\sbr{\frac{\lambda_1}{\lambda_2}\alpha_k(\vh, \cY_k) \theta_k(\cZ_k)}-\lambda_1 \norm{\theta_k}_{\mu_k, 2}^2}\nend
    &\ge \frac{\lambda_2}{\lambda_1}\cdot \rbr{\EE_{\mu_k}\sbr{\frac{\lambda_1}{\lambda_2}\alpha_k(\vh, \cY_k) \cdot \frac{\lambda_1}{\lambda_2}\theta_k^{\lambda_1}( \cZ_k;\vh)}-\lambda_1 \nbr{\frac{\lambda_1}{\lambda_2}\theta_k^{\lambda_1}(\cZ_k;\vh)}_{\mu_k, 2}^2}\nend 
    &\ge \frac{\lambda_1}{\lambda_2} \cL^{ \lambda_1}_k(\vh),\label{eq:4 fr}
\end{align}
where the first inequality holds by letting $\theta_k = \lambda_1\theta_k^{\lambda_1}/\lambda_2$ where $\lambda_1\theta_k^{\lambda_1}/\lambda_2\in\Theta_k$ is guaranteed by noting that $\lambda_1\le\lambda_2$ and that $\Theta_k$ is star-shaped.
Plugging \eqref{eq:4 fr} into \eqref{eq:3 fr}, we see that
\begin{align}
    \cL_{k, \cD}(\vhHstar)
    &\overset{\cE}{\lesssim} 4\cL^{1/2}_k(\vhHstar) - \inf_{\theta_k\in\Theta_k}\rbr{\frac 1 8   \norm{\theta_k}^2_{\mu_k, 2} - L_\alpha\eta_k\norm{\theta_k}_{\mu_k, 2}} + \frac 5 4 \eta_k^2\nend 
    &\le 4\cL^{1/2}_k(\vhHstar) + 
    \rbr{2L_\alpha^2 + \frac 5 4} \eta_k^2, \label{eq:5 fr}
\end{align}
where the second inequality holds by a simple calculation of the infimum. 
Summing up \eqref{eq:5 fr} for each $k$, we have
\begin{align}
    \cL_\cD(\vhHstar)
    &\overset{\cE}{\lesssim} \sum_{k=1}^K 4\cL_k(\vhHstar) + \rbr{2 L_\alpha^2+\frac 5 4}\sum_{k=1}^K\eta_k^2\nend
    &\le 2\nbr{\vT\vhHstar}_{2, \mu}^2+ \rbr{2 L_\alpha^2+\frac 5 4}\eta^2\nend
    & \le 2\vareH^2 + \rbr{2 L_\alpha^2+\frac 5 4}\eta^2,\label{eq:fr 3}
\end{align}
where the second equality holds by noting that
\begin{align*}
    \cL_k(\vhHstar)&=\sup_{\theta_k\in\Theta_k}\EE_{\mu_k}\sbr{\alpha_k(\vh(X), \vY_k) \theta_k(\vZ_k)}-\frac 1 2 \norm{\theta_k}_{\mu_k, 2}^2\nend
    &\le \sup_{\theta_k}\EE_{\mu_k}\sbr{\alpha_k(\vh(X), \vY_k) \theta_k(\vZ_k)}-\frac 1 2 \norm{\theta_k}_{\mu_k, 2}^2\nend
    &=\frac 1 2 \nbr{\cT_k \vh}_{\mu_k, 2}^2.
\end{align*}
The third inequality in \eqref{eq:fr 3} holds by Assumption \ref{asp:Realizability} of the realizability error.
By the nonnegtivity of the metric $\cL_{k, \cD}(\cdot)$, it follows on event $\cE$ that
\begin{align*}
    \cL_{k, \cD}(\vhHstar) - \inf_{\vh\in\vH}\cL_{k, \cD}(\vh)
    \overset{\cE}{\lesssim}
     4\cL^{1/2}_k(\vhHstar)+ \rbr{2 L_\alpha^2+\frac 5 4}\eta_k^2,
\end{align*}
and that
\begin{align*}
    \cL_{\cD}(\vhHstar) - \inf_{\vh\in\vH}\cL_{\cD}(\vh) 
    &\overset{\cE}{\lesssim} 2\vareH^2 + \rbr{2 L_\alpha^2+\frac 5 4}\eta^2. 
\end{align*}
Therefore, by definition of the confidence set in \eqref{def:CI}, with $e_\cD>2\vareH^2+(2L_\alpha^2+5/4)\eta^2$, it holds on $\cE$ that $\vhHstar\in\CI_{\vH, \cD}(e_\cD)$ and thereby $\gHstar\in\CICATE(e_\cD)$.

\paragraph{Part \RNum{3}.}
We derive the upper bound for $\cL_\cD(\vh)$ where $\vh\in\CIH(e_\cD)$ in this part.
We first give the lower bound on the empirical loss $\cL_{k,\cD}(\vh)$. 
It holds for all $\vh\in\CI_{\cD}(e_\cD)$ that
\begin{align} \label{eq:L_D(h_CI)-LB-1}
    \cL_{k,\cD_k}(\vh) 
    &=\sup_{\theta_k\in\Theta_k}\EE_{\cD_k}\sbr{\alpha_k(\vh, \cY_k) \theta_k(\cZ_k)}-\frac 1 2 \norm{\theta_k}_{\cD_k, 2}^2\nend
    &\ge \sup_{\theta_k\in\Theta_k} \Big\{-\abr{\EE_{\cD_k}\sbr{\alpha_k(\vh, \cY_k) \theta_k(\cZ_k)} -\EE_{\mu_k}\sbr{\alpha_k(\vh, \cY_k) \theta_k(\cZ_k)}} \nend
    &\quad -\frac 1 2 \abr{\norm{\theta_k}^2_{\cD_k, 2}-\norm{\theta_k}^2_{\mu_k, 2}} + \EE_{\mu_k}\sbr{\alpha_k(\vh, \cY_k) \theta_k(\cZ_k)} - \frac 1 2 \nbr{\theta_k}_{\mu_k, 2}^2\Big\}\nend
    &\overset{\cE}{\gtrsim} \sup_{\theta_k\in\Theta_k} \Big\{- \eta_k\rbr{L_\alpha\nbr{\theta_k}_{\mu_k, 2}+\eta_k} - \frac 1 4\rbr{\norm{\theta_k}^2_{\mu_k, 2}+\eta_k^2}\nend
    &\quad + \EE_{\mu_k}\sbr{\alpha_k(\vh, \cY_k) \theta_k(\cZ_k)} - \frac 1 2 \nbr{\theta_k}_{\mu_k, 2}^2\Big\}.
\end{align}
where the second equality holds by the definition of $\cE$.
Let $\Theta_k^+(\vh)=\{\theta_k\in\Theta_k: \EE_{\mu_k}\sbr{\alpha_k(\vh, \vY_k) \theta_k(\vZ_k)}>0\}$. For any $\theta_k^+\in\Theta_k^+(\vh)$, suppose that $\EE_{\mu_k}\sbr{\alpha_k(\vh, \vY_k) \theta_k^+(\vZ_k)}=\beta \nbr{\theta_k^+}_{\mu_k, 2}^2$. By definition of $\theta_k^+$, we have that $\beta > 0$.
We let $0<\kappa\le 1$. Note that $\Theta_k$ is star-shaped, and it follows that $\kappa \theta_k^+\in\Theta_k$. Therefore, by plugging in $\kappa \theta_k^+$ in \eqref{eq:L_D(h_CI)-LB-1},  we have for any $\vh\in\CI_{\vH, \cD}(e_\cD)$ and $\theta_k^+\in\Theta_k^+(\vh)$ that
\begin{align}\label{eq:L_D(h_CI)-LB-2}
    \cL_{k,\cD_k}(\vh) 
    &\overset{\cE_{k, 3}}{\gtrsim} \kappa\rbr{\beta-\frac 3 4 \kappa}\nbr{\theta_k^+}_{\mu_k, 2}^2 - \eta_k L_\alpha \kappa\nbr{\theta_k^+}_{\mu_k, 2} -  \frac 5 4\eta_k^2.
\end{align}
Recall the definition of the confidence set $\CIH(e_\cD)$. For any $\vh\in\CIH(e_\cD)$, it holds that
\begin{align}\label{eq:19}
    \cL_{\cD}(\vh)
    &\le \inf_{\vh\in\vH}\cL_{\cD}(\vh) + e_\cD\nend
    &\le \cL_{\cD}(\vhHstar) + e_\cD\nend
    &\overset{\cE}{\lesssim} 2\vareH^2 + \rbr{2 L_\alpha^2+\frac 5 4}\eta^2 + e_\cD,
\end{align}
where the second inequality holds by noting that $\vhHstar\in\vH$ and the last inequality follows from \eqref{eq:fr 3}.  By noting that $\cL_{k, \cD_k}(\cdot)\le \cL_\cD(\cdot)$, we can substitute the left-hand side of \eqref{eq:L_D(h_CI)-LB-2} by \eqref{eq:19} and obtain
\begin{align}\label{eq:quadratic inequation}
    \kappa\rbr{\beta-\frac 3 4 \kappa}\nbr{\theta_k^+}_{\mu_k, 2}^2 - \eta_k L_\alpha \kappa\nbr{\theta_k^+}_{\mu_k, 2} - \Delta_{k, \cD} \overset{\cE}{\lesssim} 0,
\end{align}
where 
\begin{align}
    \Delta_{k, \cD}= \frac 5 4\eta_k^2+2\vareH^2 + \rbr{2 L_\alpha^2+\frac 5 4}\eta^2 + e_\cD. \label{def:Delta}
\end{align} 
Note that \eqref{eq:quadratic inequation} holds for any $0<\kappa\le 1$.
By letting $\kappa = \min\{1, \beta\}$, we see that $\beta-3\kappa/4>0$. by solving the quadratic inequality in \eqref{eq:quadratic inequation}, it holds on event $\cE$ for all $\vh\in\CI_{\vH, \cD}(e_\cD)$ and $\theta_k^+\in\Theta_k^+(\vh)$ that
\begin{align}
    \nbr{\theta_k^+}_{\mu_k, 2}\overset{\cE}{\lesssim} \frac{\sqrt{\rbr{\eta_k L_\alpha \kappa}^2+\kappa\rbr{4\beta-3\kappa}\Delta_{k, \cD}}+\eta_k L_\alpha \kappa}{\kappa\rbr{2\beta-\frac 3 2 \kappa}}.\label{eq:fr 2}
\end{align}
We consider the following two cases.
\paragraph{Case (i) where $\beta\ge 1$.} If $\beta \ge 1$, we just plug in $\kappa=\min\{\beta, 1\}=1$. it holds on $\cE$ that
\begin{align}
    \nbr{\theta_k^+}_{\mu_k, 2}^2 &\overset{\cE}{\lesssim}
    \rbr{\frac{\sqrt{\rbr{\eta_k L_\alpha }^2+\rbr{4\beta-3}\Delta_{k, \cD}}+\eta_k L_\alpha }{\rbr{2\beta-\frac 3 2}}}^2\nend
    &\le 4\rbr{\sqrt{\rbr{\eta_k L_\alpha}^2+\Delta_{k, \cD}}+\eta_k L_\alpha}^2\nend
    &\le 8\rbr{2\rbr{\eta_k L_\alpha}^2+\Delta_{k, \cD}},\label{eq:7 fr}
\end{align}
where the first inequality holds by \eqref{eq:fr 2}, the second inequality holds by noting that $\beta=1$ will maximize the right-hand side.
Rearranging \eqref{eq:quadratic inequation} with $\kappa =1 $, we have on event $\cE$ that
\begin{align}
    \EE_{\mu_k}\sbr{\alpha_k(\vh, \vY_k) \theta_k^+(\vZ_k)} &= \beta \nbr{\theta_k^+}_{\mu_k, 2}^2\nend
    &\overset{\cE}{\lesssim} \frac 3 4 \nbr{\theta_k^+}_{\mu_k, 2}^2 + \eta_k L_\alpha \nbr{\theta_k^+}_{\mu_k, 2} + \Delta_{k, \cD}\nend
    &\le 16\rbr{\eta_k L_\alpha}^2+ 9\Delta_{k, \cD}, \label{eq:8 fr}
\end{align}
where the last inequality holds by the upper bound of $\nbr{\theta_k^+}_{\mu_k, 2}$ in \eqref{eq:7 fr}.
\paragraph{Case (ii) where $\beta<1$. } If $\beta < 1$, we just plug in $\kappa=\beta$. It holds on $\cE$ that
\begin{align*}
    \beta \nbr{\theta_k^+}_{\mu_k, 2} \overset{\cE}{\lesssim} 2\rbr{\sqrt{\rbr{\eta_k L_\alpha \kappa}^2+\Delta_{k, \cD}}+\eta_k L_\alpha}\le 2\rbr{2\eta_k L_\alpha + \sqrt{\Delta_{k, \cD}}},
\end{align*}
which suggests that 
\begin{align}
    \EE_{\mu_k}\sbr{\alpha_k(\vh, \vY_k) \theta_k^+(\vZ_k)} =  \beta \nbr{\theta_k^+}_{\mu_k, 2}^2 \overset{\cE}{\lesssim} 2\rbr{\sqrt{\Delta_{k, \cD}}+2\eta_k L_\alpha} \nbr{\theta_k^+}_{\mu_k, 2}.\label{eq:9 fr}
\end{align}
\paragraph{Combination of Case (i) and Case (ii).}
Combining \eqref{eq:8 fr} and \eqref{eq:9 fr} in these two cases, we then have on event $\cE$ for any $\vh\in\CI_{\vH, \cD}(e_\cD)$ and $\theta_k^+\in \Theta_k^+(\vh)$ that
\begin{align*}
    \EE_{\mu_k}\sbr{\alpha_k(\vh, \vY_k) \theta_k^+(\vZ_k)} &\overset{\cE}{\lesssim} \max\cbr{2\rbr{\sqrt{\Delta_{k, \cD}}+2\eta_k L_\alpha} \nbr{\theta_k}_{\mu_k, 2}, 16\rbr{\eta_k L_\alpha}^2+ 9\Delta_{k, \cD}}\nend
    &=\max\cbr{C_\cD \nbr{\theta_k}_{\mu_k, 2}, C_\cD^2},
\end{align*}
where $C_\cD = 4 \eta_k L_\alpha+ 3\sqrt{\Delta_{k, \cD}}$.
Considering the fact that $\EE_{\mu_k}\sbr{\alpha_k(\vh, \vY_k) \theta_k(\vZ_k)}<0$ for $\theta_k\in\Theta_k\backslash \Theta_k^+(\vh)$, it follows for any $\vh\in\CIH(e_\cD)$, $\theta_k\in\Theta_k$ and $k\in\{1, \cdots, K\}$ that
\begin{align}
    \EE_{\mu_k}\sbr{\alpha_k(\vh, \vY_k) \theta_k(\vZ_k)} \overset{\cE}{\lesssim} \max\cbr{C_\cD \nbr{\theta_k}_{\mu_k, 2}, C_\cD^2}.\label{eq:fr 1}
\end{align}
Recall the definition of the linear operator $\cT_k$ in \eqref{def:cT}. We let $\thetaTstark{k}(\vZ_k;\vh)=\arg\min_{\theta_k\in\Theta_k} \nbr{\theta_k-\cT_k\vh}_{\mu_k, 2}$. By \eqref{eq:fr 1}, it then holds for $h\in\CI_{\vH, \cD}(e_\cD)$ that
\begin{align}
    \EE_{\mu_k}\sbr{\cT_k\vh(\vZ_k) \thetaTstark{k}(\vZ_k;\vh)} &\overset{\cE}{\lesssim}
    \max\cbr{C_\cD \nbr{\thetaTstark{k}(\vh)}_{\mu_k, 2}, C_\cD^2}.\label{eq:fr 5}
\end{align}
For the left-hand side of \eqref{eq:fr 5}, we have
\begin{align}
    \EE_{\mu_k}\sbr{\cT_k\vh(\vZ_k)\thetaTstark{k}(\vZ_k;\vh)}
    &=\EE_{\mu_k}\sbr{\cT_k\vh(\vZ_k)\rbr{\thetaTstark{k}(\vZ_k;\vh)-\cT_k\vh(\vZ_k)+\cT_k\vh(\vZ_k)}}\nend
    &\ge \nbr{\cT_k\vh}_{\mu_k, 2}^2 - \nbr{\cT_k\vh}_{\mu_k, 2}\nbr{\thetaTstark{k}(\vh)-\cT_k\vh}_{\mu_k, 2}\nend
    &\ge \nbr{\cT_k\vh}_{\mu_k, 2}^2 - \nbr{\cT_k\vh}_{\mu_k, 2}\vareTheta, \label{eq:fr 6}
\end{align}
where the first inequality holds by the  Cauchy-Schwartz inequality and the last equality holds by Assumption \ref{asp:compatibility} on the compatibility of the dual function class $\Theta_k$.
For the right-hand side of \eqref{eq:fr 5}, we have,
\begin{align}
    \max\cbr{C_\cD \nbr{\thetaTstark{k}(\vh)}_{\mu_k, 2}, C_\cD^2} 
    &\le C_\cD \rbr{C_\cD + \nbr{\thetaTstark{k}(h)-\cT_k\vh+\cT_k\vh}_{\mu_k, 2}}\nend
    &\le C_\cD\rbr{C_\cD+\nbr{\cT_k\vh}_{\mu_k, 2}+\vareTheta},  \label{eq:fr 7}
\end{align}
where the last inequality holds by the triangular inequality and Assumption \ref{asp:compatibility}.
Combining \eqref{eq:fr 6} and \eqref{eq:fr 7} with \eqref{eq:fr 5}, for all $\vh\in\CI_{\vH, \cD}(e_\cD)$ and $k\in\{1, \cdots, K\}$, it holds on event $\cE$ that
\begin{align*}
    \nbr{\cT_k\vh}_{\mu_k, 2}^2 - (\vareTheta+C_\cD) \nbr{\cT_k\vh}_{\mu_k, 2}-C_\cD(C_\cD+\vareTheta) \overset{\cE}{\lesssim} 0, 
\end{align*}
which gives that
\begin{align*}
    \nbr{\cT_k\vh}_{\mu_k, 2} &\overset{\cE}{\lesssim} \frac 1 2 \rbr{\sqrt{(\vareTheta+C_\cD)^2+4C_\cD(C_\cD+\vareTheta)}+(\vareTheta+C_\cD)}\nend
    &\le \vareTheta+2C_\cD\nend
    &= \vareTheta+ 8\eta_k L_\alpha + 6\sqrt{\frac 5 4\eta_k^2+2\vareH^2 + \rbr{2 L_\alpha^2+\frac 5 4}\eta^2 + e_\cD}\nend
    & = O(\vareTheta) + O(\vareH) + O\rbr{\sqrt{e_\cD}} + O\rbr{\eta} , 
\end{align*}
where the first equality holds by definition of $C_\cD$ and the definition of $\Delta_{k, \cD}$ in \eqref{def:Delta}. The last inequality holds by noting that $\eta_k\le \eta$.
Hence, we complete the proof of Theorem \ref{thm:Fast rate}.

\subsection{Decomposition of the Sub-optimality with Pessimism}\label{proof: pessimism}
In this section, we study the sub-optimality of the estimated policy $\piepessi$ with pessimism.
The result in this section will be utilized in \S\ref{proof: subopt of IV} for the proof of sub-optimality of CCB-IV in Theorem \ref{thm:IV subopt}, \S\ref{proof: subopt of PV} for the proof of sub-optimality of CCB-PV in Theorem \ref{thm:PV subopt}, and \S\ref{proof:extended PV Subopt} for the proof of sub-optimality of CCB-PV with extended policy class in Theorem \ref{thm:extended PV subopt}.
Recall the definition of $\gpessi{\pi}$, 
\begin{align}
    \gpessi{\pi}=\arg\inf_{g\in\CICATE(e_\cD)} v(g, \pi),\label{eq:gpessi}
\end{align}
and the definition of $\piepessi$, 
\begin{align}
    \piepessi = \arg\sup_{\pie\in\Pie} \inf_{g\in\CICATE(e_\cD)} v(g, \pi) =  \arg\sup_{\pie\in\Pie} v(\gpessi{\pie}, \pi). \label{eq:piepessi}
\end{align}
On the event $\cE$ defined by \eqref{def:cE}, the regret of policy $\piepessi$ is given by
\begin{align}\label{eq:regret}
    \text{SubOpt}(\piepessi) &=v^{\piestar}-v^{\piepessi} \nend
    &= \underbrace{v^{\piestar} - v(\gpessi{\piestar}, \piestar)}_{\text{(i)}} + \underbrace{v(\gpessi{\piestar}, \piestar) - v(\gpessi{\piepessi}, \piepessi)}_{\text{(ii)}} + \underbrace{v(\gpessi{\piepessi}, \piepessi) - v(\gHstar, \piepessi)}_{\text{(iii)}} + \underbrace{v(\gHstar, \piepessi) - v^{\piepessi}}_{\text{(iv)}}\nend
    &\overset{\cE}{\lesssim} \underbrace{v^{\piestar} - v(\gpessi{\piestar}, \piestar)}_{\text{(i)}} + \underbrace{v(\gHstar, \piepessi) - v^{\piepessi}}_{\text{(iv)}}, 
\end{align}
where $\overset{\cE}{\lesssim}$ means that the inequality holds on event $\cE$ defined in  \eqref{def:cE}. 
Here, $\text{(ii)}\le 0$ holds by definition of $\piepessi$ in \eqref{eq:piepessi} and $\text{(iii)}\overset{\cE}{\lesssim} 0$ holds by definition of $\gpessi{\pie}$ in \eqref{eq:gpessi} and the fact that $\gHstar\in\CICATE(e_\cD)$ on event $\cE$ by Theorem \ref{thm:Fast rate}. Moreover, we show that (iv) is bounded by,
\begin{align}
    \text{(iv)} &= v(\gHstar, \piepessi) - v^{\piepessi}\nend
    & = \int_{\cX\times\cA}\rbr{\gHstar(x, a) - \gstar(x, a)}\tpr(x)\piepessi(a\given x)\rd x\rd a\nend
    & \le \sup_{v\in\cV} \nbr{\gHstar-\gstar}_{v, 2}\le \vareH,\label{eq:iv upper}
\end{align}
where the first inequality holds by the definition of $\cV$ that $\cV=\{v: v(x, a)=\tpr(x)\pie(a\given x), \forall \pie\in\Pie\}$ and the last inequality holds by Assumption \ref{asp:Realizability} on the realizability of the hypothesis class.
Therefore, we just need to bound (i). The upper bound for (i) in CCB-IV and CCB-PV are given in \S\ref{proof: subopt of IV} and \S\ref{proof: subopt of PV}, respectively.

\subsection{Proof of Theorem \ref{thm:IV subopt}} \label{proof: subopt of IV}
\begin{proof}
In this section, we study the estimation error of the average reward function with respect to the optimal interventional policy $\piestar$ under the CCB-IV setting, i.e., term (i) in \eqref{eq:regret}. The error in the estimated CATE is given by
\begin{align*}
    &\EEob\sbr{g^*(A, X)-\gpessi{\piestar}(A, X)\given Z, R_Z=1}\nend
    &=\EEob\sbr{\EEob\sbr{g^*(A, X)-\gpessi{\piestar}(A, X)\given A, X, Z, (R_X, R_Z)=\ind}\given Z, R_Z=1} \nend
    &=\EEob\sbr{\EEob\sbr{\hstark{1}(Y, A, Z)-\gpessi{\piestar}(A, X)\given A, X, Z, (R_X, R_Z)=\ind}\given Z, R_Z=1} \nend
    &=\EEob\sbr{\EEob\sbr{\hstark{1}(Y, A, Z)-\hpessik{1}{\piestar}(Y, A, Z) \given A, X, Z, (R_X, R_Z)=\ind}\given Z, R_Z=1} \nend
    &\quad +\EEob\sbr{\EEob\sbr{\hpessik{1}{\piestar}(Y, A, Z) - \gpessi{\piestar}(A, X)\given A, X, Z, (R_X, R_Z)=\ind}\given Z, R_Z=1},
\end{align*}
where the second equality holds by \eqref{eq:IV bridge 2} with $h_1, g$ substituted by the optimal bridge functions $\hstark{1}, g^*$. Note that $R_X\indep (Y, R_Z)\given (A, X, Z)$, it then follows that
\begin{align}
    &\EEob\sbr{g^*(A, X)-\gpessi{\piestar}(A, X)\given Z, R_Z=1}\nend
    &=\EEob\sbr{\hstark{1}(Y, A, Z)-\hpessik{1}{\piestar}(Y, A, Z)\given Z, R_Z=1} \nend
    &\quad +\EEob\sbr{\EEob\sbr{\hpessik{1}{\piestar}(Y, A, Z) - \gpessi{\piestar}(A, X)\given A, X, Z, (R_X, R_Z)=\ind}\given Z, R_Z=1},\nend
    &=\EEob\sbr{Y-\hpessik{1}{\piestar}(Y, A, Z)\given Z, R_Z=1}\nend
    &\quad
    +\EEob\sbr{\EEob\sbr{\hpessik{1}{\piestar}(Y, A, Z) - \gpessi{\piestar}(A, X)\given A, X, Z, (R_X, R_Z)=\ind}\given Z, R_Z=1}\nend
    & = -\cT_1\vhpessi{\piestar}(Z) -  \EEob\sbr{\cT_2\vhpessi{\piestar}(A, X, Z) \given Z, R_Z=1}, 
    \label{eq:IV 1}
\end{align}
where the second equality holds by plugging in \eqref{eq:IV bridge 1} for the optimal bridge function $\hstark{1}$. 
A change of base distribution in term (i) of the sub-optimality \eqref{eq:regret} gives
\begin{align}
    &v^{\piestar} - v(\gpessi{\piestar}, \piestar) \nend
    & = \EEob\sbr{\rbr{\gstar(A, X)-\gpessi{\piestar}(A, X)}\frac{\tpr(X)\piestar(A\given X)}{\pob(X, A\given R_Z=1)}\given R_Z=1}.\label{eq:IV 2}
\end{align}
By assumption of Theorem \ref{thm:IV subopt} that there exists $b_1: \cZ\rightarrow \RR$ satisfying
\begin{align*}
    \EEob\sbr{b_1(Z)\given A, X, R_Z=1} = \frac{\tpr(X)\piestar(A\given X)}{\pob(X, A\given R_Z=1)}, 
\end{align*}
it holds for \eqref{eq:IV 2} that
\begin{align}
    v^{\piestar} - v(\gpessi{\piestar}, \piestar) 
    &=\EEob\sbr{\rbr{\gstar(A, X)-\gpessi{\piestar}(A, X)}\EEob\sbr{b_1(Z)\given A, X, R_Z=1}\given R_Z=1} \nend
    & = \EEob\sbr{\EEob\sbr{\gstar(A, X)-\gpessi{\piestar}(A, X)\given Z, R_Z=1}b_1(Z)\given R_Z=1}\nend
    & = -\EEob\sbr{\rbr{\cT_1\vhpessi{\piestar}(Z) +  \EEob\sbr{\cT_2\vhpessi{\piestar}(A, X, Z) \given Z, R_Z=1}} b_1(Z)\given R_Z=1}\nend
    & = -\EEob\sbr{\cT_1\vhpessi{\piestar}(Z)b_1(Z)\given R_Z=1} \nend
    &\quad - \EEob\sbr{\cT_2\vhpessi{\piestar}(A, X, Z)b_1(Z)\frac{\pob(X, A, Z\given R_Z=1)}{\pob(A, X, Z\given (R_X, R_Z)=\ind)}\given (R_X, R_Z)=\ind}.\label{eq:IV 4}
\end{align}
where the third equality holds by plugging in \eqref{eq:IV 1}.
We define $b_2:\cA\times\cX\times\cZ\rightarrow \RR$ by
\begin{align*}
    b_2(a, x, z)=\frac{b_1(z)\pob(a, x, z\given R_Z=1)}{\pob(a, x, z\given(R_X, R_Z)=\ind)}.
\end{align*}
Hence, using the Cauchy-Schwarz inequality, \eqref{eq:IV 4} is further bounded by
\begin{align}
    v^{\piestar} - v(\gpessi{\piestar}, \piestar)
    &\le \nbr{\cT_1\vhpessi{\piestar}}_{\mu_1, 2} \nbr{b_1}_{\mu_1, 2} + \nbr{\cT_2\vhpessi{\piestar}}_{\mu_2, 2}\nbr{b_2}_{\mu_2, 2}\nend
    &\le \rbr{\nbr{b_1}_{\mu_1, 2}+\nbr{b_2}_{\mu_2, 2}} \max_{k\in\{1, 2\}} \nbr{\cT_k\vhpessi{\piestar}}_{\mu_k, k}\nend
    &\overset{\cE}{\lesssim} \sum_{k=1}^2\nbr{b_k}_{\mu_k, 2} \rbr{O(\vareTheta) + O(\vareH) + O\rbr{\sqrt{e_\cD}} +  O\rbr{\eta}},\label{eq:IV 5}
\end{align}
where the last inequality holds by Theorem \ref{thm:Fast rate}.
Now combining \eqref{eq:iv upper} and \eqref{eq:IV 5} with \eqref{eq:regret}, we arrive at
\begin{align*}
    \SubOpt(\piepessi) 
    &\overset{\cE}{\lesssim} \sum_{k=1}^2 \nbr{b_k}_{\mu_k, 2}\cdot \rbr{O(\vareTheta) + O(\vareH) + O\rbr{\sqrt{e_\cD}} +  O\rbr{\eta}} + \vareH\nend
    &\le \sum_{k=1}^2 \nbr{b_k}_{\mu_k, 2}\cdot \rbr{O(\vareTheta) + O(\vareH) + O\rbr{\sqrt{e_\cD}} +  O\rbr{\eta}}, 
\end{align*}
where the last inequality holds by noting that $\nbr{b_k}^2_{\mu_k, 2}\ge 1$, which follows from the non-negativity of the chi-squared distance, i.e., 
\begin{align*}
    \chi^2(p, \mu)=\EE_\mu\sbr{\frac{p^2}{\mu^2}-1}=\EE_\mu\sbr{\rbr{\frac{p-\mu}{\mu}}^2}\ge 0.
\end{align*}
Hence, we complete the proof of Theorem \ref{thm:IV subopt}.
\end{proof}

\subsection{Proof of Theorem \ref{thm:PV subopt}}\label{proof: subopt of PV}
\begin{proof}
In \eqref{eq:PV ID 1} and \eqref{eq:PV ID 2}, $h_1$ serves as the bridge function to overcome the problem of missingness in $W$, and $h_2$ is the actual bridge function that we care about. Therefore, we study the difference between $\hpessik{2}{\piestar}$ and $\hstark{2}$ by
\begin{align}
    &\EEob\sbr{\hstark{2}(A, W, X)-\hpessik{2}{\piestar}(A, W, X)\given A, X, Z, (R_X, R_Z)=\ind}\nend
    &=\EEob\sbr{\EEob\sbr{\hstark{2}(A, W, X)-\hpessik{2}{\piestar}(A, W, X)\given A, W, X, Z, (R_W, R_X, R_Z)=\ind}\given A, X, Z, (R_X, R_Z)=\ind}\nend
    &=\EEob\sbr{\hstark{1}(Y, A, X, Z)-\hpessik{1}{\piestar}(Y, A, X, Z)\given A, X, Z, (R_X, R_Z)=\ind} \nend
    &\quad + \EEob\sbr{\EEob\sbr{\hpessik{1}{\piestar}(Y, A, X, Z)-\hpessik{2}{\piestar}(A, W, X)\given A, W, X, Z, (R_W, R_X, R_Z)=\ind}\given A, X, Z, (R_X, R_Z)=\ind}\nend
    &=\cT_1\vhpessi{\piestar}(A, X, Z) + \EEob\sbr{\cT_2\vhpessi{\piestar}(A, W, X, Z)\given A, X, Z, (R_X, R_Z)=\ind}, \label{eq:PV 1}
\end{align}
where the second equality holds by \eqref{eq:PV ID 2} which states that $$\EEob\sbr{\hstark{2}(A, W, X) - \hstark{1}(Y, A, X, Z)\given A, W, X, Z, (R_W, R_X, R_Z)=\ind}=0, $$
and noting that $R_W\indep (R_X, R_Z, Z)\given (A, X, W)$. The third equality holds by the definition of the linear operator $\cT_k$.
Now that we have characterized the difference between $\hpessik{2}{\piestar}$ and $\hstark{2}$, it still remains to see the error in the estimated CATE. 
\begin{align*}
    &\gstar(X, A')-\gpessi{\piestar}(X, A')\nend
    &=\EEob\sbr{\rbr{\hstark{3}(Y, A, X; A') - \hpessik{3}{\piestar}(Y, A, X; A')} + \rbr{\hpessik{3}{\piestar}(Y, A, X; A')-\gpessi{\piestar}(X, A')}\given X, A', R_X=1}\nend
    &=\EEob\sbr{\EEob\sbr{\hstark{3}(Y, A, X;A')-\hpessik{3}{\piestar}(Y, A, X; A')\given A, W, X, A', (R_W, R_X)=\ind}\given X, A', R_X=1}\nend
    &\quad - \cT_4\vhpessi{\piestar}(X, A'),
\end{align*}
where the first equality holds by \eqref{eq:PV ID 6} which states that 
$$\EEob\sbr{\gstar(X, A') - \hstark{3}(Y, A, X, A')\given X, A', R_X=1}=0.$$
The second equality holds also by noting that $R_W\indep (Y, R_X)\given(A, X, W)$ and the definition of $\cT_4$ in the CCB-PV case. We continue with \eqref{eq:PV ID 5} which states that
$$\EEob\sbr{\hstark{3}(Y, A, X,  A')-\hstark{2}(A', W, X)\given A, W, X, A', (R_W, R_X)=\ind}=0, $$
and it holds for $\gstar(X, A')-\gpessi{\piestar}(X, A')$ that
\begin{align}
    &\gstar(X, A')-\gpessi{\piestar}(X, A')\nend
    &=\EEob\sbr{\EEob\sbr{\hstark{2}(A', W, X)-\hpessik{2}{\piestar}(A', W, X)\given A, W, X, A', (R_W, R_X)=\ind}\given X, A', R_X=1}\nend
    &\quad + \EEob\sbr{\EEob\sbr{\hpessik{2}{\piestar}(A', W, X)-\hpessik{3}{\piestar}(Y, A, X, A')\given A, W, X, A', (R_W, R_X)=\ind}\given X, A', R_X=1}\nend
    &\quad  - \cT_4\vhpessi{\piestar}(X, A')\nend
    &=\EEob\sbr{\hstark{2}(A', W, X)-\hpessik{2}{\piestar}(A', W, X)\given X, A', R_X=1} \nend
    &\quad -\EEob\sbr{\cT_3\vhpessi{\piestar}(A, W, X, A')\given X, A', R_X=1} -\cT_4\vhpessi{\piestar}(X, A'),\label{eq:PV 2}
\end{align}
where the second equality holds by definition of $\cT_3$ in the CCB-PV case. 
Now we plug \eqref{eq:PV 2} into (i) of \eqref{eq:regret} and it follows that
\begin{align}
    \text{(i)} &= v^{\piestar} - v(\gpessi{\piestar}, \piestar)\nend
    &=\int_{\cX\times\cA} \rbr{\gstar(x, a')-\gpessi{\piestar}(x, a')}\tpr(x)\piestar(a'\given x)\rd x\rd a'\nend
    &=\underbrace{\int_{\cX\times\cA}\EEob\sbr{\hstark{2}(A', W, X)-\hpessik{2}{\piestar}(A', W, X)\given X, A', R_X=1} \tpr(x)\piestar(a'\given x)\rd x\rd a'}_{(a)} \nend
    &\quad \underbrace{-\int_{\cX\times\cA}\EEob\sbr{\cT_3\vhpessi{\piestar}(A, W, X, A')\given X, A', R_X=1} \tpr(x)\piestar(a'\given x)\rd x\rd a'}_{(b)} \nend
    &\quad \underbrace{-\int_{\cX\times\cA}\cT_4\vhpessi{\piestar}(X, A') \tpr(x)\piestar(a'\given x)\rd x\rd a'}_{(c)}.\label{eq:PV 3}
\end{align}
To upper bound (b) and (c), we define two ratio functions $b_4:\cX\times\cA\rightarrow \RR$ and $b_3:\cW\times\cX\times\cA\times\cA'\rightarrow \RR$ by
\begin{gather}
    b_4(x, a') = \frac{\tpr(x)\piestar(a'\given x)}{\pob(x\given R_X=1)u(a')}, \label{eq:PV b4}\\
    b_3(w, x, a, a') = \frac{\tpr(x)\piestar(a'\given x)\pob(a, w\given x, R_X=1)}{u(a')\pob(x, a, w\given (R_W, R_X)=\ind)}.\label{eq:PV b3}
\end{gather}
For (b), with $b_3$ defined in \eqref{eq:PV b3} we have
\begin{align}
    (b) = -\EEob\sbr{\cT_3\vhpessi{\piestar}(A, W, X, A')b_3(W, X, A, A')\given (R_W, R_X)=\ind}\le \nbr{\cT_3\vhpessi{\piestar}}_{\mu_3, 2}\nbr{b_3}_{\mu_3, 2}. \label{eq:PV (b)}
\end{align}
Similarly, for (c) with $b_4$ defined in \eqref{eq:PV b4} we have
\begin{align}
    (c) = \EEob\sbr{-\cT_4\vhpessi{\piestar}(X, A')b_4(X, A')\given R_X=1}\le \nbr{\cT_4\vhpessi{\piestar}}_{\mu_4, 2}\nbr{b_4}_{\mu_4, 2}.\label{eq:PV (c)}
\end{align}
In addition, it holds for (a) that
\begin{align*}
    (a)&=\int_{\cX\times\cA} \EEob\sbr{\hstark{2}(A', W, X)-\hpessik{2}{\piestar}(A', W, X)\given X, A', R_X=1} \tpr(x)\piestar(a'\given x) \rd x\rd a'\nend
    &=\int_{\cX\times\cA} \EEob\sbr{\hstark{2}(a, W, X)-\hpessik{2}{\piestar}(a, W, X)\given X=x, R_X=1} \tpr(x)\piestar(a\given x) \rd x\rd a \nend
    &=\int_{\cX\times\cA}\EEob\sbr{\EEob\sbr{\hstark{2}(A, W, X)-\hpessik{2}{\piestar}(A, W, X)\given X, U, A=a, (R_X, R_Z)=\ind}\given X=x, R_X=1}
    \nend
    &\quad \cdot \tpr(x)\piestar(a\given x) \rd x\rd a, 
\end{align*}
where the second equality holds by noting that $A'\indep W, X$, and the third equality holds by noting that $R_X\indep W\given X$, $A\indep W\given (X, U)$, and $(R_X, R_Z)\indep W\given (A, U, X)$. Now we continue by 
\begin{align}
    (a)
    & = \EEob\bigg[\EEob\sbr{\hstark{2}(A, W, X)-\hpessik{2}{\piestar}(A, W, X)\given X, U, A, (R_X, R_Z)=\ind} \nend
    &\quad \cdot \frac{\pob(U\given X, R_x=1)\tpr(X)\piestar(A\given X)}{\pob(U, X, A\given (R_X, R_Z)=\ind)}\given (R_X, R_Z)=\ind\bigg], \label{eq:PV 4}
\end{align}
which prompts us to introduce another ratio function. 
Since $Z$ is over-complete over $U$, there exists $b_1: \cX\times\cA\times\cZ\rightarrow \RR$ such that
\begin{align}
    \EEob\sbr{b_1(X, A, Z)\given X, U, A, (R_X, R_Z)=\ind} = \frac{\pob(U\given X, R_x=1)\tpr(X)\piestar(A\given X)}{\pob(U, X, A\given (R_X, R_Z)=\ind)}.\label{eq:PV 5}
\end{align}
Plugging \eqref{eq:PV 5} into \eqref{eq:PV 4}, it holds that
\begin{align*}
    (a) 
    &= \EEob\bigg[\EEob\sbr{\hstark{2}(A, W, X)-\hpessik{2}{\piestar}(A, W, X)\given X, U, A, (R_X, R_Z)=\ind}\nend
    & \quad \cdot \EEob\sbr{b_1(X, A, Z)\given X, U, A, (R_X, R_Z)=\ind}\given (R_X, R_Z)=\ind\bigg]\nend
    & = \EEob\sbr{\rbr{\hstark{2}(A, W, X)-\hpessik{2}{\piestar}(A, W, X)}b_1(X, A, Z)\given  (R_X, R_Z)=\ind}\nend
    & = \EEob\sbr{\EEob\sbr{\hstark{2}(A, W, X)-\hpessik{2}{\piestar}(A, W, X)\given X, A, Z, (R_X, R_Z)=\ind} b_1(X, A, Z)\given  (R_X, R_Z)=\ind},
\end{align*}
where the second equality holds by noting that $W\indep (Z, R_Z)\given (X, A, U, R_X)$. Now combining \eqref{eq:PV 1}, we have
\begin{align*}
    (a) &= \EEob\bigg[\rbr{\cT_1\vhpessi{\piestar}(A, X, Z) + \EEob\sbr{\cT_2\vhpessi{\piestar}(A, W, X, Z)\given A, X, Z, (R_X, R_Z)=\ind}}\nend
    &\quad \cdot b_1(X, A, Z)\given (R_X, R_Z)=\ind\bigg]\nend
    &= \EEob\sbr{\cT_1\vhpessi{\piestar}(A, X, Z) b_1(X, A, Z)\given (R_X, R_Z)=\ind}\nend
    &\quad + \EEob\sbr{\EEob\sbr{\cT_2\vhpessi{\piestar}(A, W, X, Z)b_1(X, A, Z)\given A, W, X, (R_W, R_X, R_Z)=\ind}\given (R_X, R_Z)=\ind}\nend
    &= \EEob\sbr{\cT_1\vhpessi{\piestar}(A, X, Z) b_1(X, A, Z)\given (R_X, R_Z)=\ind}\nend
    &\quad + \EEob\sbr{\cT_2\vhpessi{\piestar}(A, W, X, Z)b_1(X, A, Z)\frac{\pob(A, W, X\given (R_X, R_Z)=\ind)}{\pob(A, W, X\given (R_W, R_X, R_Z)=\ind)}\given (R_W, R_X, R_Z)=\ind} ,
\end{align*}
where the second equality holds by $R_W\indep (Z, R_X, R_Z)\given (A, W, X)$.
We thereby define $b_2: \cA\times\cW\times\cX\times\cZ\rightarrow \RR$ by
\begin{align*}
    b_2(A, W, X, Z) = b_1(X, A, Z)\frac{\pob(A, W, X\given (R_X, R_Z)=\ind)}{\pob(A, W, X\given (R_W, R_X, R_Z)=\ind)}.
\end{align*}
We then arrive at
\begin{align}
    (a)&=\EEob\sbr{\cT_1\vhpessi{\piestar}(A, X, Z) b_1(X, A, Z)\given (R_X, R_Z)=\ind}\nend
    &\quad + \EEob\sbr{\cT_2\vhpessi{\piestar}(A, W, X, Z)b_2(A, W, X, Z)\given (R_W, R_X, R_Z)=\ind}\nend
    &\le \nbr{\cT_1\vhpessi{\piestar}}_{\mu_1, 2}\nbr{b_1}_{\mu_1, 2} + \nbr{\cT_2\vhpessi{\piestar}}_{\mu_2, 2}\nbr{b_2}_{\mu_2, 2}, \label{eq:PV (a)}
\end{align}
Combining \eqref{eq:PV (a)}, \eqref{eq:PV (b)} and \eqref{eq:PV (c)} with \eqref{eq:PV 3}, we have
\begin{align*}
    v^{\piestar} - v(\gpessi{\piestar})
    &\le \sum_{k=1}^4 \nbr{\cT_k\vhpessi{\piestar}}_{\mu_k, 2}\nbr{b_k}_{\mu_k, 2}
    \le \max_{k\in\{1, 2, 3,4\}}\nbr{\cT_k\vhpessi{\piestar}}_{\mu_k, 2}  \sum_{k=1}^4 \nbr{b_k}_{\mu_k, 2}.
\end{align*}
Combined with the sub-optimality in \eqref{eq:regret} and the conclusion of Theorem \ref{thm:Fast rate},  it follows that
\begin{align*}
    \SubOpt(\piepessi) &\le \max_{k\in\{1, 2, 3,4\}}\nbr{\cT_k\vhpessi{\piestar}}_{\mu_k, 2}  \sum_{k=1}^K \nbr{b_k}_{\mu_k, 2} + \vareH\nend
    &\overset{\cE}{\lesssim} \sum_{k=1}^4 \nbr{b_k}_{\mu_k, 2} \cdot \big(\cO(\vareTheta) + \cO(\vareH) + \cO\rbr{\sqrt{e_\cD}} +  \cO\rbr{\eta}\big) + \vareH\nend
    &\le \sum_{k=1}^4 \nbr{b_k}_{\mu_k, 2} \cdot \big(\cO(\vareTheta) + \cO(\vareH) + \cO\rbr{\sqrt{e_\cD}} +  \cO\rbr{\eta}\big), 
\end{align*}
which completes the proof of Theorem \ref{thm:PV subopt}.
\end{proof}

\subsection{Sub-optimality for CCB-PV with Extended Policy Class}\label{proof:extended PV Subopt}
For CCP-PV with extended policy class, we define event $\cE$ by
\begin{align}
    \cE &= \Big\{\abr{\EE_{\cD_k}\sbr{\alpha^\pie_k(\vh, \cY_k) \theta_k(\cZ_k)} - \EE_{\mu_k}\sbr{\alpha^\pie_k(\vh, \cY_k) \theta_k(\cZ_k)}} \le \eta_k\rbr{L_\alpha\nbr{\theta_k}_{\mu_k, 2}+\eta_k},\nend
    &\quad \abr{\norm{\theta_k}^2_{\cD_k, 2}-\norm{\theta_k}^2_{\mu_k, 2}}\le \frac 1 2\rbr{\norm{\theta_k}^2_{\mu_k, 2}+\eta_k^2}, 
    \forall \vh\in \vH, \forall \pie\in\Pie, \forall \theta_k\in\Theta_k, \forall k\in\{1, \cdots, K\}\Big\}, \label{def:extended cE}
\end{align}
where $\eta$ bounds the critical radius for $\alpha^\pie_k(\vh, \cY_k) \theta_k(\cZ_k)$ and $\theta_k$.
In contrast to the definition in \eqref{def:cE}, $\eta$ also bounds the critical radius of the policy class $\Pie$.
We remark that Theorem \ref{thm:Fast rate} still holds on the event defined by \eqref{def:extended cE}.
Similarly, we define the pessimistic $\gpessi{\pie}$ by
\begin{align}
    \gpessi{\pie}=\arg\inf_{g\in\CICATE^\pie(e_\cD)} v(g). \label{def:extended gpessi}
\end{align}
The sub-optimality of policy $\piepessi$ is given by
\begin{align}\label{eq:extended regret}
    \text{SubOpt}(\piepessi) &=v^{\piestar}-v^{\piepessi} \nend
    &= \underbrace{v^{\piestar} - v(\gpessi{\piestar})}_{\text{(i)}} + \underbrace{v(\gpessi{\piestar}) - v(\gpessi{\piepessi})}_{\text{(ii)}} + \underbrace{v(\gpessi{\piepessi}) - v(\gH{\piepessi})}_{\text{(iii)}} + \underbrace{v(\gH{\piepessi}) - v^{\piepessi}}_{\text{(iv)}}\nend
    &\overset{\cE}{\lesssim} \underbrace{v^{\piestar} - v(\gpessi{\piestar})}_{\text{(i)}} + \underbrace{v(\gH{\piepessi}) - v^{\piepessi}}_{\text{(iv)}}, 
\end{align}
where $\text{(ii)}\le 0$ holds by the definition of $\piepessi$ in \eqref{def:extended piepessi} and $\text{(iii)} \overset{\cE}{\lesssim} 0$ holds by definition of $\gpessi{\pie}$ in \eqref{def:extended gpessi} and the fact that $\gH{\piepessi}\in\CICATE^\piepessi(e_\cD)$ on event $\cE$ by theorem \ref{thm:Fast rate}. Following \eqref{eq:iv upper}, we have,
\begin{align*}
    \text{(iv)} &= v(\gH{\piepessi}) - v^{\piepessi}\nend
    & = \int_{\cX\times\cA}\rbr{\gH{\piepessi}(X) - g^\piepessi(X)}\tpr(x)\rd x\rd a\nend
    & \le \sup_{\pie\in\Pie}\nbr{\gH{\pie}-g^\pie}_{ \tpr, 2}\le \vareH.
\end{align*}
Similar to \S\ref{proof: subopt of PV}, we give a brief proof to bound (i) in \eqref{eq:extended regret}.
We first study the difference between $\hpessik{2}{\piestar}$ and $\h{2}{\piestar}$ by
\begin{align}
    &\EEob\sbr{\h{2}{\piestar}(A, W, X)-\hpessik{2}{\piestar}(A, W, X)\given A, X, Z, (R_X, R_Z)=\ind}\nend
    &=\EEob\sbr{\EEob\sbr{\h{2}{\piestar}(A, W, X)-\hpessik{2}{\piestar}(A, W, X)\given A, W, X, Z, (R_W, R_X, R_Z)=\ind}\given A, X, Z, (R_X, R_Z)=\ind}\nend
    &=\EEob\sbr{\h{1}{\piestar}(Y, A, X, Z)  -\hpessik{1}{\piestar}(Y, A, X, Z) \given A, X, Z, (R_X, R_Z)=\ind} \nend
    &\quad + \EEob\sbr{\EEob\sbr{\hpessik{1}{\piestar}(Y, A, X, Z)-\hpessik{2}{\piestar}(A, W, X)\given A, W, X, Z, (R_W, R_X, R_Z)=\ind}\given A, X, Z, (R_X, R_Z)=\ind}\nend
    &=- \cT_1\vhpessi{\piestar}(A, X, Z) - \EEob\sbr{\cT_2\vhpessi{\piestar}(A, W, X, Z)\given A, X, Z, (R_X, R_Z)=\ind}, \label{eq:extended PV 1}
\end{align}
which is identical to the proof in \S\ref{proof: subopt of PV} and the equality holds following the same reasons for \eqref{eq:PV 1}. 
For the error in the estimated CATE, we have
\begin{align*}
    &g^\piestar(X)-\gpessi{\piestar}(X)\nend
    &=\EEob\sbr{\rbr{\h{3}{\piestar}(Y, A, X) - \hpessik{3}{\piestar}(Y, A, X)} + \rbr{\hpessik{3}{\piestar}(Y, A, X)-\gpessi{\piestar}(X)}\given X, R_X=1}\nend
    &=\EEob\sbr{\EEob\sbr{\h{3}{\piestar}(Y, A, X)-\hpessik{3}{\piestar}(Y, A, X)\given A, W, X, (R_W, R_X)=\ind}\given X, R_X=1} - \cT_4\vhpessi{\piestar}(X),
\end{align*}
where the first equality holds by \eqref{eq:PV ID extension 4} which states that 
$$\EEob\sbr{g^\piestar(X, A') - \h{3}{\piestar}(Y, A, X, A')\given X, A', R_X=1}=0.$$
The second equality holds also by noting that $R_W\indep (Y, R_X)\given(A, X, W)$ and the definition of $\cT_4$ in the CCB-PV case. We continue with \eqref{eq:PV ID extension 3} which states that
$$\EEob\sbr{\h{3}{\piestar}(Y, A, X)-\sum_{a'\in\cA}\h{2}{\piestar}(a', W, X)\given A, W, X, (R_W, R_X)=\ind}=0, $$
and it holds for $g^\piestar(X)-\gpessi{\piestar}(X)$ that
\begin{align}
    &g^\piestar(X)-\gpessi{\piestar}(X)\nend
    &=\EEob\sbr{\EEob\sbr{\sum_{a'\in\cA}\rbr{\h{2}{\piestar}(a', W, X)-\hpessik{2}{\piestar}(a', W, X)}\given A, W, X, (R_W, R_X)=\ind}\given X, R_X=1}\nend
    &\quad + \EEob\sbr{\EEob\sbr{\sum_{a'\in\cA}\hpessik{2}{\piestar}(a', W, X)-\hpessik{3}{\piestar}(Y, A, X)\given A, W, X, (R_W, R_X)=\ind}\given X, R_X=1}\nend
    &\quad  - \cT_4\vhpessi{\piestar}(X)\nend
    &=\sum_{a'\in\cA}\EEob\sbr{\h{2}{\piestar}(a', W, X)-\hpessik{2}{\piestar}(a', W, X)\given X, R_X=1} \nend
    &\quad -\EEob\sbr{\cT_3\vhpessi{\piestar}(A, W, X)\given X, R_X=1} -\cT_4\vhpessi{\piestar}(X),\label{eq:extended PV 2}
\end{align}
where the second equality holds by definition of $\cT_3$ in the CCB-PV case. 
Now we plug \eqref{eq:extended PV 2} into (i) of \eqref{eq:extended regret} and it follows that
\begin{align}
    \text{(i)} &= v^{\piestar} - v(\gpessi{\piestar})\nend
    &=\int_{\cX} \rbr{g^\piestar(x)-\gpessi{\piestar}(x)}\tpr(x)\rd x\nend
    &=\underbrace{\int_{\cX}\EEob\sbr{\sum_{a'\in\cA}\rbr{\h{2}{\piestar}(a', W, X)-\hpessik{2}{\piestar}(a', W, X)}\given X=x,  R_X=1} \tpr(x)\rd x}_{(a)} \nend
    &\quad \underbrace{-\int_{\cX}\EEob\sbr{\cT_3\vhpessi{\piestar}(A, W, X)\given X=x, R_X=1} \tpr(x)\rd x}_{(b)} \underbrace{-\int_{\cX}\cT_4\vhpessi{\piestar}(x) \tpr(x)\rd x}_{(c)}.\label{eq:extended PV 3}
\end{align}
To upper bound (b) and (c), we define two ratio functions $b_4:\cX\times\cA\rightarrow \RR$ and $b_3:\cW\times\cX\times\cA\times\cA'\rightarrow \RR$ by
\begin{gather}
    b_4(x, a') = \frac{\tpr(x)}{\pob(x\given R_X=1)}, \label{eq:extended PV b4}\\
    b_3(w, x, a, a') = \frac{\tpr(x)\pob(a, w\given x, R_X=1)}{\pob(x, a, w\given (R_W, R_X)=\ind)}.\label{eq:extended PV b3}
\end{gather}
For (b), with $b_3$ defined in \eqref{eq:extended PV b3} we have
\begin{align}
    (b) = -\EEob\sbr{\cT_3\vhpessi{\piestar}(A, W, X, A')b_3(W, X, A, A')\given (R_W, R_X)=\ind}\le \nbr{\cT_3\vhpessi{\piestar}}_{\mu_3, 2}\nbr{b_3}_{\mu_3, 2}. \label{eq:extended PV (b)}
\end{align}
Similarly, for (c) with $b_4$ defined in \eqref{eq:extended PV b4} we have
\begin{align}
    (c) = \EEob\sbr{-\cT_4\vhpessi{\piestar}(X, A')b_4(X, A')\given R_X=1}\le \nbr{\cT_4\vhpessi{\piestar}}_{\mu_4, 2}\nbr{b_4}_{\mu_4, 2}.\label{eq:extended PV (c)}
\end{align}
In addition, it holds for (a) that
\begin{align}
    (a)&=\int_{\cX}\EEob\sbr{\sum_{a'\in\cA}\rbr{\h{2}{\piestar}(a', W, X)-\hpessik{2}{\piestar}(a', W, X)}\given X=x,  R_X=1} \tpr(x)\rd x\nend
    &=\int_{\cX}\EEob\sbr{\sum_{a'\in\cA}\EEob\sbr{\h{2}{\piestar}(A, W, X)-\hpessik{2}{\piestar}(A, W, X)\given X, U, A=a, (R_X, R_Z)=\ind}\given X=x, R_X=1}
    \nend
    &\quad \cdot \tpr(x) \rd x \nend
    &= \EEob\bigg[\EEob\sbr{\h{2}{\piestar}(A, W, X)-\hpessik{2}{\piestar}(A, W, X)\given X, U, A, (R_X, R_Z)=\ind} \nend
    &\quad \cdot \frac{\pob(U\given X, R_x=1)\tpr(X)}{\pob(U, X, A\given (R_X, R_Z)=\ind)}\given (R_X, R_Z)=\ind\bigg], \label{eq:extended PV 4}
\end{align}
where the second equality holds by noting that $R_X\indep W\given X$, $A\indep W\given (X, U)$, and $(R_X, R_Z)\indep W\given (A, U, X)$. 
which prompts us to introduce another ratio function. 
Since $Z$ is overcomplete over $U$, there exists $b_1: \cX\times\cA\times\cZ\rightarrow \RR$ such that
\begin{align}
    \EEob\sbr{b_1(X, A, Z)\given X, U, A, (R_X, R_Z)=\ind} = \frac{\pob(U\given X, R_x=1)\tpr(X)}{\pob(U, X, A\given (R_X, R_Z)=\ind)}.\label{eq:extended PV 5}
\end{align}
Plugging \eqref{eq:extended PV 5} into \eqref{eq:extended PV 4}, it holds that
\begin{align*}
    (a) 
    &= \EEob\bigg[\EEob\sbr{\h{2}{\piestar}(A, W, X)-\hpessik{2}{\piestar}(A, W, X)\given X, U, A, (R_X, R_Z)=\ind}\nend
    & \quad \cdot \EEob\sbr{b_1(X, A, Z)\given X, U, A, (R_X, R_Z)=\ind}\given (R_X, R_Z)=\ind\bigg]\nend
    & = \EEob\sbr{\rbr{\h{2}{\piestar}(A, W, X)-\hpessik{2}{\piestar}(A, W, X)}b_1(X, A, Z)\given  (R_X, R_Z)=\ind}\nend
    & = \EEob\sbr{\EEob\sbr{\h{2}{\piestar}(A, W, X)-\hpessik{2}{\piestar}(A, W, X)\given X, A, Z, (R_X, R_Z)=\ind} b_1(X, A, Z)\given  (R_X, R_Z)=\ind},
\end{align*}
where the second equality holds by noting that $W\indep (Z, R_Z)\given (X, A, U, R_X)$. Now plugging in \eqref{eq:extended PV 1}, we have
\begin{align*}
    (a) &= -\EEob\bigg[\rbr{\cT_1\vhpessi{\piestar}(A, X, Z) + \EEob\sbr{\cT_2\vhpessi{\piestar}(A, W, X, Z)\given A, X, Z, (R_X, R_Z)=\ind}}\nend
    &\quad \cdot b_1(X, A, Z)\given (R_X, R_Z)=\ind\bigg]\nend
    &= -\EEob\sbr{\cT_1\vhpessi{\piestar}(A, X, Z) b_1(X, A, Z)\given (R_X, R_Z)=\ind}\nend
    &\quad - \EEob\sbr{\EEob\sbr{\cT_2\vhpessi{\piestar}(A, W, X, Z)b_1(X, A, Z)\given A, W, X, (R_W, R_X, R_Z)=\ind}\given (R_X, R_Z)=\ind}\nend
    &= -\EEob\sbr{\cT_1\vhpessi{\piestar}(A, X, Z) b_1(X, A, Z)\given (R_X, R_Z)=\ind}\nend
    &\quad - \EEob\sbr{\cT_2\vhpessi{\piestar}(A, W, X, Z)b_1(X, A, Z)\frac{\pob(A, W, X\given (R_X, R_Z)=\ind)}{\pob(A, W, X\given (R_W, R_X, R_Z)=\ind)}\given (R_W, R_X, R_Z)=\ind} ,
\end{align*}
where the second equality holds by $R_W\indep (Z, R_X, R_Z)\given (A, W, X)$.
We thereby define $b_2: \cA\times\cW\times\cX\times\cZ\rightarrow \RR$ by
\begin{align*}
    b_2(A, W, X, Z) = b_1(X, A, Z)\frac{\pob(A, W, X\given (R_X, R_Z)=\ind)}{\pob(A, W, X\given (R_W, R_X, R_Z)=\ind)}.
\end{align*}
We then arrive at
\begin{align}
    (a)&=-\EEob\sbr{\cT_1\vhpessi{\piestar}(A, X, Z) b_1(X, A, Z)\given (R_X, R_Z)=\ind}\nend
    &\quad - \EEob\sbr{\cT_2\vhpessi{\piestar}(A, W, X, Z)b_2(A, W, X, Z)\given (R_W, R_X, R_Z)=\ind}\nend
    &\le \nbr{\cT_1\vhpessi{\piestar}}_{\mu_1, 2}\nbr{b_1}_{\mu_1, 2} + \nbr{\cT_2\vhpessi{\piestar}}_{\mu_2, 2}\nbr{b_2}_{\mu_2, 2}, \label{eq:extended PV (a)}
\end{align}
Combining \eqref{eq:extended PV (a)}, \eqref{eq:extended PV (b)} and \eqref{eq:extended PV (c)} with \eqref{eq:extended PV 3}, we have
\begin{align*}
    v^{\piestar} - v(\gpessi{\piestar})
    \le \sum_{k=1}^4 \nbr{\cT_k\vhpessi{\piestar}}_{\mu_k, 2}\nbr{b_k}_{\mu_k, 2} \le \max_{k\in\{1, 2, 3,4\}}\nbr{\cT_k\vhpessi{\piestar}}_{\mu, 2}  \sum_{k=1}^4\nbr{b_k}_{\mu_k, 2}, 
\end{align*}
Therefore, for the sub-optimality given in \eqref{eq:extended regret}, it holds by the conclusion of Theorem \ref{thm:Fast rate} that
\begin{align*}
    \SubOpt(\piepessi)\overset{\cE}{\lesssim} \sum_{k=1}^4 \nbr{b_k}_{\mu_k, 2} \cdot \big(\cO(\vareTheta) + \cO(\vareH) + \cO\rbr{\sqrt{e_\cD}} +  \cO\rbr{\eta}\big),
\end{align*}
which completes the proof of Theorem \ref{thm:extended PV subopt}.
\section{Completeness Conditions in Linear Inverse Problem}\label{app:linear inverse}
In this section, we discuss existence of a solution $h$ to the following linear inverse problem,
\begin{align}
    g(c, d, e)=\EE\sbr{h(A, B, C, d)\given B=b, C=c, E=e}.\label{eq:linear inverse problem}
\end{align}
Our discussion follows \citet[Theorem 2.41]{miao2018identifying, carrasco2007linear}.
Let $L^2(t, p)$ denote the space of all the square integrable functions of $t$ with respect to a distribution $p(t)$, which yields a Hilbert space with inner product $\langle g, h\rangle =\int g(t)h(t) p(t)\rd t$.
Let $K_{b,c}$ denote the conditional expectation operator:$L^2(F(a))\rightarrow L^2(F(d))$, $K_{b, c}f(e)=\EE\sbr{f(A)\given B=b, C=c, E=e}$ for $f\in L^2(a, p(a\given b, c, e))$.
Let $\{(\lambda_{b,c,n}, \phi_{b,c,n}, \psi_{b,c,n})\}_{n=1}^\infty$ denote a singular value decomposition of $K_{b, c}$.
By assuming the following four regularity conditions to hold,
\begin{itemize}
    \item[(i)] $\int\int f(a\given b, c, e)f(e\given b, c, a)\rd a\rd e<\infty$ for all $b, c$,
    \item[(ii)] $\int g(c, d, e)^2 p(e\given b, c)\rd e <\infty$ for all $b, c, d$,
    \item[(iii)] $\sum_{n=1}^\infty \lambda_{b, c, n}^{-2}\abr{\langle g(c, d, \cdot), \psi_{b, c, n}(\cdot)\rangle_{p(\cdot\given b, c)}}^2\le \infty$ for all $b, c, d$,
    \item[(iv)] $\EE\sbr{\sigma(E)\given A=a, B=b, C=c}\aseq 0$ for all $a, b, c$ if and only if $\sigma(e)\aseq 0$, 
\end{itemize}
the solution to \eqref{eq:linear inverse problem} must exist.

\subsection{Fenchel Duality}\label{lem:Fenchel}
By the property of Fenchel Duality, any convex function $f$ can be equivalently written as
\begin{align*}
    f(x)=\sup_{\zeta} x\cdot \zeta - f^*(\zeta),
\end{align*}
where $f^*$ is the Fenchel duality \citep{borwein2006convex}. In case of $f(x)=x^2/2$, the Fenchel duality is given by $f^*(z)=\zeta^2/2$. Therefore, we can equivalently express $\nbr{\cT_k\vh}^2_{\mu_k, 2}$ as 
\begin{align*}
    \nbr{\cT_k\vh}^2_{\mu_k, 2} 
    = \EE_{\mu_k}\sbr{\sup_{\zeta} \cT_k\vh(Z_k) \zeta - \frac 1 2 \zeta^2}.
\end{align*}
Note that the supremum is achieved when $\zeta=\cT_k\vh(Z_k)$.
By assuming $\cT_k\vh\in\Theta_k$, it follows that
\begin{align*}
    \nbr{\cT_k\vh}^2_{\mu_k, 2} 
    = \EE_{\mu_k}\sbr{\sup_{\theta_k\in\Theta_k} \cT_k\vh(Z_k) \theta_k(Z_k) - \frac 1 2 \theta_k(Z_k)^2}.
\end{align*}
Following the interchangeability principle \citep{dai2017learning}, we can therefore swap the position of expectation and supremum and obtain
\begin{align*}
    \nbr{\cT_k\vh}^2_{\mu_k, 2} 
    =\sup_{\theta_k\in\Theta_k} \EE_{\mu_k}\sbr{ \cT_k\vh(Z_k) \theta_k(Z_k) - \frac 1 2 \theta_k(Z_k)^2}.
\end{align*}
\section{Critical Radius}\label{app:critical radius}
In this section, we study the critical radius for the linear two-step DTRs and the linear one-step POMDP using the techniques given in \S\ref{app:technical}.
\subsection{Critical Radius for Linear Two-step DTRs}\label{app:DTRs critical radius}
In this section, we calculate the critical radius of the function class, 
\begin{align*}
    \cQ_k=\{ \alpha_k(\vh(\vX), \vY_k)\theta_k(\vZ_k): \vh\in\vH, \theta_k\in\Theta_k\}, 
\end{align*}
in the two-step DTRs example.
We first summarize the linear function classes as follows,
\begin{gather*}
    \cH_1=\{w_1\in\RR^{m_1}:\cA\times \cY\rightarrow w_1^\top\phi_1(\cdot), \nbr{w_1}_2\le C_1, \nbr{\phi_1(\cdot)}_2\le 1\}, \\
    \cG=\{w_2\in\RR^{m_2}:\cX\times \cA \rightarrow w_2^\top\phi_2(\cdot), \nbr{w_2}_2\le C_2, \nbr{\phi_2(\cdot)}_2\le 1\}, \\
    \Theta_1=\{\beta_1\in\RR^{d_1}:\cY_1\rightarrow \beta_1^\top \psi_1(\cdot), \nbr{\beta_1}\le D_1, \nbr{\psi_1(\cdot)}_2\le 1\},\\
    \Theta_2=\{\beta_2\in\RR^{d_2}:\cX\times\cA\times \cY_1\rightarrow \beta_2^\top \psi_2(\cdot), \nbr{\beta_2}\le D_2, \nbr{\psi_2(\cdot)}_2\le 1\}, \\
    \cQ_1=\{\cA\times\cY\rightarrow (h_1(a, y)-y_2)\theta_1(y_1): h_1\in\cH_1, \theta_1\in\Theta_1\}, \\
    \cQ_2=\{\cX\times\cA\times\cY\rightarrow (g(x,a)-h_1(a, y)) \theta_2(x, a, y_1): h_1\in\cH_1, g\in\cG, \theta_2\in\Theta_2\}.
\end{gather*}
Note that $h_1(a, y)-y$ can be captured by the following linear function class
\begin{align*}
    \cU_1=\cbr{\tilde w_1=
    \sqrt{2}\begin{bmatrix}
    w_1\\
    t
    \end{bmatrix}
    \in\RR^{m_1+1}:\cA\times\cY\rightarrow  \tilde w_1^\top \begin{bmatrix}
    \frac{\phi_1(a, y)}{\sqrt 2}\\
    \frac{y_2}{\sqrt 2 L_{Y_2}}
    \end{bmatrix}
    , \nbr{\tilde w_1}_2\le \sqrt{2\rbr{C_1^2+L_{Y_2}^2}}},
\end{align*}
and $g(x, a)-h_1(a, y)$ is captured by the following linear function class, 
\begin{align*}
    \cU_2=\cbr{\tilde w_2=
    \sqrt{2}\begin{bmatrix}
    w_1\\
    w_2
    \end{bmatrix}
    \in\RR^{m_1+m_2}:\cA\times\cY\rightarrow  \tilde w_2^\top \begin{bmatrix}
    \frac{\phi_1(a, y)}{\sqrt 2}\\
    \frac{\phi_2(x, a)}{\sqrt 2}
    \end{bmatrix}
    , \nbr{\tilde w_2}_2\le \sqrt{2\rbr{C_1^2+C_2^2}}}.
\end{align*}
By Lemma \ref{lem:critical radius for product}, the maximal critical radius for $\cQ_1$ and $\cQ_2$, which are denoted by $\eta_1$ and $\eta_2$, respectively, are bounded with probability $1-\delta$ by
\begin{gather*}
    \eta_1\le \cO\rbr{
    \sqrt{2\rbr{C_1^2+L_{Y_2}^2}
    \cdot 
    \frac{m_1+d_1+1}{T_1}\log \rbr{\frac{T_1}{m_1+d_1+1}}} + \sqrt{\frac{\log \rbr{1/\delta}}{T_1}}} ,\\
    \eta_2\le \cO\rbr{
    \sqrt{2\rbr{C_1^2+C_2^2}
    \cdot 
    \frac{m_1+m_2+d_2}{T_2}\log \rbr{\frac{T_2}{m_1+m_2+d_2}}} + \sqrt{\frac{\log \rbr{1/\delta}}{T_2}}}.
\end{gather*}

\subsection{Critical Radius for Linear One-step POMDP}\label{app:POMDP critical radius}
In this section, we calculate the critical radius of the function class 
\begin{align*}
    \cQ_k=\{ \alpha^\pie_k(\vh(\vX), \vY_k)\theta_k(\vZ_k): \vh\in\vH, \theta_k\in\Theta_k, \pie\in\Pie\}, 
\end{align*}
in the two-step DTRs example.
We first summarize the linear function classes as follows,
\begin{gather*}
\Pie \subseteq \cbr{\pie \bigg | \pie(a\given o, o^-)=\frac{\exp\rbr{w_0^\top\phi_0(a, o, o^-)}}{\sum_{a'\in\cA}\exp\rbr{w_0^\top\phi_0(a', o, o^-)}}, \nbr{w_0}_2\le C_0,  \nbr{\phi_0(\cdot)}_2\le 1}, \\
    \cH_k=\{ w_k^\top\phi_k(\cdot): w_k\in\RR^{m_k} \nbr{w_k}_2\le C_k, \nbr{\phi_k(\cdot)}_2\le 1\}, \quad k=1, 2, 3, \\
    \cG=\{w_4\in\RR^{m_4}:\cO^-\rightarrow w_4^\top\phi_4(\cdot), \nbr{w_4}\le C_3\nbr{W_6}_F, \nbr{\phi_4(\cdot)}_2\le 1\}, \\
    \Theta_k=\{\beta_k^\top \psi_k(\cdot): \beta_k\in\RR^{d_k},  \nbr{\beta_k}\le D_k, \nbr{\psi_k(\cdot)}_2\le 1\}, \quad k=1, 2, 3, \\
    \Theta_4=\{\beta_4\in\RR^{m_4}:\cO^-\rightarrow \beta_4^\top \phi_4(\cdot), \nbr{\beta_4}\le D_4, \nbr{\phi_4(\cdot)}_2\le 1\}, \\
    \cQ_k=\{ \alpha^\pie_k(\vh, y_k)\theta_k(z_k): \vh\in\vH, \theta_k\in\Theta_k, \pie\in\Pie\}, \quad k=1, 2, 3, 4,\\
    \sY=\cbr{f:\cY\rightarrow\RR\given f(y)=\lambda\cdot\frac{y}{L_Y}, \abr{\lambda}\le L_Y}.
\end{gather*}
Here $\cY$ can be viewed as an one-dimensional linear function class that $y$ falls into.
Notice that $\alpha^\pie_1=h_1$ is captured by $\vH_1$. Therefore, the critical radius of the product function class $\cQ_1$ is bounded by 
\begin{align*}
    \eta_1\le \cO\rbr{D_1C_1\sqrt{\frac{m_1+d_1}{T_1}\log\rbr{\frac{T_1}{m_1+d_1}}}}, 
\end{align*}
following Lemma \ref{lem:critical radius for product}.
For $\alpha^\pie_2=h_2-h_1-Y\pie$, we have
\begin{align}
    N(t;\cQ_2, \nbr{\cdot}_\cD)
    &\le N\rbr{\frac{t}{3};\cbr{h_2\theta_2},\nbr{\cdot}_\cD}\cdot N\rbr{\frac{t}{3};\cbr{h_1\theta_2},\nbr{\cdot}_\cD}\cdot  N\rbr{\frac{t}{3};\cbr{Y\pie\theta_2},\nbr{\cdot}_\cD}\nend
    &\le N\rbr{\frac{t}{6D_2};\cH_2,\nbr{\cdot}_\cD} N\rbr{\frac{t}{6C_2};\Theta_2,\nbr{\cdot}_\cD}\nend
    &\qquad \cdot N\rbr{\frac{t}{6D_2};\cH_1, \nbr{\cdot}_\cD}
    N\rbr{\frac{t}{6C_1};\Theta_2,\nbr{\cdot}_\cD}\nend
    &\qquad \cdot N\rbr{\frac{t}{9D_2};\sY,\nbr{\cdot}_\cD}
    N\rbr{\frac{t}{9L_Y D_2};\Pie, \nbr{\cdot}_\cD}
    N\rbr{\frac{t}{9L_Y};\Theta_2, \nbr{\cdot}_\cD},\label{eq:POMDP critic 1}
\end{align}
where the first inequality holds by Lemma \ref{lem:covering number summation} on the covering number of summation function class and the second inequality holds by Lemma \ref{lem:covering number product} on the covering number of product function class. Following  Lemma \ref{lem:covering number linear} on the covering number for the linear function class and Lemma \ref{lem:covering number Pie} on the covering number of the policy class, \eqref{eq:POMDP critic 1} is further bounded by
\begin{align*}
    \log N(t;\cQ_2,\nbr{\cdot}_\cD)
    \le a \log\rbr{1+\frac{C}{t}},
\end{align*}
where $a=7\max\{m_0, m_1, m_2, d_1, d_2\}$ and $C=\max\{12 C_2 D_2, 12 D_2 C_1, 12 C_1 D_2, 18 D_2 L_Y, 54 L_Y D_2, 18 L_Y D_2\}$.
By Lemma \ref{lem:critical radius & covering number}, the critical radius $\eta_2$ is bounded by
\begin{align*}
    \eta_2\le \cO\rbr{C\sqrt{\frac{a}{T_2}\cdot\log \frac{T_2}{a}}}.
\end{align*}
For $\alpha^\pie_3=h_3(y, a, o^-)-\sum_{a'\in\cA}h_2(a', o, o^-)$, it is captured by
\begin{align*}
    \cU_3=\cbr{\tilde w_3^\top 
    \begin{bmatrix}
    \frac{\sum_{a'\in\cA} \phi_2(a', o, o^-)}{\sqrt{2}|\cA|}\\
    \frac{\phi_3(y, a, o^-)}{\sqrt 2}
    \end{bmatrix}:
    \tilde w_3\in\RR^{m_2+m_3}, \nbr{\tilde w_3}_2\le \sqrt{2\rbr{|\cA|^2C_2^2+C_3^2}}
    }, 
\end{align*}
and $\alpha^\pie_4$ is captured by the following linear function class
\begin{align*}
    \cU_4=\cbr{
    \tilde w_4^\top \begin{bmatrix}
    \frac{\phi_3(a, y)}{\sqrt 2}\\
    \frac{\phi_4(x, a)}{\sqrt 2}
    \end{bmatrix}:
    \tilde w_4
    \in\RR^{m_3+m_4}, 
    \nbr{\tilde w_4}_2\le \sqrt{2\rbr{C_3^2+C_4^2}}}.
\end{align*}
By Lemma \ref{lem:critical radius for product} and Corollary 5 of \cite{dikkala2020minimax}, the maximum critical radius for $\cQ_k$ denoted by $\eta_k$ is bounded with probability $1-\delta$ by
\begin{gather*}
    \eta_1\le \cO\rbr{ D_1 C_1
    \sqrt{
    \frac{m_1+d_1}{T_1}\log \rbr{\frac{T_1}{m_1+d_1}}} + \sqrt{\frac{\log \rbr{1/\delta}}{T_1}}},\\
    \eta_2\le \cO\rbr{ C\sqrt{\frac{\max\{m_0, m_1, m_2, d_1, d_2\}}{T_2}\log\rbr{\frac{T_2}{\max\{m_0, m_1, m_2, d_1, d_2\}}}}+\sqrt{\frac{\log(1/\delta)}{T_2}}}, \\
    \eta_3\le \cO\rbr{ D_3
    \sqrt{ 2 \rbr{|\cA|^2C_2^2+C_3^2}
    \cdot 
    \frac{m_2+m_3+d_3}{T_3}\log \rbr{\frac{T_3}{m_2+m_3+d_3}}} + \sqrt{\frac{\log \rbr{1/\delta}}{T_3}}},\\
    \eta_4\le \cO\rbr{ D_4 
    \sqrt{ 2\rbr{C_3^2+C_4^2}
    \cdot 
    \frac{m_3+m_4}{T_4}\log \rbr{\frac{T_4}{m_3+m_4}}} + \sqrt{\frac{\log \rbr{1/\delta}}{T_4}}}.
\end{gather*}

\section{Technical Results}\label{app:technical}
\begin{lemma}[Corollary 14.3 in \cite{wainwright2019high}]
Let $N_\cD(t;\BB_\cD(\delta;\sF))$ denote the $t$-covering number of the set $\BB_\cD(\delta;\sF)=\{f\in\sF: \nbr{f}_\cD\le \delta\}$ in the empirical $L^2(\PP_\cD)$-norm. Then the empirical version of critical inequality
\begin{align*}
    \cR_\cD(\eta;\cF)\le \frac{\eta^2}{C}
\end{align*}
is satisfied for any $\eta>0$ such that 
\begin{align*}
    \frac{64}{\sqrt{n}}\int_{\frac{\eta^2}{2C}}^{\eta}\sqrt{\log N_\cD(t;\BB_\cD(\eta;\sF))}\rd t\le \frac{\eta^2}{C}.
\end{align*}
\end{lemma}

\begin{lemma}[Covering number for summation class]\label{lem:covering number summation}
Let $N(t;\cF,\nbr{\cdot})$ denote the $t$-covering number of a set $\cF$ on a metric space equipped with norm $\nbr{\cdot}$ such that the triangle inequality holds. For function classes $\cF_1, \cF_2$, let $\cF$ denote their summation class, 
\begin{align*}
    \cF=\cbr{f\given f=f_1+f_2, f_1\in\cF_1, f_2\in\cF_2}.
\end{align*}
The $t$-covering number for $\cQ$ satisfies
\begin{align}
    N(t;\cF, \nbr{\cdot})\le N\rbr{\frac{t}{2};\cF_1, \nbr{\cdot}} \cdot N\rbr{\frac{t}{2};\cF_2, \nbr{\cdot}}.\label{eq:covering number summation}
\end{align}
\begin{proof}
Suppose that $\Theta_1$ is a $t/2$-covering of $\cF_1$, $\Theta_2$ is a $t/2$-covering of $\cF_2$. We construct
\begin{align*}
    \Theta = \cbr{\theta\given \theta=\theta_1+\theta_2, \theta_1\in\Theta_1, \theta_2\in\Theta_2}.
\end{align*}
For any $f\in\cF$, there exist $f_1\in\cF_1, f_2\in\cF_2$ such that $f=f_1+f_2$. Moreover, by definition of the covering set, there exist $\theta_1\in\Theta_1, \theta_2\in\Theta_2$ such that $\nbr{f_1-\theta_1}\le t/2$ and that $\nbr{f_2-\theta_2}\le t/2$. Let $\theta\in\Theta$ such that $\theta=\theta_1+\theta_2$, it follows that 
\begin{align}
    \nbr{f-\theta}
    &= \nbr{(f_1+f_2)-(\theta_1+\theta_2)}\nend
    &\le \nbr{f_1-\theta_1} +\nbr{f_2-\theta_2}\nend
    &\le t,\label{app:tech 1}
\end{align}
where the first inequality holds by the triangle inequality of metric $\nbr{\cdot}$. It follows from \eqref{app:tech 1} that the product set $\Theta$ is a $t$-covering set of $\cF$. Therefore, we conclude that \eqref{eq:covering number summation} holds.
\end{proof}
\end{lemma}

\begin{lemma}[Covering number for product class]\label{lem:covering number product}
Let $N(t;\cF,\nbr{\cdot})$ denote the $t$-covering number of a set $\cF$ on a metric space equipped with norm $\nbr{\cdot}$ such that the triangle inequality holds. For uniformly bounded function classes $\cF_1, \cF_2$ such that $\nbr{\cF_1}_{\infty}\le C_1$ and $\nbr{\cF_2}_{\infty}\le C_2$, let $\cF$ denote their product class,
\begin{align*}
    \cF=\cbr{f\given f=f_1\cdot f_2, f_1\in\cF_1, f_2\in\cF_2}.
\end{align*}
The $t$-covering number for $\cF$ satisfies
\begin{align*}
    N(t;\cF, \nbr{\cdot})\le N\rbr{\frac{t}{2C_2};\cF_1, \nbr{\cdot}} \cdot N\rbr{\frac{t}{2C_1};\cF_2, \nbr{\cdot}}.
\end{align*}
\begin{proof}
Suppose that $\Theta_1$ is a $t/2C_2$-covering of $\cF_1$, $\Theta_2$ is a $t/2C_1$-covering of $\cF_2$. We construct
\begin{align*}
    \Theta = \cbr{\theta\given \theta=\theta_1\cdot \theta_2, \theta_1\in\Theta_1, \theta_2\in\Theta_2}.
\end{align*}
For any $f\in\cF$, there exist $f_1\in\cF_1, f_2\in\cF_2$ such that $f=f_1\cdot f_2$. Moreover, by definition of the covering set, there exist $\theta_1\in\Theta_1, \theta_2\in\Theta_2$ such that $\nbr{f_1-\theta_1}\le t/2C_2$ and that $\nbr{f_2-\theta_2}\le t/2C_1$.
Let $\theta\in\Theta$ such that $\theta=\theta_1\cdot\theta_2$, it follows that 
\begin{align*}
    \nbr{f-\theta}
    &= \nbr{(f_1\cdot f_2)-(\theta_1\cdot\theta_2)}\nend
    &\le \nbr{f_1-\theta_1}\cdot C_2 +\nbr{f_2-\theta_2}\cdot C_1\nend
    &\le t,
\end{align*}
where the first inequality holds by the triangle inequality of metric $\nbr{\cdot}$. It follows from \eqref{app:tech 1} that the product set $\Theta$ is a $t$-covering set of $\cF$.
\end{proof}
\end{lemma}

\begin{lemma}[Covering number for bounded linear class]\label{lem:covering number linear}
Suppose that $\cF$ is a bounded linear function class defined as
\begin{align*}
    \cF=\cbr{f\given f(x)=w^\top \phi(x), w\in\RR^d, \phi:\cX\rightarrow \RR^d, \nbr{\phi}_{2, \infty}\le 1, \nbr{w}_2\le C}.
\end{align*}
The covering number $N(t;\cF, \nbr{\cdot}_\cD)$ with respect to norm $\nbr{\cdot}_\cD$ is bounded by
\begin{align*}
    \log N(t;\cF,\nbr{\cdot}_\cD)\le d\log\rbr{1+\frac{2C}{t}}.
\end{align*}
\begin{proof}
This lemma is a conclusion of Example 5.8 in \cite{wainwright2019high} which states that 
\begin{align*}
    \log N(t;\BB, \nbr{\cdot})\le d\log\rbr{1+\frac 2 \delta},
\end{align*}
if $\BB$ is also a unit ball under norm $\nbr{\cdot}$. In case of $\cF$, we construct a function class $\cS$ defined as
\begin{align}
    \cS=\cbr{w^\top\phi\given \nbr{w^\top \phi(\cdot)}_{\cD}\le C}.\label{eq:covering unit ball}
\end{align}
It is straightforward that $\cF\subseteq\cS$. Therefore, it follows from \eqref{eq:covering unit ball} that
\begin{align*}
    \log N(t;\cF, \nbr{\cdot}_\cD)\le \log N(t;\cS, \nbr{\cdot}) \le  d\log\rbr{1+\frac{2C}{t}}
\end{align*}
\end{proof}
\end{lemma}

\begin{lemma}[critical radius for product of bounded linear function classes]\label{lem:critical radius for product}
Consider two linear function class $\cF_1$ and $\cF_2$ defined as
\begin{align*}
    \cF_1=\cbr{w_1\in\RR^{d_1}:\cX\rightarrow w_1^\top\phi_1(\cdot), \nbr{w_1}_2\le C_1, \nbr{\phi_1(\cdot)}_2\le 1}, \nend
    \cF_2 =\cbr{w_2\in\RR^{d_2}:\cX\rightarrow w_2^\top \phi_2(\cdot), \nbr{w_2}_2\le C_2, \nbr{\phi_2(\cdot)}_2\le 1}.
\end{align*}
The product space of $\cF_1$ and $\cF_2$ is defined as
\begin{align*}
    \cQ=\cbr{\cX\rightarrow f_1(\cdot) f_2(\cdot): f_1\in\cF_1, f_2\in\cF_2}, 
\end{align*}
and the critical radius of $\cQ$ is bounded by 
\begin{align*}
    \eta \le 64 b\sqrt{\frac{d_1+d_2}{n}\cdot \log \rbr{\rbr{1+8\max\cbr{\frac{1}{C_1}, \frac{1}{C_2}}}\frac{n}{64^2(d_1+d_2)}}}, 
\end{align*}
where $b =\sqrt{C\rbr{8\max\{C_1, C_2\}+C}}$ and $C=C_1C_2$.
\begin{proof}
Note that we always have $\nbr{f_1(\cdot)}_\infty\le \nbr{w_1}_2\nbr{\phi_1(\cdot)}_2\le C_1$ for any $f_1\in\cF_1$. It also holds that $\nbr{f_2(\cdot)}_\infty\le C_2$ and that $\nbr{q}_\infty\le C_1C_2=C$ for any $q\in\cQ$. The critical radius $\eta$ for $\cQ$ then satisfies,
\begin{align}
    \cR_\cD(\eta;\cQ)\le \frac{\eta^2}{C}.\label{eq:6 critical radius}
\end{align}
Let $N_\cD(t;\BB_\cD(\eta;\cQ))$ denote the $t$-covering number of the set $\BB_\cD(\eta;\cQ)=\{f\in\cQ: \nbr{f}_\cD\le \eta\}$ in the empirical $L^2(\PP_\cD)$-norm. Then the empirical version of critical inequality \eqref{eq:6 critical radius}
is satisfied for any $\eta>0$ such that 
\begin{align}
    \frac{64}{\sqrt{n}}\int_{\frac{\eta^2}{2C}}^{\eta}\sqrt{\log N_\cD(t;\BB_\cD(\eta;\cQ))}\rd t\le \frac{\eta^2}{C}.\label{eq:1 critical radius}
\end{align}
Such a property holds by Corollary 14.3 in \cite{wainwright2019high}. By definition of $\BB_\cD(\eta;\cQ)$, it holds directly 
\begin{align*}
    \log N_\cD(t;\BB_\cD(\eta;\cQ))&\le \log N_\cD(t;\cQ)\nend
    &\le \log N_\cD\rbr{\frac{t}{2C_2};\cF_1} + \log N_\cD\rbr{\frac{t}{2C_1};\cF_2}\nend
    &\le \log N_\cD\rbr{\frac{t}{2C_2};\cS_\cD(C_1;\phi_1)} + \log N_\cD\rbr{\frac{t}{2C_1};\cS_\cD(C_2;\phi_2)},
\end{align*}
where we define $\cS(C;\phi)=\cbr{w\in \RR^d:  w\in\RR^{d}, \nbr{w^\top\phi(\cdot)}_\cD\le C}$. 
Here, the second inequality holds by Lemma \ref{lem:covering number product} and the last inequality holds by noting that $\cF_1\in\cS(C_1;\phi_1)$ and $\cF_2\in\cS(C_2;\phi_2)$. Note that the norms corresponding to $\cS(C;\phi)$ and covering number $N_\cD$ are both $\nbr{\cdot}_\cD$. Thereby applying Lemma 5.7 in \cite{wainwright2019high}, it follows that,
\begin{align*}
    \log N_\cD(\delta;\cS_\cD(C;\phi))\le d\log \rbr{\frac {2C}{\delta} + 1}.
\end{align*}
Therefore, we show that
\begin{align*}
    \log N_\cD(t;\BB_\cD(\eta;\cQ))\le d_1\log\rbr{\frac{4C_2}{t}+1} + d_2\log\rbr{\frac{4C_1}{t}+1}.
\end{align*}
The left hand-side of \eqref{eq:1 critical radius} is bounded by
\begin{align}
    \frac{64}{\sqrt{n}}\int_{\frac{\eta^2}{2C}}^{\eta}\sqrt{\log N_\cD(t;\BB_\cD(\eta;\cQ))}\rd t &\le \frac{64}{\sqrt{n}} \eta \sqrt{\log N_\cD\rbr{\frac{\eta^2}{2C};\BB_\cD(\eta;\cQ)}}\nend
    &\le \frac{64}{\sqrt{n}} \eta \sqrt{
    d_1\log\rbr{\frac{8C C_2}{\eta^2}+1} + d_2\log\rbr{\frac{8C C_1}{\eta^2}+1}
    }\nend
    &\le \frac{64}{\sqrt{n}} \eta \sqrt{
    (d_1+d_2)\log\rbr{\frac{8C\max\cbr{C_1, C_2}}{\eta^2}+\frac{C^2}{\eta^2}}
    },
    \label{eq:3 critical radius}
\end{align}
where the last inequality holds by noting that $\eta<C$.
Therefore, an upper bound for the critical radius is given by  plugging \eqref{eq:3 critical radius} into \eqref{eq:1 critical radius},
\begin{align}
    \frac{64}{\sqrt{n}} \eta \sqrt{
    (d_1+d_2)\log\rbr{\frac{8C\max\cbr{C_1, C_2}}{\eta^2}+\frac{C^2}{\eta^2}}}
    \le \frac{\eta^2}{C}. \label{eq:4 critical radius}
\end{align}
A little transformation of \eqref{eq:4 critical radius} gives
\begin{align*}
    a \log \frac{b^2}{\eta^2}\le 
    \frac{\eta^2}{b^2}, 
\end{align*}
where 
\begin{align}
    a=\sqrt{\frac{C}{8\max\{C_1, C_2\}+C}} \frac{64^2}{n}(d_1+d_2), \quad b =\sqrt{C\rbr{8\max\{C_1, C_2\}+C}}.\label{eq:5 critical radius}
\end{align}
By assuming that $a<1/2$, we see that $\eta=b \sqrt{a\log \frac 1 a} $ satisfies \eqref{eq:5 critical radius}.
Therefore, the critical radius $\eta$ is upper bounded by
\begin{align*}
    \eta &\le  64 b\sqrt{\frac{d_1+d_2}{n}\cdot \log \rbr{\rbr{1+8\max\cbr{\frac{1}{C_1}, \frac{1}{C_2}}}\frac{n}{64^2(d_1+d_2)}}}\nend
    &\le \cO\rbr{C\sqrt{\frac{d_1+d_2}{n}\log \rbr{\frac{n}{d_1+d_2}}}}.
\end{align*}
\end{proof}
\end{lemma}

\begin{lemma}[Covering number for the policy class $\Pie$ in \eqref{def:linear policy}]
For the policy class\label{lem:covering number Pie}
\begin{align*}
    \Pie \subseteq \cbr{\pie \bigg | \pie(a\given o, o^-;w_0)=\frac{\exp\rbr{w_0^\top\phi_0(a, o, o^-)}}{\sum_{a'\in\cA}\exp\rbr{w_0^\top\phi_0(a', o, o^-)}}, w_0\in\RR^{m_0}, \nbr{w_0}_2\le C_0,  \nbr{\phi_0(\cdot)}_{2, \infty}\le 1},
\end{align*}
the covering number $N(t;\Pie, \nbr{\cdot}_\cD)$ is bounded by,
\begin{align*}
    \log N(t;\Pie, \nbr{\cdot}_\cD)\le m_0 \log\rbr{1+\frac{6 C_0}{t}}.
\end{align*}
\begin{proof}
For simplicity, let $\zeta(a, \tau;w)=\exp(w^\top \phi_0(a, o, o^-))$ where $\tau=(o, o^-)$. For $w$ and $w'$ with bounded quadratic norms, the policy difference of such a softmax policy class can be bounded by
\begin{align}
    &\abr{\pie(a\given \tau;w)-\pie(a\given \tau;w')}\nend
    &\le \frac{\abr{\zeta(a, \tau;w)-\zeta(a, \tau;w')}\sum_{a'}\zeta(a', \tau;w') + \zeta(a, \tau;w')\abr{\sum_{a'}\zeta(a', \tau;w')-\sum_{a'}\zeta(a',\tau;w)}}{\sum_{a'}\zeta(a', \tau;w) \cdot \sum_{a'}\zeta(a', \tau;w')}.\label{eq:pie diff}
\end{align}
Without loss of generality, we assume that $\sum_{a'}\zeta(a',\tau;w')\le \sum_{a'}\zeta(a',\tau;w)$.
Therefore, \eqref{eq:pie diff} can be further bounded by
\begin{align}
    &\abr{\pie(a\given \tau;w)-\pie(a\given \tau;w')}\nend
    &\le \frac{\abr{\zeta(a, \tau;w)-\zeta(a, \tau;w')}}{\sum_{a'}\zeta(a', \tau;w)} + \frac{\sum_{a'}\abr{\zeta(a', \tau;w')-\zeta(a', \tau;w)}}{\sum_{a'}\zeta(a', \tau;w)}\nend
    &\le \frac{\abr{\zeta(a, \tau;w)-\zeta(a, \tau;w')}}{\max\cbr{\zeta(a, \tau;w), \zeta(a, \tau;w')}} + 2\cdot\frac{\sum_{a'}\abr{\zeta(a', \tau;w')-\zeta(a', \tau;w)}}{\sum_{a'}\zeta(a', \tau;w)+\sum_{a'}\zeta(a', \tau;w')}\nend
    &\le \frac{\abr{\zeta(a, \tau;w)-\zeta(a, \tau;w')}}{\max\cbr{\zeta(a, \tau;w), \zeta(a, \tau;w')}} + 2\cdot\frac{\sum_{a'}\abr{\zeta(a', \tau;w')-\zeta(a', \tau;w)}}{\sum_{a'}\max\cbr{\zeta(a', \tau;w),\zeta(a', \tau;w')}},\label{eq:pie diff 1}
\end{align}
where the second inequality holds by noting that $\sum_{a'}\zeta(a', \tau;w)\ge \sum_{a'}\zeta(a', \tau;w')\ge \zeta(a, \tau;w')$.
For given $w_1, w_2$, recall the definition of $\zeta$ and assume without loss of generality that $\zeta(a,\tau;w_1)\ge \zeta(a,\tau;w_2)$. We have
\begin{align*}
    \frac{\abr{\zeta(a, \tau;w_1)-\zeta(a, \tau;w_2)}}{\max\cbr{\zeta(a, \tau;w_1), \zeta(a, \tau;w_2)}}= 1-\exp\cbr{(w_2-w_1)^\top \phi_0(a,\tau)}
    \le \abr{(w_1-w_2)^\top \phi_0(a, \tau)}.
\end{align*}
Plugging the result into \eqref{eq:pie diff 1}, we conclude that
\begin{align*}
    \abr{\pie(a\given \tau;w)-\pie(a\given \tau;w')}\le 3 \nbr{(w-w')^\top \phi_0(\cdot, \tau)}_{\infty},
\end{align*}
which further yields
\begin{align}
\nbr{\pie(\cdot\given\cdot;w)-\pie(\cdot\given\cdot;w')}_\cD\le 
\nbr{\pie(\cdot\given\cdot;w)-\pie(\cdot\given\cdot;w')}_{\infty}\le 3\nbr{(w-w')^\top\phi_0(\cdot, \cdot)}_\infty.\label{eq:pie diff bound}
\end{align}
Following \eqref{eq:pie diff bound}, the covering number of $\Pie$ with respect to norm $\nbr{\cdot}_\cD$ is upper bounded by
\begin{align*}
    N(t;\Pie, \nbr{\cdot}_\cD)\le N\rbr{\frac t 3;\cF, \nbr{\cdot}_\infty}\le N\rbr{\frac t 3;\cS, \nbr{\cdot}_\infty},
\end{align*}
where $\cF=\{f\given f(a, \tau)=w^\top\phi_0(a,\tau), \nbr{w}_2\le C_0, \nbr{\phi_0(a, \tau)}_{2, \infty}\le 1\}$ and $\cS=\{w^\top \phi_0(a, \tau)\given \allowbreak\nbr{w^\top \phi_0(a, \tau)}_\infty\le C_0\}$.
The second inequality holds by noting that $\cF\subseteq\cS$.
Thus, by Example 5.8 in \cite{wainwright2019high} and noting that the covering number and function class $\cS$ are equipped with the same norm $\nbr{\cdot}_\infty$, we conclude that
\begin{align*}
    \log N(t;\Pie, \nbr{\cdot}_\cD)\le m_0 \log\rbr{1+\frac{6 C_0}{t}}.
\end{align*}
\end{proof}
\end{lemma}

\begin{lemma}[Bounding critical radius with covering number]\label{lem:critical radius & covering number}
If the covering number of $C_0$-uniformly bounded function class $\cF$ satisfies $N(t;\cF, \nbr{\cdot}_\cD)\le a\log\rbr{1+C_1/t}$, the maximal covering number of $\cF$ converges at a rate of $\cO(\sqrt{(a\log T)/T})$, where $C=\max\{C_0, C_1\}$ and $T$ is the size of $\cD$.
\begin{proof}
The critical radius $\eta$ for $\cF$ then satisfies,
\begin{align*}
    \cR_\cD(\eta;\cF)\le \frac{\eta^2}{C_0}.
\end{align*}
We denote an $\eta$-ball in the empirical $L^2(\PP_\cD)$-norm by $\BB_\cD(\eta;\cF)=\{f\in\cF\given \nbr{f}_\cD\le \eta\}$. By Corollary 14.3 in \cite{wainwright2019high}, an upper bound for $\eta$ is given by the following inequality,
\begin{align*}
    \frac{64}{\sqrt{T}}\int_{\frac{\eta^2}{2C_0}}^{\eta}\sqrt{\log N(t;\BB_\cD(\eta;\cQ),\nbr{\cdot}_\cD)}\rd t\le \frac{\eta^2}{C_0}.
\end{align*}
Since $\BB_\cD(\eta;\cF)\subseteq \cF$, the critical radius $\eta$ is further bounded by the following inequality
\begin{align*}
    \frac{64}{\sqrt{T}}\int_{\frac{\eta^2}{2C_0}}^{\eta}\sqrt{\log N(t;\cF,\nbr{\cdot}_\cD)}\rd t\le \frac{\eta^2}{C_0}.
\end{align*}
Noting that $\log N(t;\cF,\nbr{\cdot}_\cD) \le a\log(1+2C_1C_0/\eta^2)$ for $t\in(\eta^2/2C_0, \eta)$, we thus have $\eta$ bounded by
\begin{align}
    \frac{64}{\sqrt{T}}\cdot \eta \cdot \sqrt{a\log\rbr{1+\frac{2C_0C_1}{\eta^2}}}\le \frac{\eta^2}{C_0}.\label{eq:critic radius bound}
\end{align}
For simplicity, we use $C=\max\cbr{C_0, C_1}$ to replace $C_0, c_1$ and conclude that $\eta$ is bounded by
\begin{align*}
    \frac{2048 a}{T} \cdot {\log\rbr{1+\frac{2C^2}{\eta^2}}}\le \frac{\eta^2}{2C^2}
\end{align*}
satisfies \eqref{eq:critic radius bound}.
As $T\rightarrow\infty$, we have $T\ge (\sqrt{2}-1)2048a$. 
Thereby, it is easy to verify that
\begin{align*}
    \eta_0=64C\cdot\sqrt{\frac a T \log\rbr{1+\frac{T}{2048a}}}
\end{align*}
satisfies \eqref{eq:critic radius bound} and we conclude that $\eta\sim \cO(C\sqrt{(a\log T)/T})$.
\end{proof}
\end{lemma}

\end{document}